\documentclass{article}

\usepackage{microtype}
\usepackage{graphicx}
\usepackage{subcaption}
\usepackage{xcolor}
\usepackage{colortbl}
\usepackage{booktabs} 
\usepackage{multirow}
\usepackage{wrapfig}
\usepackage{subcaption}

\usepackage{hyperref}

\usepackage[accepted]{icml2025}

\usepackage{amsmath}
\usepackage{amssymb}
\usepackage{mathtools}
\usepackage{amsthm}

\usepackage[capitalize,noabbrev]{cleveref}

\theoremstyle{plain}

\theoremstyle{definition}

\theoremstyle{remark}

\usepackage[textsize=tiny]{todonotes}

\icmltitlerunning{Predicting High-precision Depth on Low-Precision Devices Using 2D Hilbert Curves}

\begin{document}

\twocolumn[
\icmltitle{Predicting~High-precision~Depth~on~Low-Precision Devices Using~2D~Hilbert~Curves}




\begin{icmlauthorlist}
\icmlauthor{Mykhailo~Uss}{srukr,khnau}
\icmlauthor{Ruslan~Yermolenko}{srukr,knu}
\icmlauthor{Oleksii~Shashko}{srukr,nure}
\icmlauthor{Olena~Kolodiazhna}{ntuu}
\icmlauthor{Ivan~Safonov}{srukr}
\icmlauthor{Volodymyr~Savin}{srukr,ntuu}
\icmlauthor{Yoonjae~Yeo}{sr}
\icmlauthor{Seowon~Ji}{konkuk}
\icmlauthor{Jaeyun~Jeong}{sr}
\end{icmlauthorlist}

\icmlaffiliation{srukr}{Samsung R\&D Institute Ukraine, Kyiv 01032, Ukraine}
\icmlaffiliation{sr}{Samsung Research, Seoul 06765, Republic of Korea}
\icmlaffiliation{khnau}{Department of Information-Communication Technologies, National Aerospace University, Kharkiv 61070, Ukraine}
\icmlaffiliation{ntuu}{Institute of Physics and Technology, NTUU "Igor Sikorsky Kyiv Polytechnic Institute", Kyiv 01032, Ukraine}
\icmlaffiliation{knu}{Faculty of Physics, Taras Shevchenko National University of Kyiv, Kyiv 01032, Ukraine}
\icmlaffiliation{nure}{Department of Artificial Intelligence, Kharkiv National University of Radio Electronics, Kharkiv 61070, Ukraine}
\icmlaffiliation{konkuk}{Department of Computer Science and Engineering, Konkuk University,  Seoul, Republic of Korea}

\icmlcorrespondingauthor{Mykhailo~Uss}{m.uss@samsung.com}

\icmlkeywords{Machine Learning, ICML}

\vskip 0.3in
]



\printAffiliationsAndNotice{}  

\begin{abstract}
Dense depth prediction deep neural networks (DNN) have achieved impressive results for both monocular and binocular data, but still they are limited by high computational complexity, restricting their use on low-end devices. For better on-device efficiency and hardware utilization, weights and activations of the DNN should be converted to low-bit precision. However, this precision is not sufficient to represent high dynamic range depth. In this paper, we aim to overcome this limitation and restore high-precision depth from low-bit precision predictions. To achieve this, we propose to represent high dynamic range depth as two low dynamic range components of a Hilbert curve, and to train the full-precision DNN to directly predict the latter. For on-device deployment, we use standard quantization methods and add a post-processing step that reconstructs depth from the Hilbert curve components predicted in low-bit precision. Extensive experiments demonstrate that our method increases the bit precision of predicted depth by up to three bits with little computational overhead. We also observed a positive side effect of quantization error reduction by up to 4.6 times. Our method enables effective and accurate depth prediction with DNN weights and activations quantized to eight-bit precision.
\end{abstract}
\vspace*{-0.3cm}

\begin{figure}[!ht]
    \centering
    \begin{subfigure}{0.48\columnwidth}
        \includegraphics[width=\columnwidth]{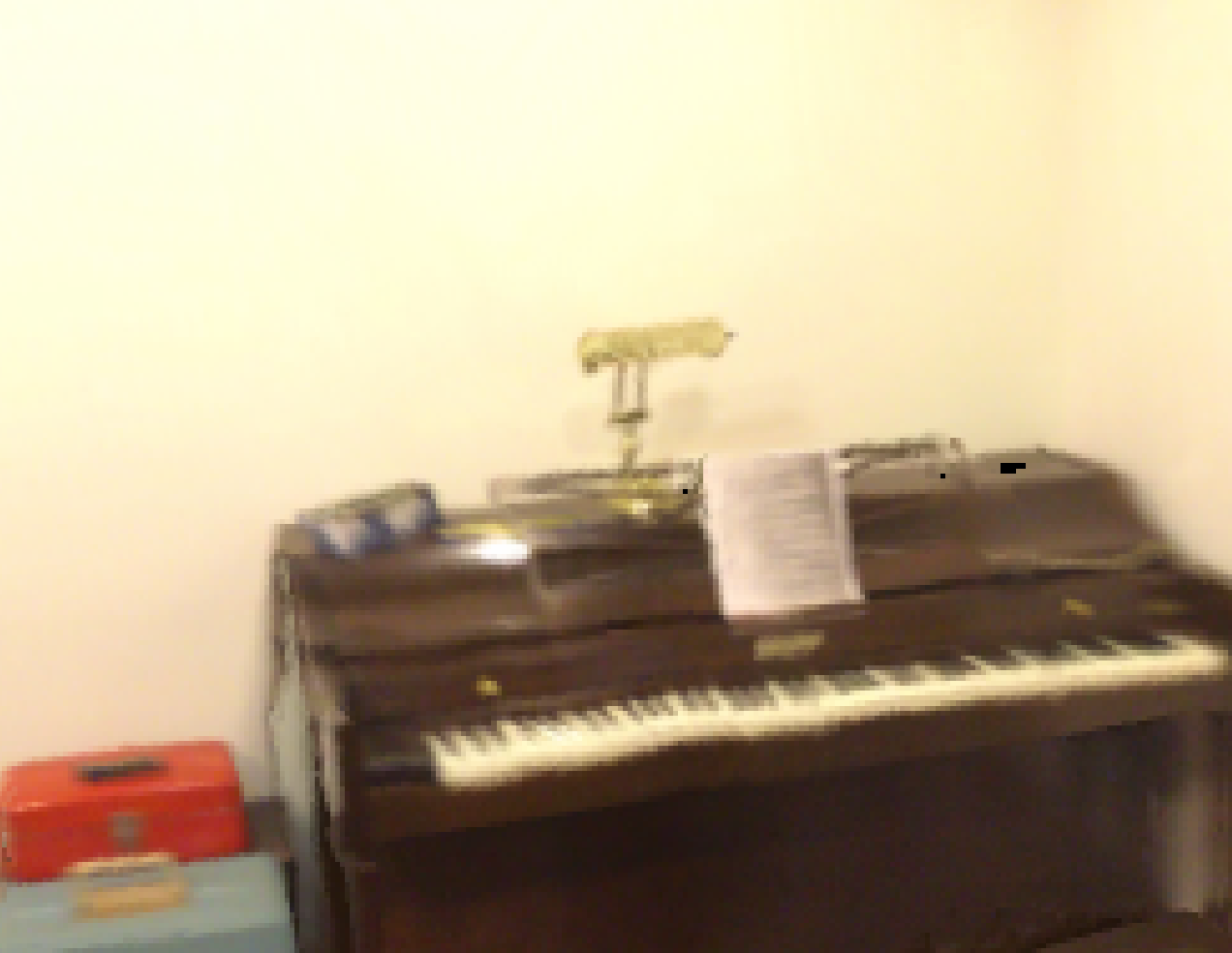}
        \caption{Input (left)}
        \label{fig:2-1-image}
    \end{subfigure}
    \centering
    \begin{subfigure}{0.48\columnwidth}
        \includegraphics[width=\columnwidth]{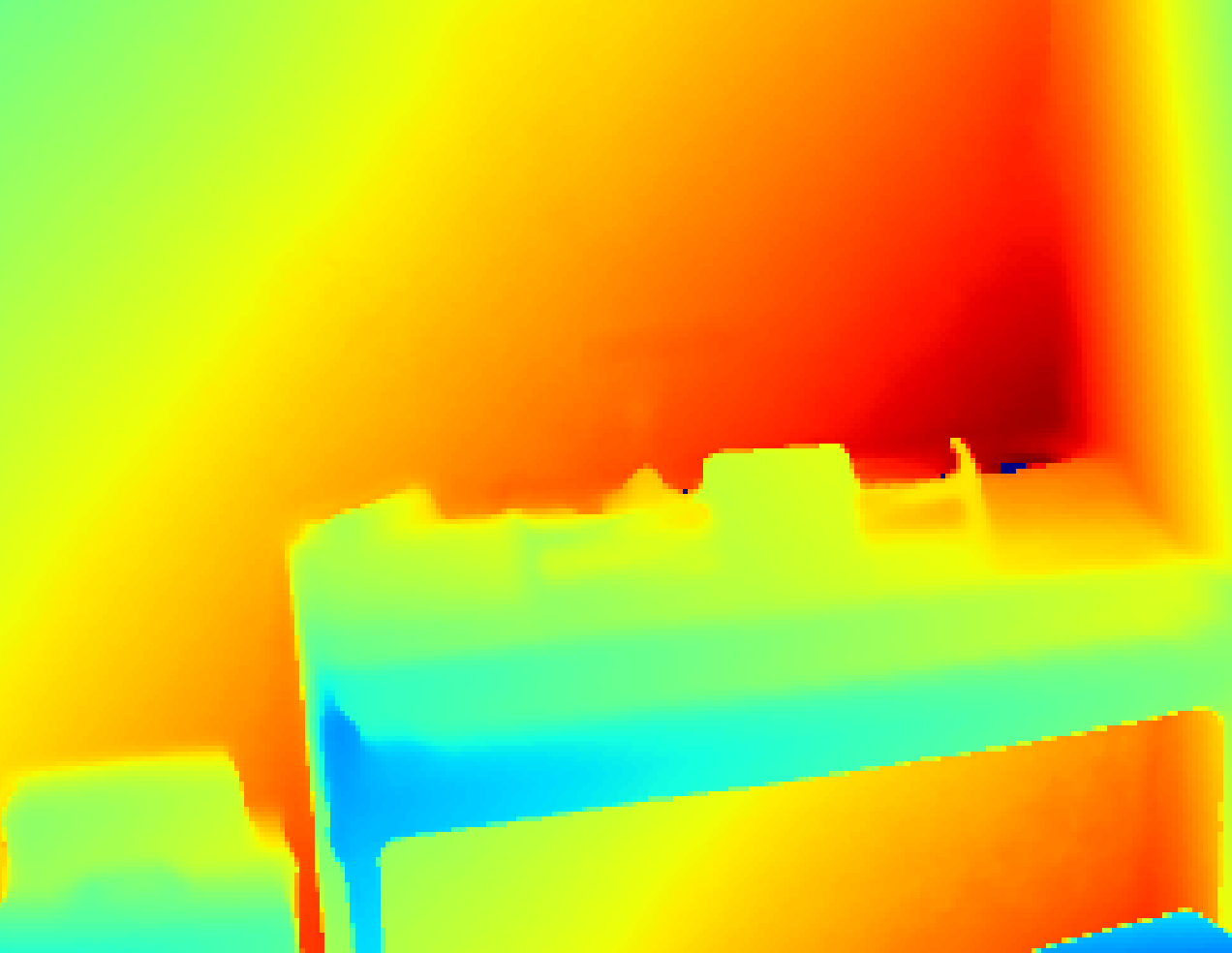}
        \caption{GT depth}
        \label{fig:2-1-gt depth}
    \end{subfigure}
    \centering
    \begin{subfigure}{0.48\columnwidth}
        \includegraphics[width=\columnwidth]{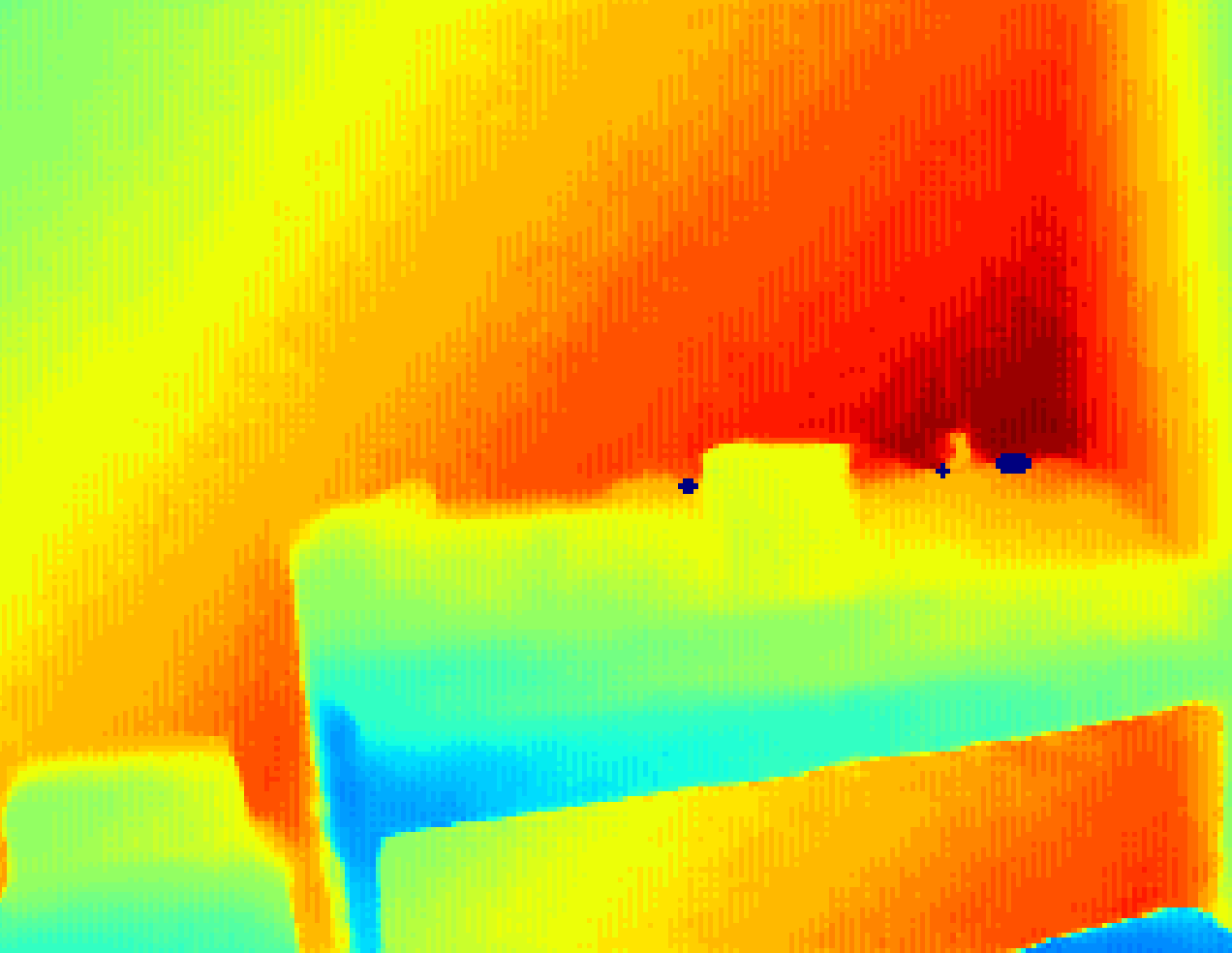}
        \caption{DispNet}
        \label{fig:2-1-dispnetnaseint8}
    \end{subfigure}
    \centering
    \begin{subfigure}{0.48\columnwidth}
        \includegraphics[width=\columnwidth]{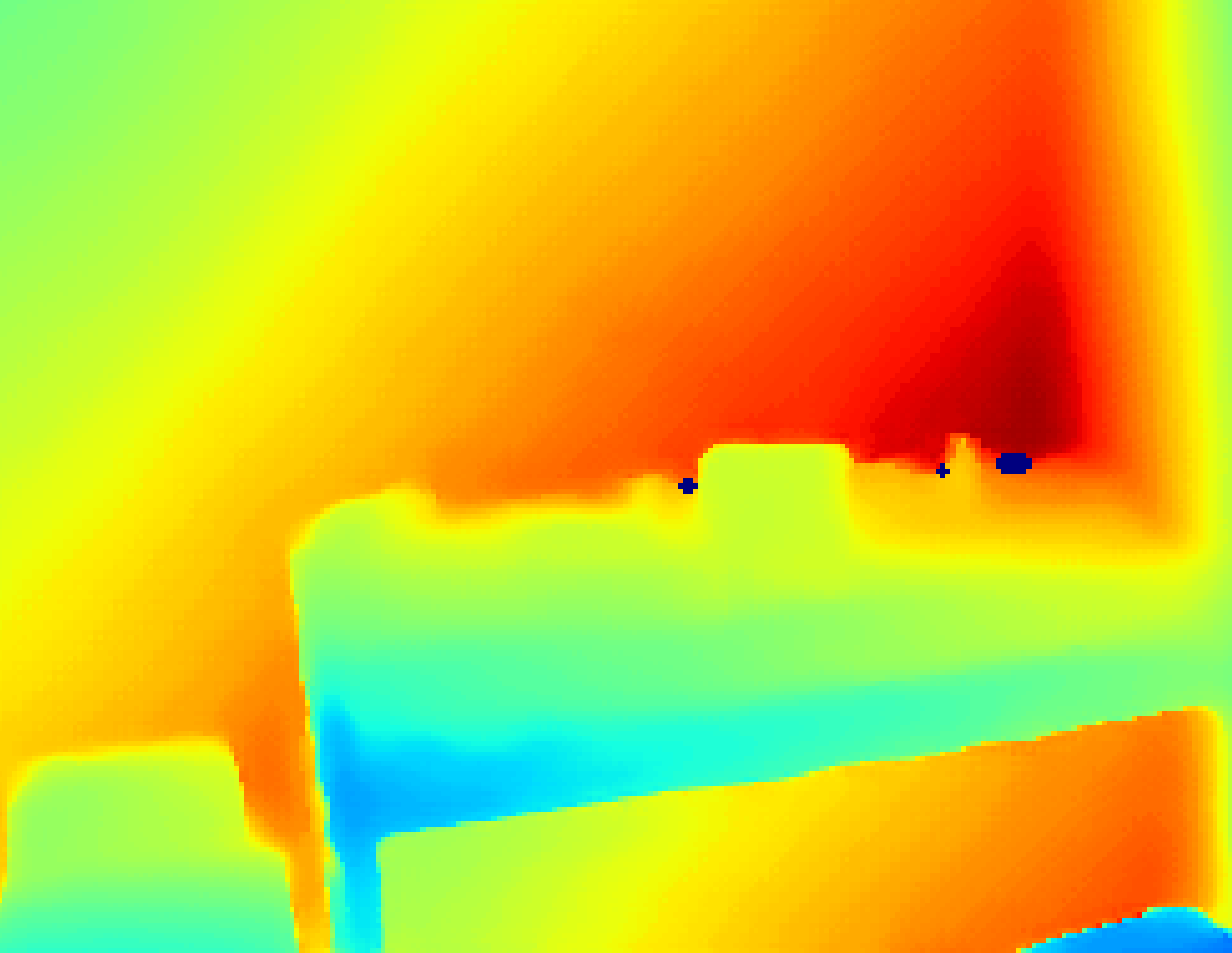}
        \caption{Our result}
        \label{fig:2-1-ourdispnetnaseint8}
    \end{subfigure}
    \caption{Illustration of DispNet~\cite{DispNet} quantization to INT8 precision (W8A8). Running inference of the quantized model on Qualcomm Hexagon~DSP results in depth precision loss and quantization artifacts~(c). Our method increases depth bit-width and reduces quantization error~(d).}
    \label{fig:1-1-abstract_demo}
    \vspace*{-0.4cm}
\end{figure}

\section{Introduction}
\label{sec:intro}

Dense depth prediction has received considerable attention in recent years due to its wide-ranging applicability in such areas as scene understanding~\cite{ScanNet,ScanNet++}, autonomous driving~\cite{PseudoLidarDriving, PseudoLiDAR++}, robotics~\cite{wofk2019fastdepth}, augmented reality / virtual reality~\cite{ARVR}, Internet of Things~\cite{Ignatov2022EfficientSD}, drones~\cite{DDOSDrones, ChannelAwareDT}. Usually, applications in these areas have strong limitations on hardware capabilities and energy consumption. Accurate depth estimation plays a vital role in enabling robust perception and decision-making processes in these contexts. A majority of recent developments in monocular~\cite{DenseTransformer, BEiT, DINOv2, Marigold, DepthAnything, DepthAnythingV2, ChronoDepth} and binocular~\cite{RAFTStereo, SelectiveStereo} depth estimation employ large-scale frameworks leveraging complex convolutional neural networks (including Convolutional Gated Recurrent Units)~\cite{crnn, RAFTStereo, SelectiveStereo}, transformer~\cite{DenseTransformer, BEiT, DINOv2} and diffusion~\cite{Marigold, DepthAnything, DepthAnythingV2, ChronoDepth} architectures. Furthermore, this trend stems from attempts to develop foundational models to solve the task. However, the impressive progress made so far provokes a significant increase in the complexity of the proposed solutions to achieve better accuracy, which typically demands considerable computational and memory resources. Hardware limitations cause additional difficulties with on-device deployment of these complex models.

The efficiency of DNN inference on low-end devices can be achieved by applying several strategies. One is to optimize a model architecture to reduce number of parameters or latency, avoid using computationally intensive layers, use network pruning~\cite{wofk2019fastdepth, oh2020rrnet, liu2023lightdepthnet, song2023spatial}. Another strategy consists in usage of low-precision computations~\cite{Li2021Unleashing, Jacob2018quant}. Low-bit fixed-point representation of DNN weights and activations, such as INT8, increases the number of operations run in parallel, reduces the amount of data transferred to and from memory, and decreases the energy consumption of Multiply-Accumulate (MAC) operation~\cite{nagel2021white}. When moving to lower bit-width precision, the memory amount reduces linearly, and the computational cost of matrix multiplications reduces quadratically ~\cite{nagel2021white}. Full-precision models are usually trained in FP32 precision and quantized to low-bit precision for on-device deployment. Quantization-aware training (QAT) emulates the quantization process during full-precision model training~\cite{yang2019quantization, sakr2022optimal, bhalgat2020lsq+, li2022q}; post-training quantization (PTQ) works on trained models and seeks to convert weights into integer values with minimum quality degradation~\cite{fang2020post, yuan2022ptq4vit}.

Running inference of quantized models comes at the cost of additional low-precision computation errors and data precision loss ~\cite{nagel2021white}. Considerable research has been devoted towards reducing quantization error by DNN’s architecture modification ~\cite{QuantizationFriendlyMN, apq}, quantization techniques improvement ~\cite{sakr2022optimal, bhalgat2020lsq+, li2022q, nagel2021white, fang2020post, yuan2022ptq4vit}, adaptation to specific hardware ~\cite{Shlezinger2018HardwareLimitedTQ, Kiyama2019AQN, Wang2018HAQHA}. The problem of data precision loss is largely unexplored, presumably because it does not pose a problem for models predicting RGB images, which are naturally presented as three eight-bit channels. Nevertheless, precision loss becomes an issue for High Dynamic Range images. This prevents the use of efficient homogeneous eight-bit quantization and forces to resort to less efficient mixed quantization schemes~\cite{hdrQuant}.

Depth maps require a high bit-width for accurate representation. For example, representing depth in the range $0\ldots10$~m with 1~cm accuracy requires ten bits. Depth represented in eight bits suffers from the false edges on planar surfaces (Fig.~\ref{fig:2-1-dispnetnaseint8}), loss of spatial details and distortion of objects with low depth contrast. Consequently, low-end devices with low-precision computations necessarily lead to on-device depth quality loss. To address this issue, in~\cite{jiang2022lowmemory} authors suggested keeping the final layer of the depth completion model at full floating-point precision while quantizing the weights and activations of other layers to either four-bit or eight-bit precision. This solution is not efficient because the last layer works in the highest spatial resolution and is computationally intensive. Low-end devices not always support mixed-precision quantization or different bit-width for different model layers. For example, SNPE library~\cite{SNPE_SDK}  for converting DNNs for inference on Qualcomm Hexagon Digital Signal Processor (DSP) supports the same bit-width for all layers (can be different for weights and for activations). The Coral Edge Tensor Processing Unit (TPU) supports only INT8 or UINT8 data. All operations with float-precision data are executed on CPU~\cite{CoralTPU}. It means that the depth prediction model on devices with low-precision arithmetic will suffer from depth precision loss independently of the quantization quality. This problem relates to the hardware limitations and cannot be solved by improving the quality of DNN quantization with PTQ/QAT.

In this work, we propose to overcome this limitation by using depth representation as points on a 2D Hilbert curve. This transform codes high dynamic range depth as two Hilbert curve components that can be represented in low-bit precision with minimum quality loss (Fig.~\ref{fig:1-1-abstract_demo}). Our method adds an on-device post-processing step that converts low-bit precision Hilbert components to higher precision depth. The post-processing does not involve addition or multiplication operations and can be implemented as a simple lookup table (LUT). The main difference of our method is that for improving quality of DNN inference on device we consider not only DNN architecture modification, training process (like QAT), efficient weights and activations quantization (like PTQ) but also structure of the signal predicted by the DNN. Our method does not modify quantization process by itself and is built on top of existing quantization methods (either PTQ or QAT); progress of these methods in specific domains or for specific architectures will bring improvement in the performance of our method as well.

Our main contributions can be summarized as follows:

(a) We propose a novel method for high-quality depth prediction on devices with low-precision arithmetic. The method consists of representing depth as 2D Hilbert curve components, training a full-precision model to predict 2D Hilbert curve components, and reconstructing high-precision depth from low-precision Hilbert curve components.

(b) We evaluate the proposed method and demonstrate that the modified model quantized to W8A8 format (8-bit weights and activations) can predict depth with similar or better quality than the original model in W8A16 format (8-bit weights and 16-bit activations). At the same time, the inference time and power consumption of the modified W8A8 model with post-processing are 1.5 times less than that of the original W8A16 model.

(c)	We observe that the application of the proposed method has additional positive side effect on the quantization quality, specifically quantization error reduction by up to 4.6 times.

\section{Method}
\label{sec:idea}

In this section, we introduce our method that allows high-precision depth prediction on devices with low-precision computations. The methods rests on depth representation as two components of a 2D Hilbert curve. On-device, these components are predicted in low-bit precision on TPU or NPU and then converted to high-bit precision depth using a simple post-processing algorithm run on CPU. In the following subsections, we describe the method in general (Sec.~\ref{sec:hp_depth_on_lp_devices}), discuss properties of the Hilbert curve and its utility for depth coding (Sec.~\ref{sec:selecting_curve}), explain implementation details (Sec.~\ref{sec:fb_transformations}), discuss expected change of quantization error (Sec.~\ref{sec:quatization_transformation}). Lastly, we present the loss function for training the modified model (Sec.~\ref{sec:model_loss_modifications}).

\subsection{High-precision depth on a low-precision device}
\label{sec:hp_depth_on_lp_devices}

Let us consider a DNN that predicts some quantity $q$ bounded in the range $[0,1]$. A quantized model runs on a low-end device that has an efficient DNN inferencing module (in our case, it is DSP) with $b$-bit output data representation and a general-purpose CPU with full-precision arithmetic. The DSP outputs value $q_\textrm{\tiny{quant.b}}$ in $b$-bit precision (e.g. INT8). The error between $q$ and $q_\textrm{\tiny{quant.b}}$ is called the quantization error. We denote the standard deviation (SD) of this error as $\sigma_\textrm{\tiny{quant}}$.

We seek to increase the bit-width of the predicted value of $q$ with minimum computational overhead. For the depth prediction case, $q$ corresponds to normalized depth or normalized disparity, depending on the DNN design. 

Due to DSP limitations, the bit-width of the model output cannot be increased, so we can operate only with the number and structure of the output channels. These channels should be transferred from DSP to CPU and used to reconstruct $q$ in higher bit precision. To make this scheme effective, the following key problems need to be solved: (1) minimize the amount of data transferred from DSP to CPU, (2) minimize the complexity of post-processing run on CPU. In the solution proposed in ~\cite{jiang2022lowmemory}, the last layer of the depth completion model is in float precision. This version is ineffective because it requires passing to the CPU a large amount of data from the penultimate layer and processing the high-resolution data associated with the last layer.

\begin{figure}[tb]
  \centering
  \includegraphics[width=1\linewidth]{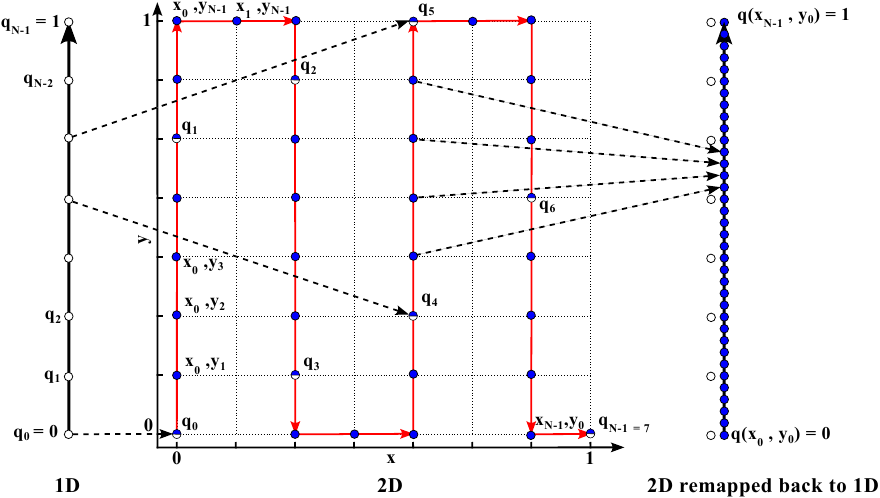}
  \caption{Idea illustration. 1D range is quantized to $N=8$ values $q_0=0\ldots q_{N-1}=1$ marked by white circles. The 1D range is mapped to a 2D curve shown in red color. Both $x$ and $y$ axes are also quantized into $N=8$ values yielding 64 2D points. Among them, 36 points lie on the curve (shown in blue color). Mapping the 2D curve back to the 1D range results in 36 different quantization values. Quantization error has effectively been reduced by the factor equal to the curve length $L=35/7=5$.}
  \label{fig:1D_vs_2D}
  \vspace*{-0.3cm}
\end{figure}

Our idea is to represent $q$ as a point on a 2D parametric curve $(x(q), y(q))$ with length $L > 1$, where both $x(q)$ and $y(q)$ are bounded in the $[0,1]$ range. The full-precision DNN is trained to directly predict $x(q)$ and $y(q)$. When running inference on DSP, the value of $q$ is calculated from $x$ and $y$ values predicted in $b$-bit precision. The parametric curve of length $L$ will pass through approximately $L \cdot 2^b$ discreet points $(x(q), y(q))$ effectively increasing precision of reconstructed value of $q$ by $\log_2 L$ bits from $b$ to $b + \log_2 L$ (Fig.~\ref{fig:1D_vs_2D}). In this implementation, the amount of data to transfer from DSP to CPU increases only twofold, post-processing on CPU is simple and implemented using LUT. We will later refer to unmodified DNN as original model and to model predicting Hilbert components (and post-processing step when it is clear from the context) as modified model.

\begin{figure}
    \centering
    \includegraphics[width=1\linewidth]{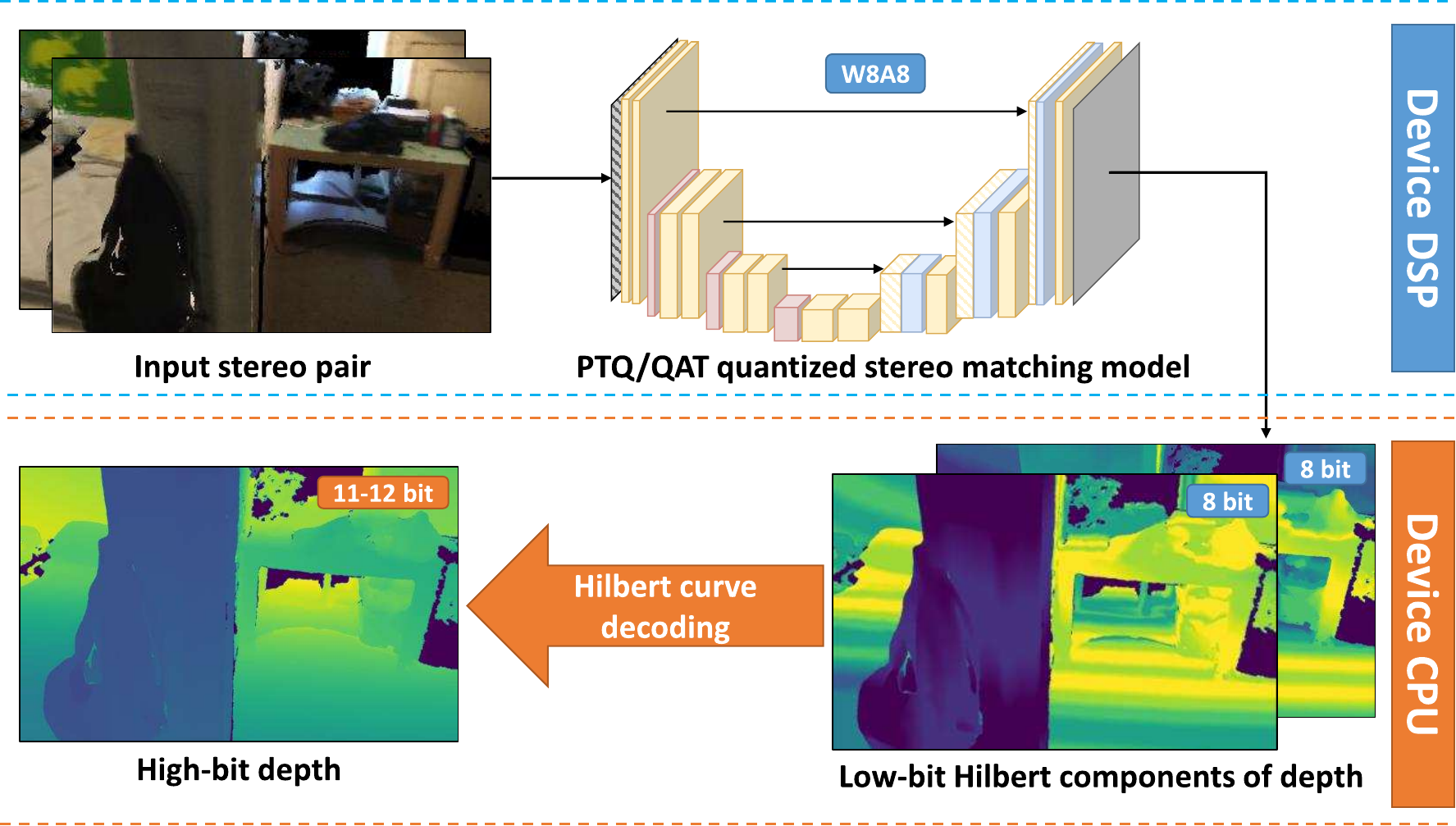}
    \caption{Scheme of the proposed method's inference pipeline on device.}
    \label{fig:inference-pipeline}
    \vspace*{-0.3cm}
\end{figure}

On-device deployment of a depth prediction DNN consists of the following steps (Fig. \ref{fig:inference-pipeline}): (a) training the full-precision model to directly predict the Hilbert curve components of depth, (b) applying standard quantization methods (either QAT during training or PTQ to trained model), (c) running inference of the modified quantized model on-device and obtaining Hilbert curve components in low-bit precision, (d) applying post-processing to Hilbert curve components and reconstruct depth in higher-bit precision.

Selection of function $(x(q), y(q))$ is crucial for successful implementation of the proposed idea.

\subsection{Selecting a suitable parametric curve}
\label{sec:selecting_curve}

\begin{figure*}[!htb]
    \centering
    \begin{subfigure}{0.2025\linewidth}
        \includegraphics[width=1.0\linewidth]{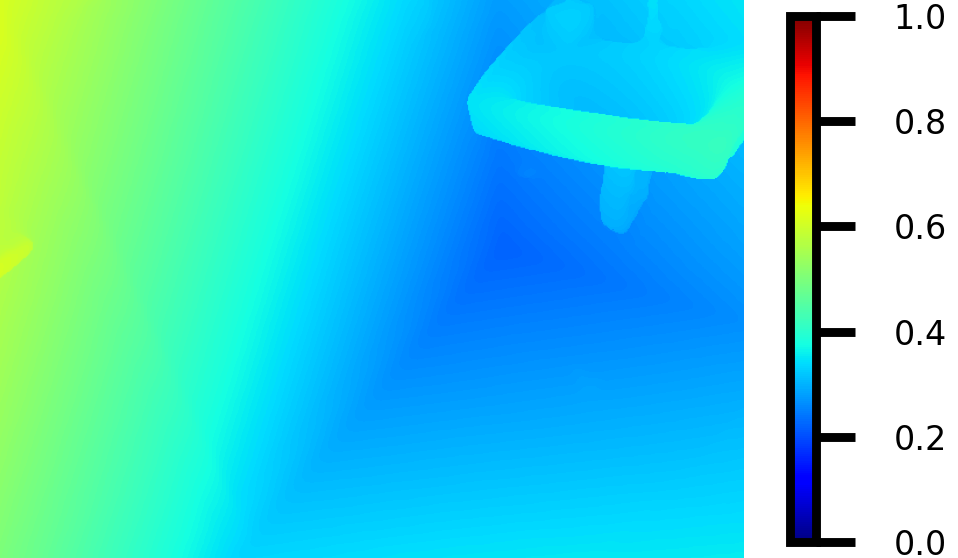}
        \caption{$\phantom{xxx}$}
        \label{fig:3-3-gt-disp}
    \end{subfigure}
    \centering
    \begin{subfigure}{0.1215\linewidth}
        \includegraphics[width=1.0\linewidth]{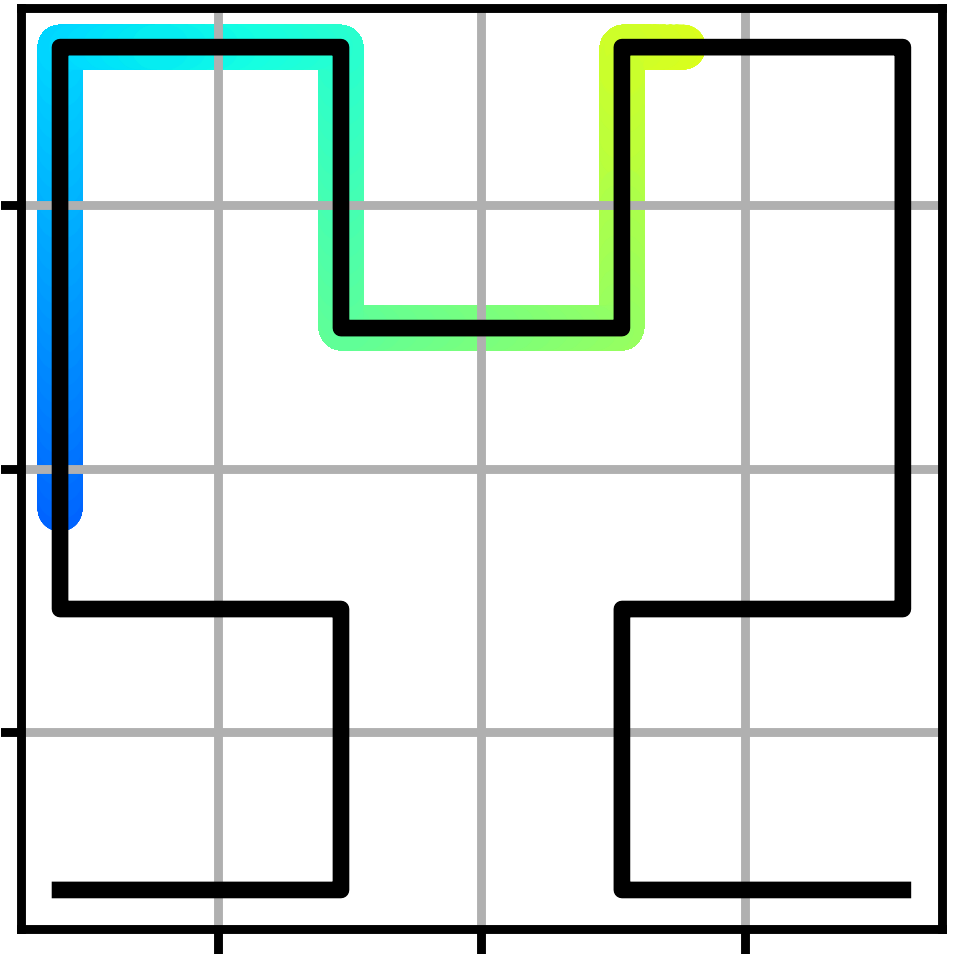}
        \caption{}
        \label{fig:3-3-h-curve-points}
    \end{subfigure}
    \centering
    \begin{subfigure}{0.16\linewidth}
        \includegraphics[width=1.0\linewidth]{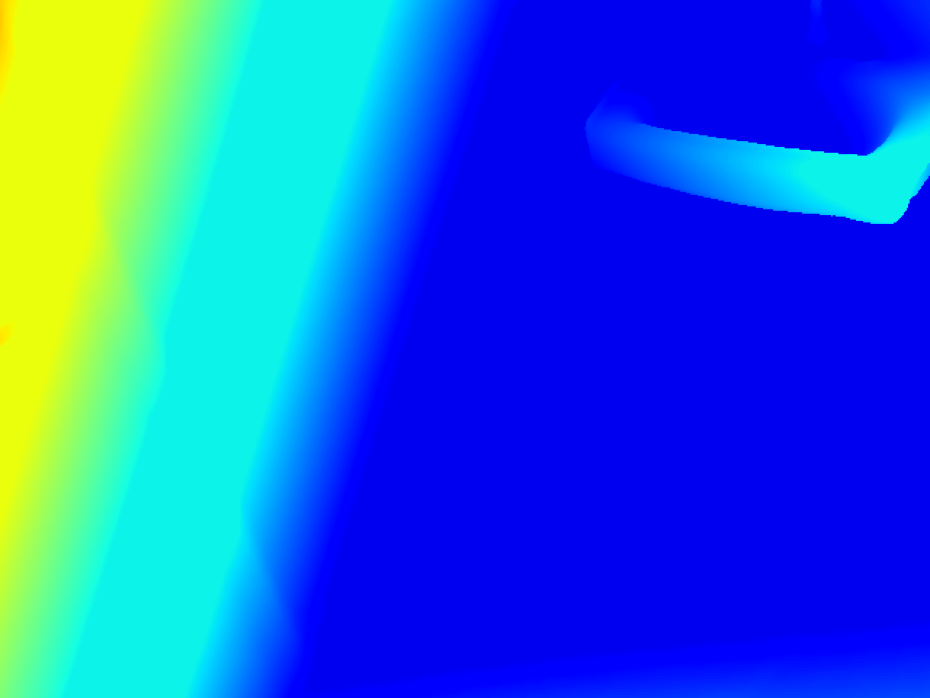}
        \caption{}
        \label{fig:3-3-h-x}
    \end{subfigure}
    \centering
    \begin{subfigure}{0.16\linewidth}
        \includegraphics[width=1.0\linewidth]{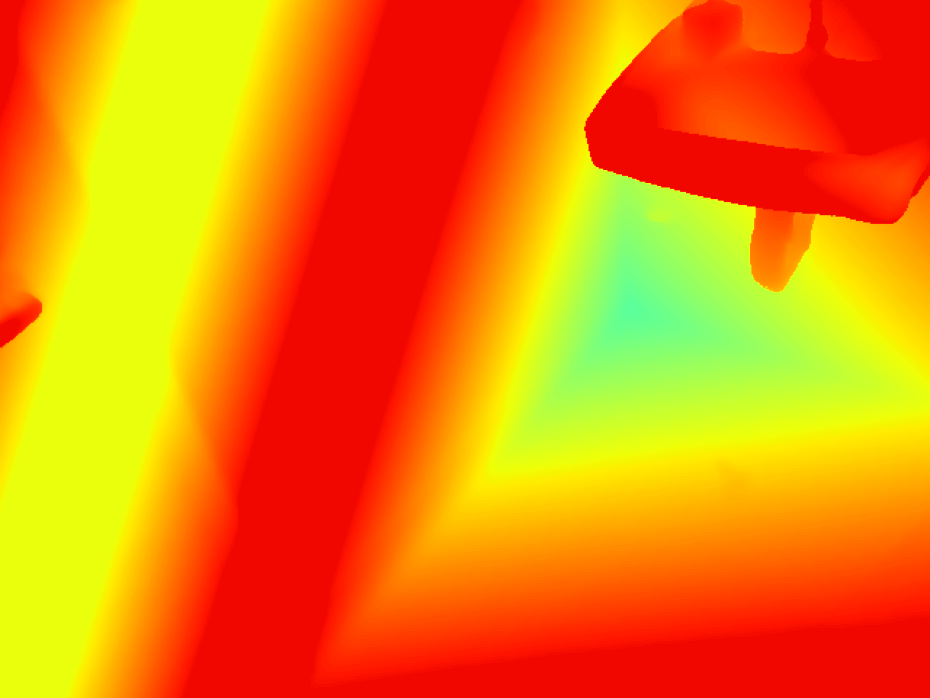}
        \caption{}
        \label{fig:3-3-h-y}
    \end{subfigure}
    \centering
    \begin{subfigure}{0.16\linewidth}
        \includegraphics[width=1.0\linewidth]{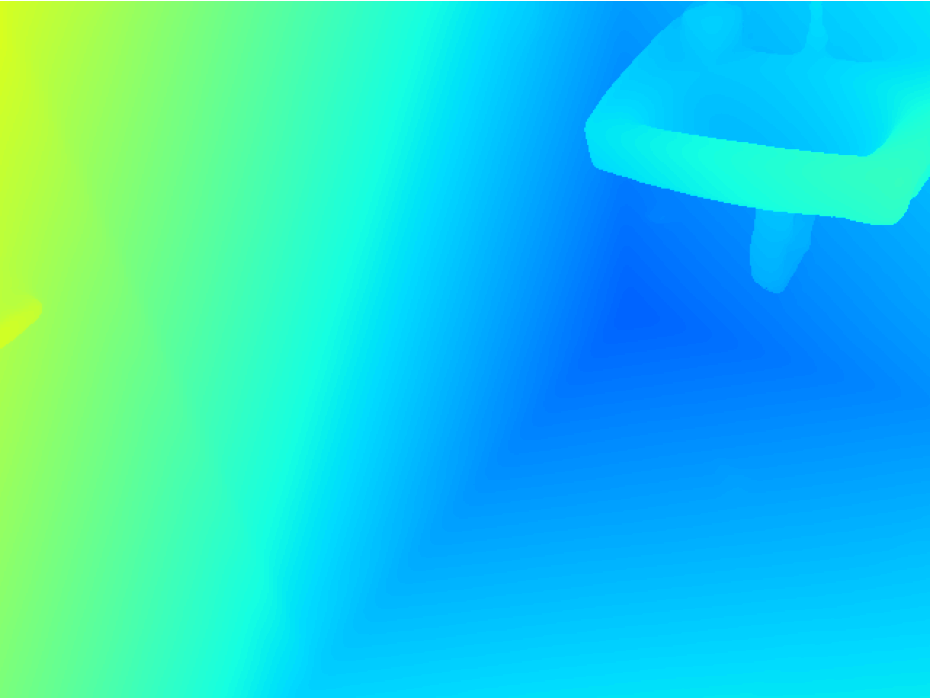}
        \caption{}
        \label{fig:3-3-coarse}
    \end{subfigure}
    \centering
    \begin{subfigure}{0.16\linewidth}
        \includegraphics[width=1.0\linewidth]{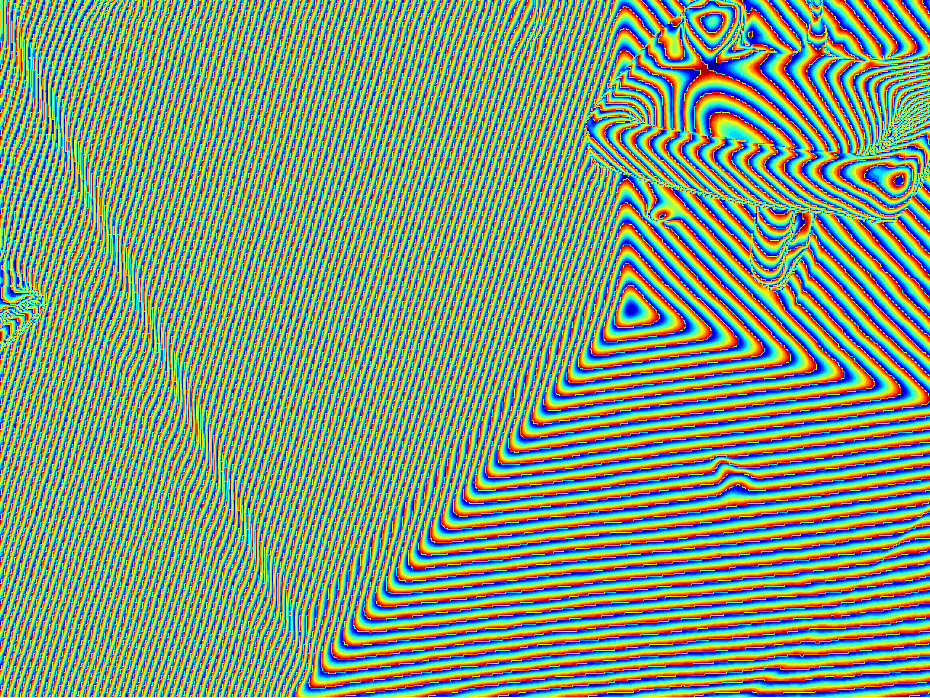}
        \caption{}
        \label{fig:3-3-fine}
    \end{subfigure}
    \caption{Illustration of disparity transforms: (a) disparity map; (b) mapping to 2D with second order Hilbert curve; (c, d) $x$ and $y$ components of the Hilbert curve; (e, f) coarse and fine details of disparity map. Fine details in (f) are the least significant byte of disparity (a) represented in 16-bit format. High-frequency oscillations make it appear different from the original disparity and difficult to predict by a DNN model.}
    \label{fig:3-3-disp-h-transform}
\end{figure*}

Let us consider the simplest function: $x(q) = \lfloor q * 255 \rfloor / 255, y(q) = q - x(q)$, where $\lfloor \cdot \rfloor$ denotes the floor function. In this case, $x(q)$ is a coarse representation of $q$ and $y(q)$ adds fine details (see example in Fig. \ref{fig:3-3-coarse} and \ref{fig:3-3-fine}). The length of this curve is 256, and it allows reconstructing $q$ in INT16 precision from $x(q)$ and $y(q)$ in INT8 precision. However, structure of $y(q)$ is unfriendly for the full-precision model training: it has multiple sharp transitions from zero to one and from one to zero. During our experiments we found that this curve degrades full-precision model training and is very susceptible to quantization errors. 

Let us discuss the desired properties of the curve $(x(q), y(q))$:
\begin{enumerate}
    \item Continuity: small changes in $q$ should result in small changes in both $x(q)$ and $y(q)$. For depth coding, this property ensures that $x(q)$ and $y(q)$ preserve spatial smoothness and do not introduce new depth discontinuities.
    \item Non-self-intersection: the curve should preserve one-to-one correspondence between $q$ and $(x(q), y(q))$.
    \item Boundedness and self-avoidance: the curve should cover the unit square uniformly and avoid close points $(x(q_1), y(q_1))$ and $(x(q_2), y(q_2))$ for distant $q_1$ and $q_2$.
\end{enumerate}

Curves with the desired properties are known as space-filling curves or Peano curves~\cite{sagan2012space}. For the goals of this work, we found one particular curve, namely the Hilbert curve~\cite{sagan2012space, bader2012space}, to be the simplest and the most flexible option. We describe this choice in more detail in Appendix~\ref{sec:curves}.

\begin{figure}[tb]
  \centering
  \includegraphics[width=1\linewidth]{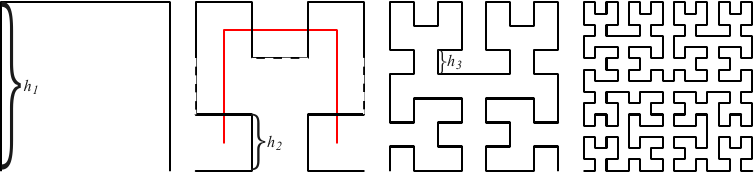}
  \caption{Hilbert curves for orders $p=1,2,3,4$ (from left to right). Every order is formed by the replacement of every node by an elementary 3-segment sequence.}
  \label{fig:Hilbert_curve_orders}
  \vspace*{-0.3cm}
\end{figure}

The Hilbert curve is a continuous fractal space-filling curve that is constructed as a limit of piece-wise linear curves~\cite{bader2012space}. The Hilbert curve starts from a single point in the middle of the unit square. Each subsequent curve order is produced by replicating and linking points of the curve of the previous order. We will later refer to the curve order as $p$. The approximating polygons for curves with orders $1-4$ are shown in Fig.~\ref{fig:Hilbert_curve_orders}. In order to avoid boundary effects, we scale each curve so that it fits into the square $\{(x, y) \, | \, k \le x, y \le 1-k\}$, where $k=0.1$. In this case, the length of the $p$-th order curve is $L_p=(2^p+1)(1 - 2k)$. The length of an edge of $p$-th order approximation polygon equals $h_p=(1 - 2k)/{(2^p-1)}$. This value also defines the minimum distance between points of different parallel edges of the Hilbert curve approximation polygon (Fig.~\ref{fig:Hilbert_curve_orders}).

\subsection{Forward and backward transformations}
\label{sec:fb_transformations}

For a full-precision and quantized models, predicted values $(x, y)$ will not have an exact match with any $(x(q), y(q))$. To convert arbitrary point $(x, y)$ back to 1D value, we link $(x, y)$ to the closest point on the Hilbert curve:
\begin{equation}
q_{xy} = \operatorname*{argmin}_{q\in [0, 1]} \lVert x - x(q), y - y(q) \rVert.
    \label{eq:hilbert_projection}
\end{equation}

Building forward (1D~$\rightarrow$~2D) and backward (2D~$\rightarrow$~1D) mappings for Hilbert curves of arbitrary order is based on iterative algorithms~\cite{bader2012space}. Because we work with one specific curve order, a faster transformation can be implemented with LUTs. We build two LUTs for corresponding mappings. The first LUT is built using bilinear interpolation of the nodes of the low-order Hilbert curve. This LUT allows us to get $(x(q), y(q))$ for the given $q$. An example of this transformation applied to the ground truth (GT) disparity map is shown in Fig.~\ref{fig:3-3-gt-disp}-\ref{fig:3-3-h-curve-points}. To map 2D values $(x, y)$ (Fig.~\ref{fig:3-3-h-x}-\ref{fig:3-3-h-y}) back to the 1D representation, we use a second LUT that is built based on Eq.~(\ref{eq:hilbert_projection}). In our experiments, this LUT has a size of 256 by 256 elements.

\subsection{Quantization error transformation}
\label{sec:quatization_transformation}

For the original model, the quantization error is measured at the model output level. For our method, this error is measured after post-processing of Hilbert components and accounts for Hilbert components quantization error and its transformation during post-processing. We denote SD of this error as $\bar\sigma_\textrm{\tiny{quant}}$ and SD of Hilbert components quantization error as $\sigma_\textrm{\tiny{xy.quant}}$.

Locally, Eq.~(\ref{eq:hilbert_projection}) reduces to the projection of $(x, y)$ to the closest Hilbert curve edge, discarding $x$ or $y$ component (depending on the edge orientation), compressing the remaining component $L$ times. We discussed above that this transform adds $\log_2 L$ bit to $q$ precision. In addition, Hilbert components quantization error is compressed $L$ times yielding $\bar\sigma_\textrm{\tiny{quant}} = \sigma_\textrm{\tiny{xy.quant}} / L$.

Quantization error compression takes place if (a) Hilbert components quantization error is independent between channels and identically distributed, and (b) if $\sigma_\textrm{\tiny{xy.quant}}$ is small enough for Hilbert components quantization error to be smaller than $h_p$. Both original and modified models have similar architecture, solve the same task, are trained on the same data, quantized by the same procedure. We can additionally assume (c) that $\sigma_\textrm{\tiny{quant}} = \sigma_\textrm{\tiny{xy.quant}}$. If this additional condition is valid then our method will reduce quantization error by $L$ times compared to the original model: $\bar\sigma_\textrm{\tiny{quant}} = \sigma_\textrm{\tiny{quant}} / L$.

We satisfy condition (b) by proper selection of the Hilbert curve order. For the stereo matching, we experimentally found that $p=2, 3$ are suitable choices, providing curve length 4--7.2 and bit-width increase by 2--2.85. Assumptions (a) and (c) can only be verified experimentally for a specific DNN architecture. We discuss their validity in detail in Section~\ref{sec:analysis_quat_reduction}. Notice, that bit-width increase will take place even if conditions (a) and (c) are not fulfilled. These additional conditions are only needed for potential quantization error reduction.

\subsection{Model and Loss Function Modification}
\label{sec:model_loss_modifications}

\begin{figure}
    \centering
        \includegraphics[width=1\linewidth]{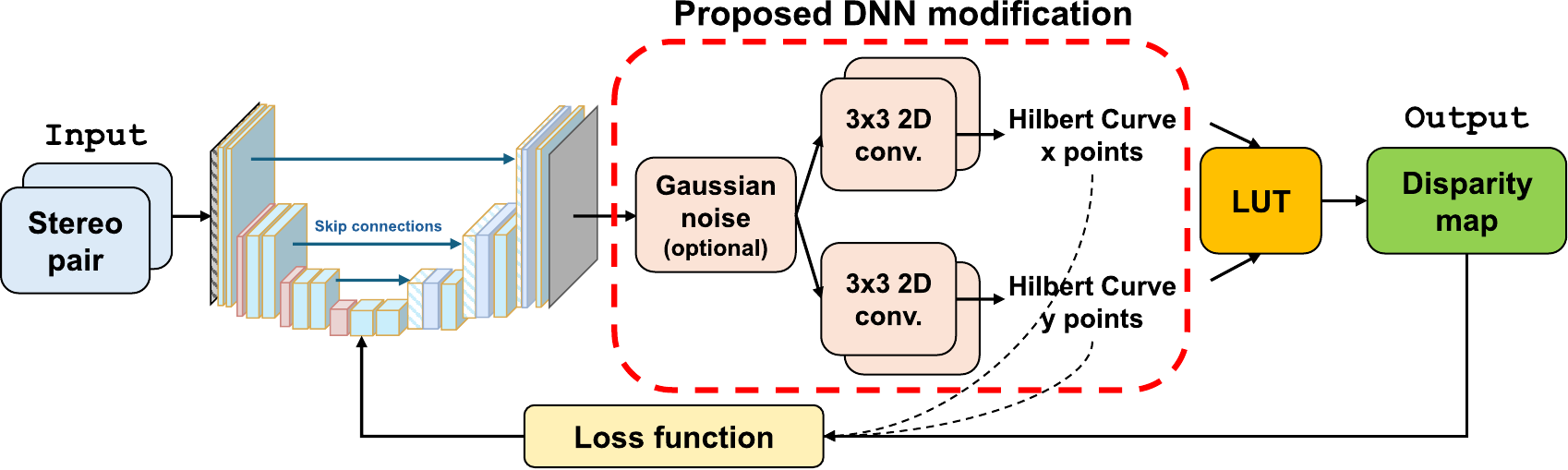}
        \caption{DispNet~\cite{DispNet} model modification required by the proposed approach. The input RGB stereo pair is processed by the encoder-decoder network from the original model. Features from the encoder-decoder are fed to the optional Gaussian noise layer, followed by two identical heads for Hilbert curve components. At the post-processing stage, Hilbert components are converted to the final disparity map.}
        \label{fig:3-3-hpdispnet-arch}
        \vspace*{-0.3cm}
\end{figure}

To implement the proposed approach, a DNN with one head predicting quantity $q$ should be modified to have two heads predicting Hilbert curve components $x$ and $y$. This modification for the DispNet model~\cite{DispNet} is illustrated in Fig.~\ref{fig:3-3-hpdispnet-arch}.

The loss function for the proposed method is composed of two terms: depth loss $\Lambda(q_{\scriptscriptstyle\textrm{GT}}, q_{xy})$~\cite{HuLoss} and additional term $\Lambda_H(x_{\scriptscriptstyle\textrm{GT}}, y_{\scriptscriptstyle\textrm{GT}}, x, y)$ that assures model convergence to the Hilbert curve-based representation:
\begin{equation}
\Lambda_{full} = \Lambda(q_{\scriptscriptstyle\textrm{GT}}, q_{xy}) + \alpha \cdot \Lambda_H(x_{\scriptscriptstyle\textrm{GT}}, y_{\scriptscriptstyle\textrm{GT}}, x, y),
    \label{eq:joint_loss}
\end{equation}
\noindent where $x_{\scriptscriptstyle\textrm{GT}}$ and $y_{\scriptscriptstyle\textrm{GT}}$ are GT values for Hilbert curve components calculated from GT value $q_{\scriptscriptstyle\textrm{GT}}$, and $\alpha$ is hyperparameter. The additional component $\Lambda_H$ is calculated as follows:
\begin{equation}
\Lambda_H(x_{\scriptscriptstyle\textrm{GT}}, y_{\scriptscriptstyle\textrm{GT}}, x, y) = (x_{\scriptscriptstyle\textrm{GT}} - x)^2 + (y_{\scriptscriptstyle\textrm{GT}} - y)^2 + \beta \cdot r_{xy}^2,
    \label{eq:HILBERT_loss}
\end{equation}
\noindent where $r_{xy} = \lVert (x - x(q_{xy}), y - y(q_{xy}) \rVert$ and $\beta$ is additional hyperparameter. The Hilbert curve loss component serves two goals: it penalizes the distance between GT and predicted points in the 2D representation; it penalizes deviation across the Hilbert curve and forces the model to predict only points that belong to the curve. We set $\alpha=1$ and $\beta=25$ in our experiments.

\section{Experiments}
\label{sec:experiment}

Two models are selected for the stereo matching experiment: DispNet with the original architecture proposed in~\cite{DispNet} and Dense Prediction Transformer (DPT)~\cite{DenseTransformer} with MobileViTv3-S~\cite{MobileViTv3} as an encoder. In all experiments, the models' input size is $384\times512$ pixels and the output size is $192\times256$ pixels. All modifications of the DPT model are described in detail in Appendix~\ref{sec:DPT}. The head predicting Hilbert curve components for the DPT model is the same as for the DispNet but includes an additional convolutional and up-sample layer in each branch to adapt to feature shapes of the DPT decoder. We found that injecting small amount of Gaussian noise at the beginning of the Hilbert components head improves quantization of modified models with the SNPE library and has no effect on unmodified models. Experimental results for modified models are presented with a Gaussian noise layer with SD equals 0.02.

\subsection{Implementation Details}

For stereo matching models training we adapted ScanNet~v2~\cite{ScanNet} dataset in the following way. Training, validation, and test data are rendered from meshes provided for each ScanNet~v2 scene using PyRender~v.0.1.45 library. The camera poses for the left camera are fixed to the same values as specified in ScanNet~v2. The right camera is shifted by 60~mm along the axis $x$ to form the horizontal baseline. Intrinsic parameters for the left and right cameras correspond to ScanNet data: pinhole camera with $f_x=f_y=577.87$, $c_x=320$, $c_y=240$. The split of the dataset into training and test parts corresponds to the official ScanNet~v2 split.

All models were quantized using SNPE SDK~v.2.24 and tested on Samsung~S24+ device with Qualcomm Snapdragon~8 Gen~3 processor and Hexagon DSP. We compare models in FP16, W8A16 and W8A8 formats running on Hexagon DSP. Power consumption is measured with Monsoon Solutions FTA22D Power Monitor in power save mode.

\subsection{Evaluation Metrics}

We characterize the quality of predicted depth using standard metrics~\cite{DepthMP, geiger2012we}: mean absolute relative error (Abs Rel), root mean square error (RMSE), end-point-error (EPE), and D1.

Pixel-level errors themselves are not sufficient to characterize quantization artifacts in the predicted depth maps. For example, INT8 depth representation has minimal impact on the Abs Rel metric. To address this issue, we experimented with the SSIM metric~\cite{SSIM, UnderstandingSSIM}. However, we found it barely affected by quantization artifacts. Therefore, we propose using cosine similarity~\cite{CosSimilarity} between discrete cosine transform (DCT)~\cite{DCT} coefficients of GT and predicted depth maps. For this, $n\times n$ DCT is applied in a scanning window manner to both GT and predicted depth maps. The DCT coefficients are flattened to vector representations and zero coefficients are discarded. Cosine similarity between flattened vectors is calculated at each scanning window position and then averaged:
\begin{equation}
  S_{C} = \frac{1}{N}\sum_{i=1}^{N}\sum_{j=1}^{M}\frac{\mathbf{c_{ij}} \cdot \mathbf{\hat{c_{ij}}}}{\|\mathbf{c_{ij}}\| \cdot \|\mathbf{\hat{c_{ij}}}\|},
  \label{eq:cos_sim}
\end{equation}
where $N$ is the number of frames, M is the number of scanning windows, $\mathbf{c_{ij}}$ and $\mathbf{\hat{c}_{ij}}$ are vectors of DCT coefficients of GT and predicted depth maps for frame ${i}$ and window $j$. Experiments show that $S_{C}$ calculated in $4\times 4$ window is sensitive to depth blurring and INT8 quantization artifacts. The value of $S_{C}$ close to unity (the maximum possible value) indicates high-quality depth maps with sharp edges and absence of artifacts in homogeneous areas.

To characterize quantization error, we use SD $\hat{\sigma}$ between raw outputs of full-precision and quantized models measured in a robust way as Scaled Median Absolute Deviation~\cite{ScaledMAD}.

\subsection{Analysis of W8A8 Models}

We trained DispNet and DPT models using $p=1, 2, 3, 4$. These models will be later referred to as hpDispNet and hpDPT, respectively. We observed that float-precision models hpDispNet and hpDPT learn to predict points close to the Hilbert curve. Depth prediction accuracy of modified models is similar to the original models w.r.t. all metrics.

Next, we compare the original and modified models in W8A8 format running on CPU and DSP. The difference is that W8A8 model runs on CPU in dequantization mode using full precision arithmetic, while on DSP inference is done in lower-precision arithmetic. Quantitative results are presented in Table~\ref{tab:3-5-keras-dlc-metrics}. For the original DispNet model, quantization leads to noticeable quality degradation on both CPU and DSP. On CPU degradation is seen for $S_c$ metric that drops from 0.86 to 0.68 reflecting loss of spatial details.

\begin{figure}[!ht]
    \centering
    \begin{subfigure}{0.95\linewidth}
        \includegraphics[width=\linewidth]{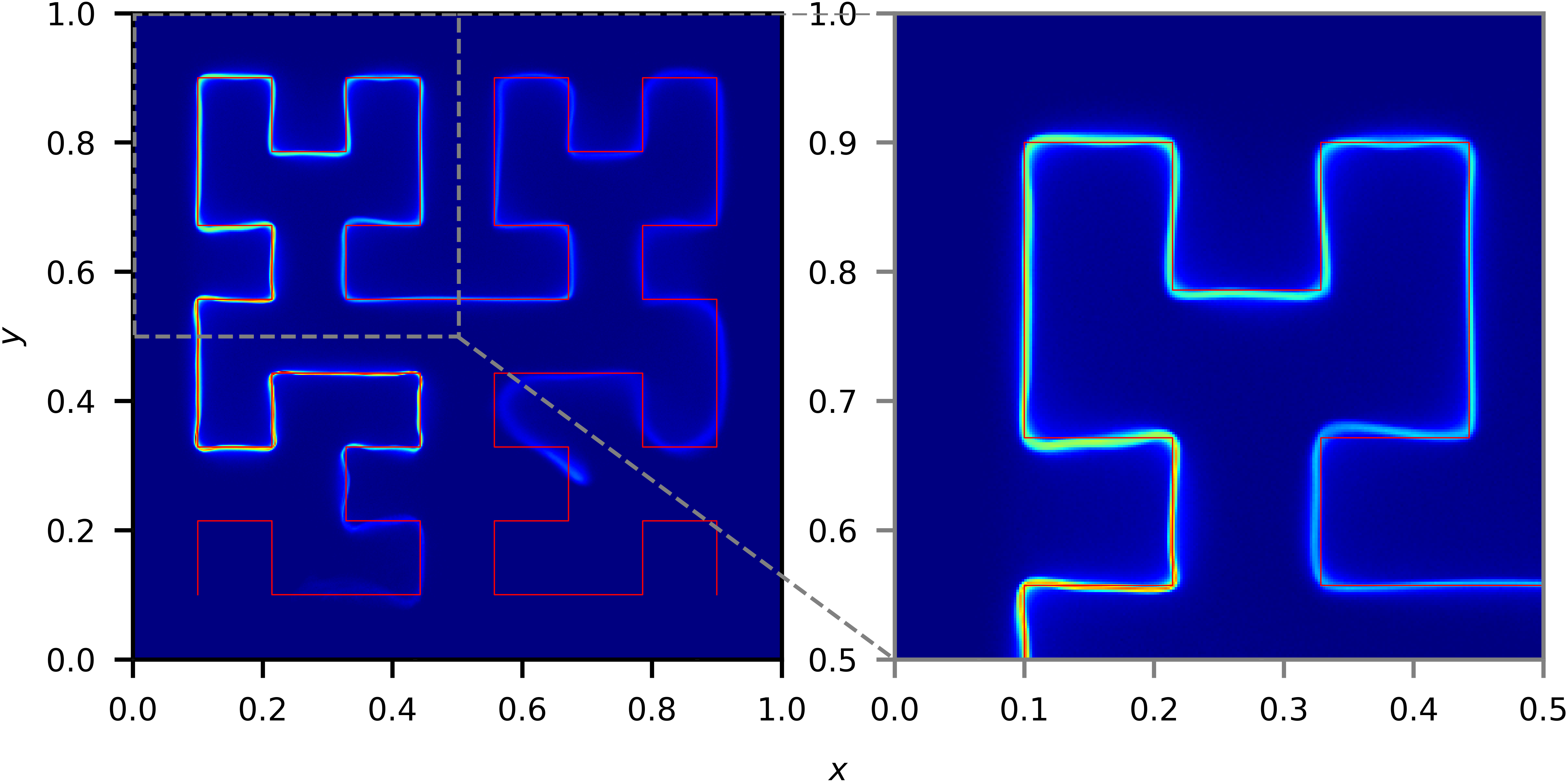}
        \caption{W8A8 model on CPU delegate}
        \label{fig:hilbert3-keras-dn}
    \end{subfigure}
    \begin{subfigure}{0.95\linewidth}
        \includegraphics[width=\linewidth]{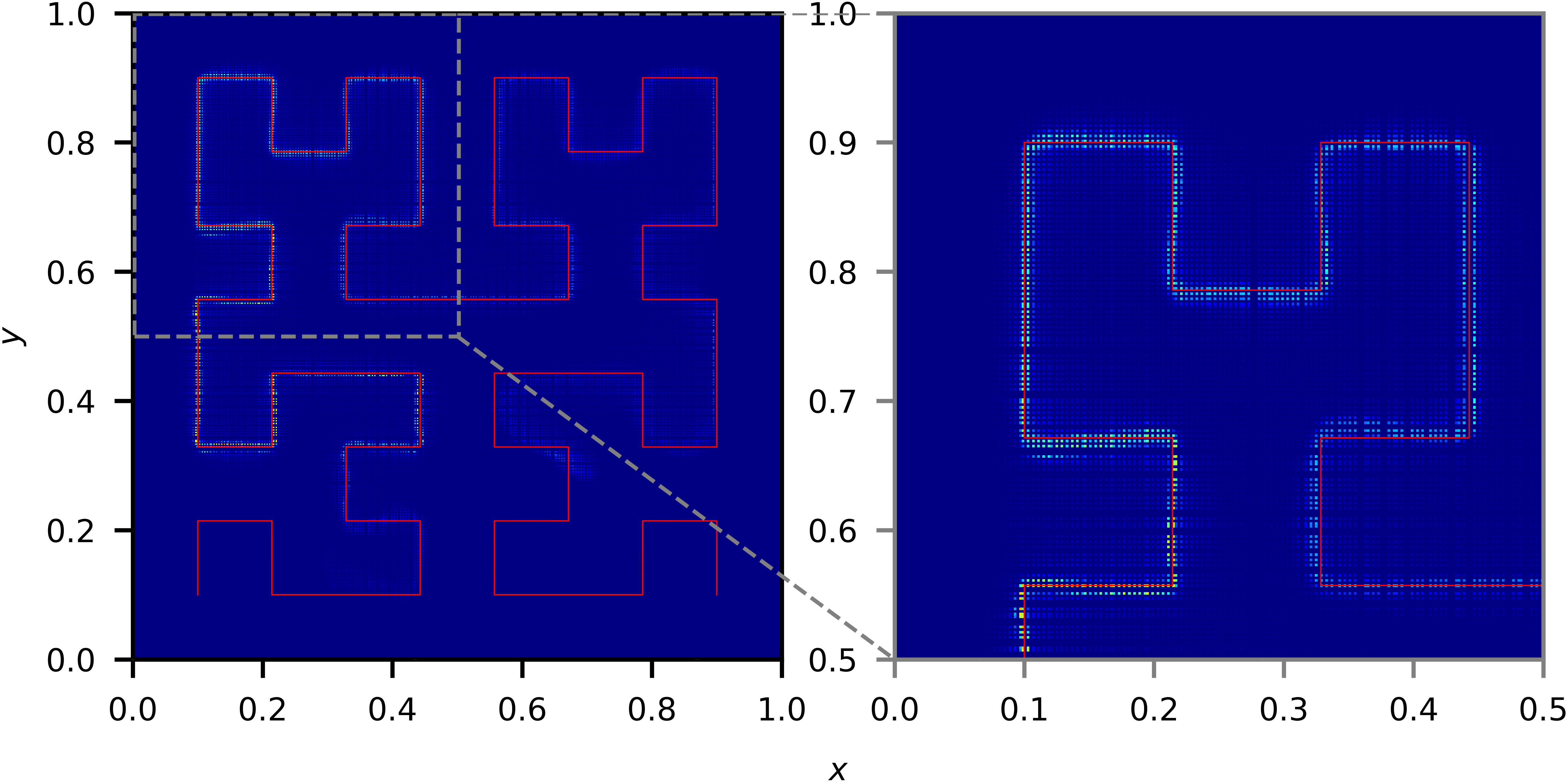}
        \caption{W8A8 model on DSP delegate}
        \label{fig:hilbert3-dsp-dn}
    \end{subfigure}
    \caption{2D histogram of h3DispNet W8A8 model output for CPU and DSP delegates.}
    \label{fig:hilbert3-dn-comp}
    \vspace*{-0.3cm}
\end{figure}

Modified models for all $p$ perform better than the original ones; the best result achieved for $p=3$. As shown in Fig.~\ref{fig:hilbert3-dn-comp}, the quantized h3DispNet model retains the ability to predict points on the Hilbert curve for both CPU and DSP inference. The h3DispNet model on CPU shows almost the same quality as the FP32 model and outperforms the original DispNet w.r.t. all metrics. On DSP, the quality drop of the original DispNet is more significant: Abs Rel increases from 1.01 to 2.03, and $S_c$ decreases from 0.86 to 0.58. The h3DispNet model compensates this drop almost completely reaching Abs Rel 0.93 and $S_c$ 0.81.

\begin{table*}[]
  \centering
 \begin{scriptsize} 
\begin{tabular}{c|ccc|ccc|ccc|ccc|ccc}
    \toprule
\multicolumn{1}{c@{\kern0.2em}|}{\multirow{3}{*}{Model}}
& \multicolumn{3}{c@{\kern0.2em}|}{{Abs Rel, \% $\downarrow$}}                              & \multicolumn{3}{c@{\kern0.2em}|}{RMSE, px$\downarrow$} 
& \multicolumn{3}{c@{\kern0.2em}|}{$S_{C}\uparrow$}
& \multicolumn{3}{c|}{EPE, px $\downarrow$}   & \multicolumn{3}{c}{D1, \% $\downarrow$}     \\ \cline{2-16}

\multicolumn{1}{c|}{}
& \multicolumn{1}{c@{\kern0.2em}}{FP32}  
& \multicolumn{1}{c@{}}{W8A8}
& \multicolumn{1}{c@{\kern0.4em}|}{W8A8}
& \multicolumn{1}{c@{\kern0.2em}}{FP32}  
& \multicolumn{1}{c@{}}{W8A8}
& \multicolumn{1}{c@{\kern0.4em}|}{W8A8}
& \multicolumn{1}{c@{\kern0.2em}}{FP32} 
& \multicolumn{1}{c@{}}{W8A8}
& \multicolumn{1}{c@{\kern0.4em}|}{W8A8}
& \multicolumn{1}{c@{\kern0.2em}}{FP32} 
& \multicolumn{1}{c@{}}{W8A8}
& \multicolumn{1}{c@{\kern0.4em}|}{W8A8}
& \multicolumn{1}{c@{\kern0.2em}}{FP32} 
& \multicolumn{1}{c@{}}{W8A8}
& \multicolumn{1}{c@{\kern0.4em}}{W8A8\vphantom{\rule{0pt}{1.1em}}}
\\
\multicolumn{1}{c@{\kern0.2em}|}{} 
& \multicolumn{1}{c@{\kern0.2em}}{CPU}  
& \multicolumn{1}{c@{\kern0.2em}}{DSP}
& \multicolumn{1}{c@{\kern0.2em}|}{CPU}
& \multicolumn{1}{c@{\kern0.2em}}{CPU}  
& \multicolumn{1}{c@{\kern0.2em}}{DSP}
& \multicolumn{1}{c@{\kern0.2em}|}{CPU}
& \multicolumn{1}{c@{\kern0.2em}}{CPU} 
& \multicolumn{1}{c@{\kern0.2em}}{DSP}
& \multicolumn{1}{c@{\kern0.2em}|}{CPU}
& \multicolumn{1}{c@{\kern0.2em}}{CPU} 
& \multicolumn{1}{c@{\kern0.2em}}{DSP}
& \multicolumn{1}{c@{\kern0.2em}|}{CPU}
& \multicolumn{1}{c@{\kern0.2em}}{CPU} 
& \multicolumn{1}{c@{\kern0.2em}}{DSP}
& \multicolumn{1}{c@{\kern0.2em}}{CPU\vphantom{$A_p$}} 
\\ \hline
\rowcolor[HTML]{EFEFEF} 
DispNet
&\multicolumn{1}{c@{\kern0.2em}}{1.01\vphantom{$A^h$}} 
&\multicolumn{1}{c@{\kern0.2em}}{2.03}
&\multicolumn{1}{c@{\kern0.2em}|}{1.15}
&\multicolumn{1}{c@{\kern0.2em}}{1.12} 
&\multicolumn{1}{c@{\kern0.2em}}{2.24}
&\multicolumn{1}{c@{\kern0.2em}|}{1.10}
&\multicolumn{1}{c@{\kern0.2em}}{0.86} 
&\multicolumn{1}{c@{\kern0.2em}}{0.58}
&\multicolumn{1}{c@{\kern0.2em}|}{0.68}
&\multicolumn{1}{c@{\kern0.2em}}{0.29} 
&\multicolumn{1}{c@{\kern0.2em}}{0.69}
&\multicolumn{1}{c@{\kern0.2em}|}{0.31}
&\multicolumn{1}{c@{\kern0.2em}}{1.81} 
&\multicolumn{1}{c@{\kern0.2em}}{5.35}
&\multicolumn{1}{c@{\kern0.2em}}{1.73}\\
h1DispNet 
&\multicolumn{1}{c@{\kern0.2em}}{1.06} 
&\multicolumn{1}{c@{\kern0.2em}}{1.50}
&\multicolumn{1}{c@{\kern0.2em}|}{1.12}
&\multicolumn{1}{c@{\kern0.2em}}{0.97} 
&\multicolumn{1}{c@{\kern0.2em}}{1.09}
&\multicolumn{1}{c@{\kern0.2em}|}{0.97}
&\multicolumn{1}{c@{\kern0.2em}}{0.86} 
&\multicolumn{1}{c@{\kern0.2em}}{0.67}
&\multicolumn{1}{c@{\kern0.2em}|}{0.79}
&\multicolumn{1}{c@{\kern0.2em}}{0.27} 
&\multicolumn{1}{c@{\kern0.2em}}{0.35}
&\multicolumn{1}{c@{\kern0.2em}|}{0.29}
&\multicolumn{1}{c@{\kern0.2em}}{1.27} 
&\multicolumn{1}{c@{\kern0.2em}}{2.25}
&\multicolumn{1}{c@{\kern0.2em}}{1.27}\\
h2DispNet 
&\multicolumn{1}{c@{\kern0.2em}}{0.85} 
&\multicolumn{1}{c@{\kern0.2em}}{0.98}
&\multicolumn{1}{c@{\kern0.2em}|}{0.88}
&\multicolumn{1}{c@{\kern0.2em}}{0.90} 
&\multicolumn{1}{c@{\kern0.2em}}{\textbf{0.94}}
&\multicolumn{1}{c@{\kern0.2em}|}{0.91}
&\multicolumn{1}{c@{\kern0.2em}}{0.87} 
&\multicolumn{1}{c@{\kern0.2em}}{0.75}
&\multicolumn{1}{c@{\kern0.2em}|}{0.83}
&\multicolumn{1}{c@{\kern0.2em}}{0.22} 
&\multicolumn{1}{c@{\kern0.2em}}{0.25}
&\multicolumn{1}{c@{\kern0.2em}|}{0.23}
&\multicolumn{1}{c@{\kern0.2em}}{1.01} 
&\multicolumn{1}{c@{\kern0.2em}}{1.26}
&\multicolumn{1}{c@{\kern0.2em}}{1.02}\\
h3DispNet
&\multicolumn{1}{c@{\kern0.2em}}{0.88} 
&\multicolumn{1}{c@{\kern0.2em}}{\textbf{0.93}}
&\multicolumn{1}{c@{\kern0.2em}|}{0.87}
&\multicolumn{1}{c@{\kern0.2em}}{1.00} 
&\multicolumn{1}{c@{\kern0.2em}}{1.03}
&\multicolumn{1}{c@{\kern0.2em}|}{1.00}
&\multicolumn{1}{c@{\kern0.2em}}{0.87} 
&\multicolumn{1}{c@{\kern0.2em}}{0.81}
&\multicolumn{1}{c@{\kern0.2em}|}{0.86}
&\multicolumn{1}{c@{\kern0.2em}}{0.24} 
&\multicolumn{1}{c@{\kern0.2em}}{\textbf{0.24}}
&\multicolumn{1}{c@{\kern0.2em}|}{0.24}
&\multicolumn{1}{c@{\kern0.2em}}{1.25} 
&\multicolumn{1}{c@{\kern0.2em}}{1.26}
&\multicolumn{1}{c@{\kern0.2em}}{1.25}\\
h4DispNet
&\multicolumn{1}{c@{\kern0.2em}}{0.90} 
&\multicolumn{1}{c@{\kern0.2em}}{0.94}
&\multicolumn{1}{c@{\kern0.2em}|}{0.92}
&\multicolumn{1}{c@{\kern0.2em}}{1.02} 
&\multicolumn{1}{c@{\kern0.2em}}{1.02}
&\multicolumn{1}{c@{\kern0.2em}|}{1.02}
&\multicolumn{1}{c@{\kern0.2em}}{0.85} 
&\multicolumn{1}{c@{\kern0.2em}}{\textbf{0.83}}
&\multicolumn{1}{c@{\kern0.2em}|}{0.85}
&\multicolumn{1}{c@{\kern0.2em}}{0.24} 
&\multicolumn{1}{c@{\kern0.2em}}{0.25}
&\multicolumn{1}{c@{\kern0.2em}|}{0.24}
&\multicolumn{1}{c@{\kern0.2em}}{1.24} 
&\multicolumn{1}{c@{\kern0.2em}}{\textbf{1.25}}
&\multicolumn{1}{c@{\kern0.2em}}{1.24}\\ \hline
\rowcolor[HTML]{EFEFEF} 
DPT 
&\multicolumn{1}{c@{\kern0.2em}}{0.75\vphantom{$A^h$}} 
&\multicolumn{1}{c@{\kern0.2em}}{4.18}
&\multicolumn{1}{c@{\kern0.2em}|}{1.48}
&\multicolumn{1}{c@{\kern0.2em}}{0.87} 
&\multicolumn{1}{c@{\kern0.2em}}{2.09}
&\multicolumn{1}{c@{\kern0.2em}|}{1.03}
&\multicolumn{1}{c@{\kern0.2em}}{0.89} 
&\multicolumn{1}{c@{\kern0.2em}}{0.52}
&\multicolumn{1}{c@{\kern0.2em}|}{0.87}
&\multicolumn{1}{c@{\kern0.2em}}{0.21} 
&\multicolumn{1}{c@{\kern0.2em}}{1.03}
&\multicolumn{1}{c@{\kern0.2em}|}{0.39}
&\multicolumn{1}{c@{\kern0.2em}}{0.95} 
&\multicolumn{1}{c@{\kern0.2em}}{5.78}
&\multicolumn{1}{c@{\kern0.2em}}{1.39}\\
h1DPT 
&\multicolumn{1}{c@{\kern0.2em}}{0.70} 
&\multicolumn{1}{c@{\kern0.2em}}{1.48}
&\multicolumn{1}{c@{\kern0.2em}|}{0.78}
&\multicolumn{1}{c@{\kern0.2em}}{0.88} 
&\multicolumn{1}{c@{\kern0.2em}}{1.41}
&\multicolumn{1}{c@{\kern0.2em}|}{0.89}
&\multicolumn{1}{c@{\kern0.2em}}{0.88} 
&\multicolumn{1}{c@{\kern0.2em}}{0.54}
&\multicolumn{1}{c@{\kern0.2em}|}{0.88}
&\multicolumn{1}{c@{\kern0.2em}}{0.20} 
&\multicolumn{1}{c@{\kern0.2em}}{0.41}
&\multicolumn{1}{c@{\kern0.2em}|}{0.22}
&\multicolumn{1}{c@{\kern0.2em}}{1.01} 
&\multicolumn{1}{c@{\kern0.2em}}{2.77}
&\multicolumn{1}{c@{\kern0.2em}}{1.03}\\
h2DPT
&\multicolumn{1}{c@{\kern0.2em}}{0.71} 
&\multicolumn{1}{c@{\kern0.2em}}{\textbf{1.12}}
&\multicolumn{1}{c@{\kern0.2em}|}{0.72}
&\multicolumn{1}{c@{\kern0.2em}}{0.91} 
&\multicolumn{1}{c@{\kern0.2em}}{\textbf{1.02}}
&\multicolumn{1}{c@{\kern0.2em}|}{0.91}
&\multicolumn{1}{c@{\kern0.2em}}{0.88} 
&\multicolumn{1}{c@{\kern0.2em}}{0.62}
&\multicolumn{1}{c@{\kern0.2em}|}{0.88}
&\multicolumn{1}{c@{\kern0.2em}}{0.20} 
&\multicolumn{1}{c@{\kern0.2em}}{\textbf{0.29}}
&\multicolumn{1}{c@{\kern0.2em}|}{0.20}
&\multicolumn{1}{c@{\kern0.2em}}{1.07} 
&\multicolumn{1}{c@{\kern0.2em}}{\textbf{1.27}}
&\multicolumn{1}{c@{\kern0.2em}}{1.08}\\
h3DPT
&\multicolumn{1}{c@{\kern0.2em}}{0.55} 
&\multicolumn{1}{c@{\kern0.2em}}{1.35}
&\multicolumn{1}{c@{\kern0.2em}|}{0.63}
&\multicolumn{1}{c@{\kern0.2em}}{0.80} 
&\multicolumn{1}{c@{\kern0.2em}}{1.33}
&\multicolumn{1}{c@{\kern0.2em}|}{0.80}
&\multicolumn{1}{c@{\kern0.2em}}{0.90} 
&\multicolumn{1}{c@{\kern0.2em}}{0.70}
&\multicolumn{1}{c@{\kern0.2em}|}{0.90}
&\multicolumn{1}{c@{\kern0.2em}}{0.15} 
&\multicolumn{1}{c@{\kern0.2em}}{0.32}
&\multicolumn{1}{c@{\kern0.2em}|}{0.17}
&\multicolumn{1}{c@{\kern0.2em}}{0.78} 
&\multicolumn{1}{c@{\kern0.2em}}{1.28}
&\multicolumn{1}{c@{\kern0.2em}}{0.78}\\
h4DPT
&\multicolumn{1}{c@{\kern0.2em}}{0.74} 
&\multicolumn{1}{c@{\kern0.2em}}{1.27}
&\multicolumn{1}{c@{\kern0.2em}|}{0.76}
&\multicolumn{1}{c@{\kern0.2em}}{0.94} 
&\multicolumn{1}{c@{\kern0.2em}}{1.38}
&\multicolumn{1}{c@{\kern0.2em}|}{0.94}
&\multicolumn{1}{c@{\kern0.2em}}{0.87} 
&\multicolumn{1}{c@{\kern0.2em}}{\textbf{0.73}}
&\multicolumn{1}{c@{\kern0.2em}|}{0.86}
&\multicolumn{1}{c@{\kern0.2em}}{0.21} 
&\multicolumn{1}{c@{\kern0.2em}}{0.32}
&\multicolumn{1}{c@{\kern0.2em}|}{0.21}
&\multicolumn{1}{c@{\kern0.2em}}{1.14} 
&\multicolumn{1}{c@{\kern0.2em}}{1.62}
&\multicolumn{1}{c@{\kern0.2em}}{1.14}  \\             \bottomrule             
\end{tabular}
    \end{scriptsize}
    \caption{Metrics of DispNet, hpDispNet, DPT and hpDPT models. All metrics are presented for FP32 model (CPU inference), W8A8 model running on DSP and W8A8 model running on CPU (dequantization mode). The best results on DSP are in bold font. Results for the original DispNet and DPT models are marked in grey color.}
    \label{tab:3-5-keras-dlc-metrics}
\end{table*}

For the DPT model, the situation is similar, but the quality drop for the quantized model is more significant on both CPU and DSP. On CPU, the h3DPT model performs better than the original FP32 DPT model. On DSP the best result shows h2DPT model with Abs Rel improving from 4.18\% to 1.12\% and $S_c$ increasing from 0.52 to 0.62 as compared to the original model. The curve of the third order provides the best compromise between Abs Rel and $S_c$ improvement for both DispNet and DPT models. This conclusion holds when RMSE, EPE or D1 are taken into account.

\begin{figure}
    \centering
    \begin{subfigure}{0.3\linewidth}
        \includegraphics[width=1.0\linewidth]{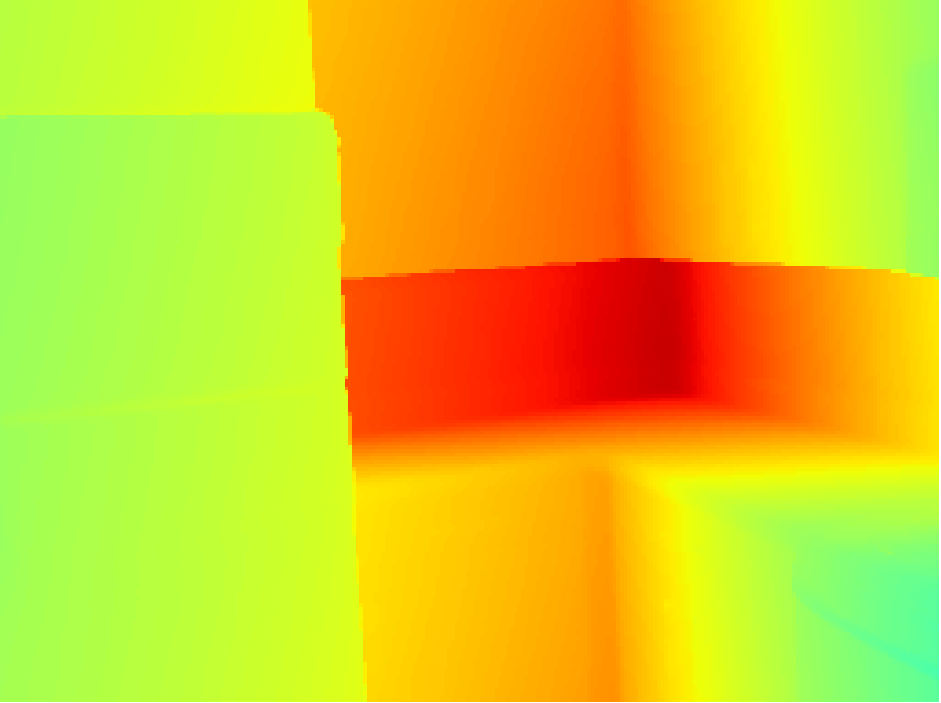}
        \caption{GT depth}
        \label{fig:4-5-gt_depth}
    \end{subfigure}
    \centering
    \begin{subfigure}{0.3\linewidth}
    \includegraphics[width=1.0\linewidth]{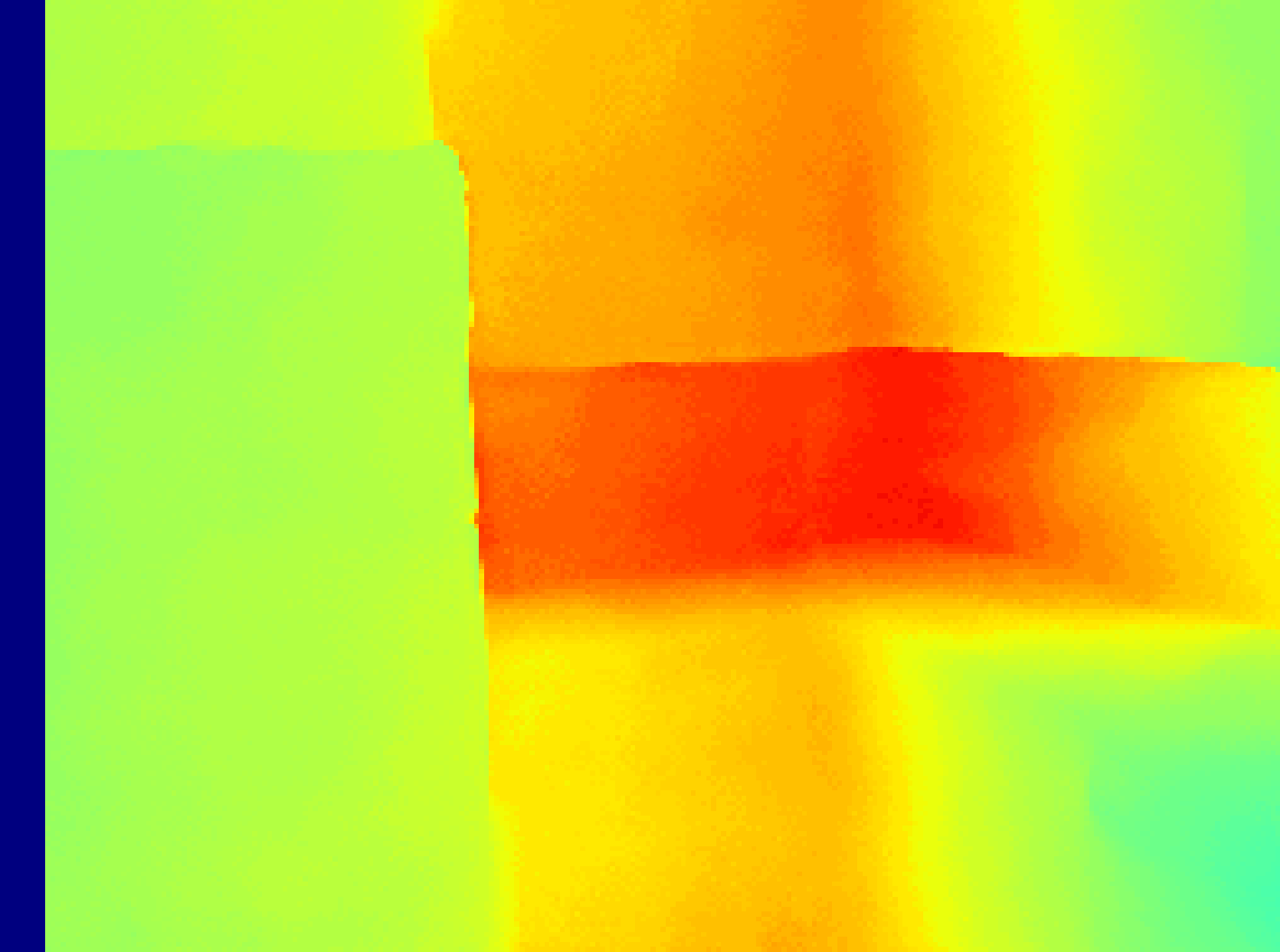}
        \caption{DPT depth}
        \label{fig:4-5-dptbase}
    \end{subfigure}
    \centering
    \begin{subfigure}{0.355\linewidth}
    \includegraphics[width=1.0\linewidth]{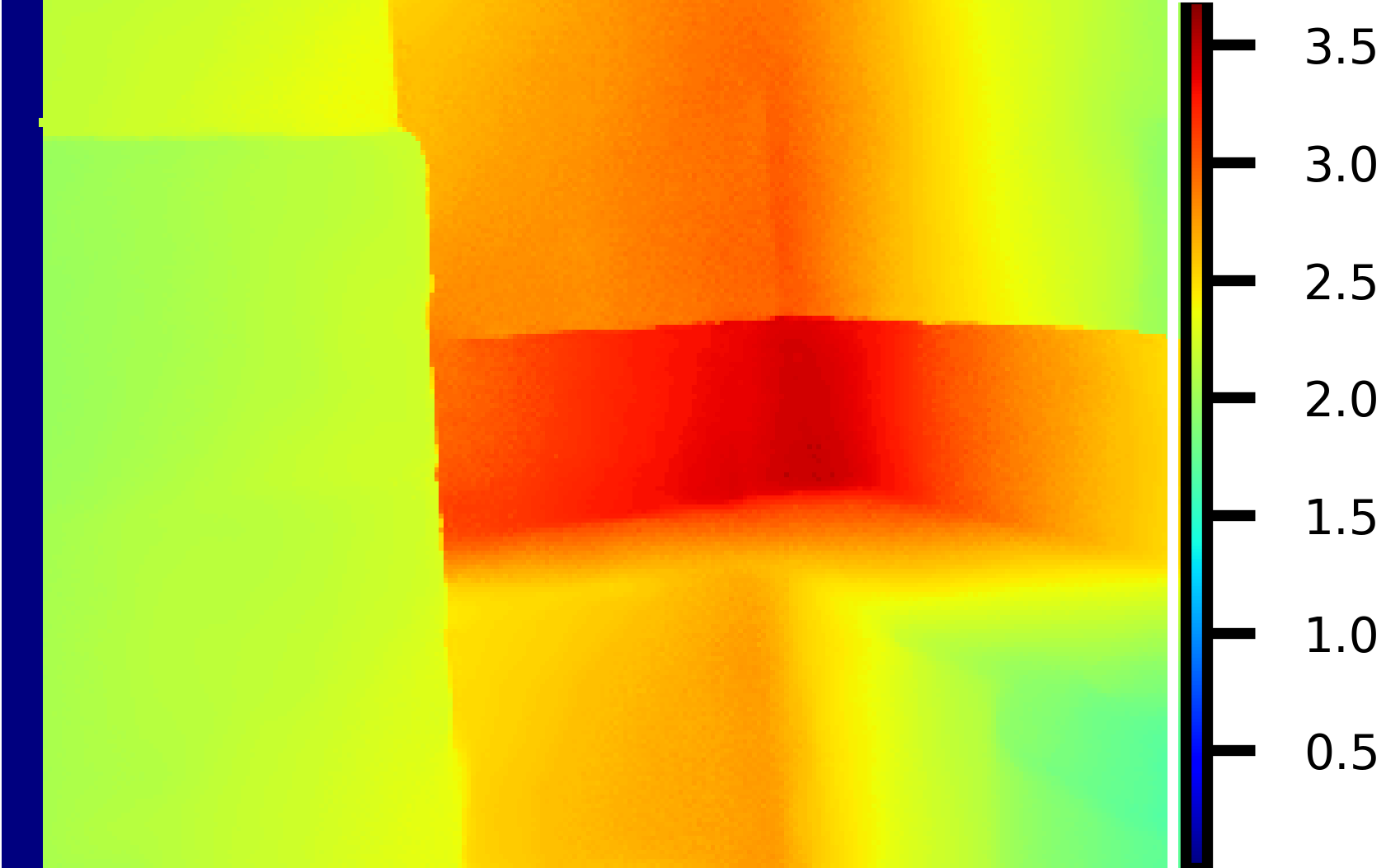}
        \caption{h3DPT depth$\phantom{xxx}$}
        \label{fig:4-5-h3dpt}
    \end{subfigure}
    \centering 
   \begin{subfigure}{0.30\linewidth}
        \includegraphics[width=1.0\linewidth]{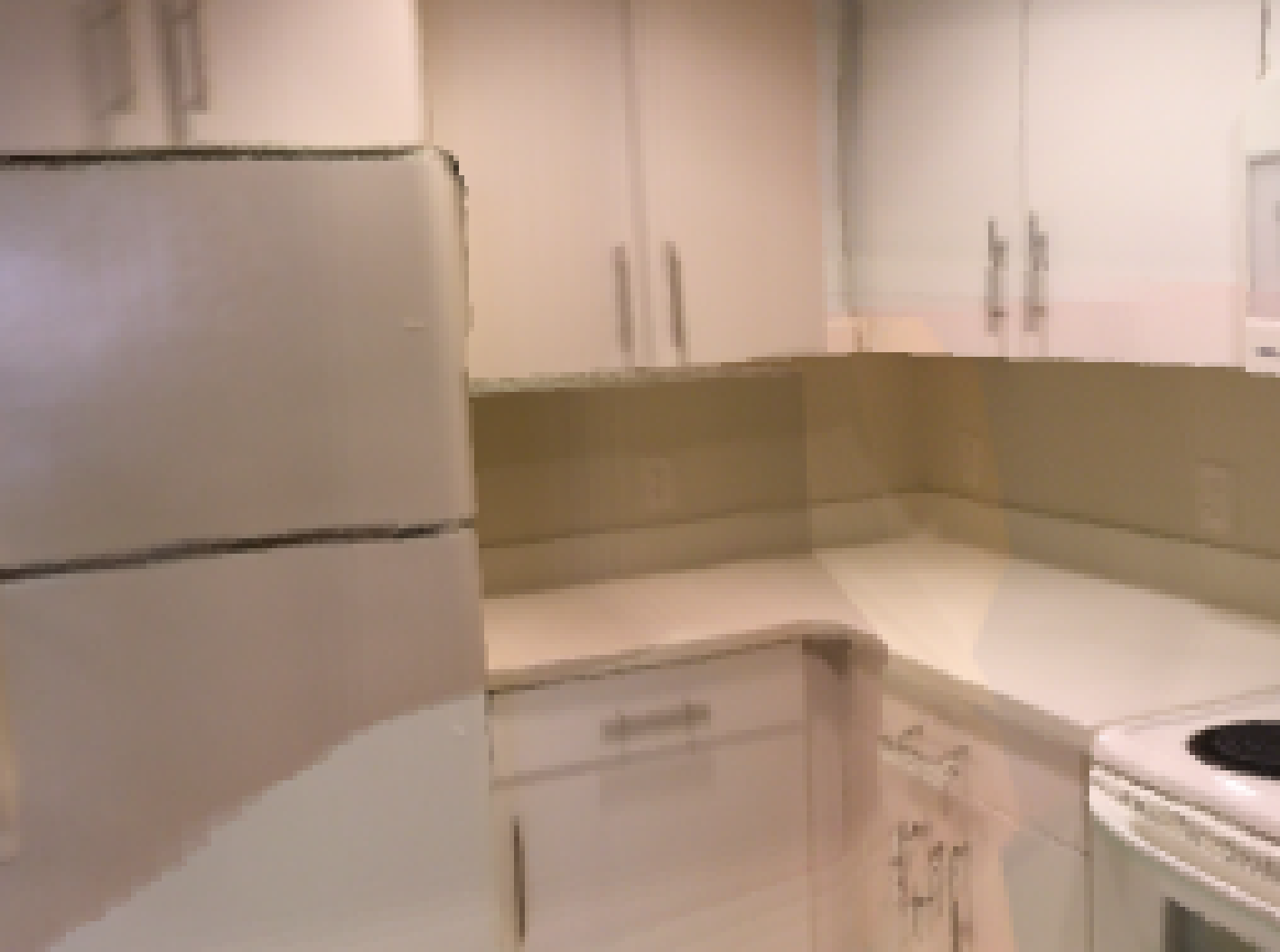}
        \caption{Image}
        \label{fig:4-5-gtdepth}
    \end{subfigure}
    \centering
    \begin{subfigure}{0.3\linewidth}
    \includegraphics[width=1.0\linewidth]{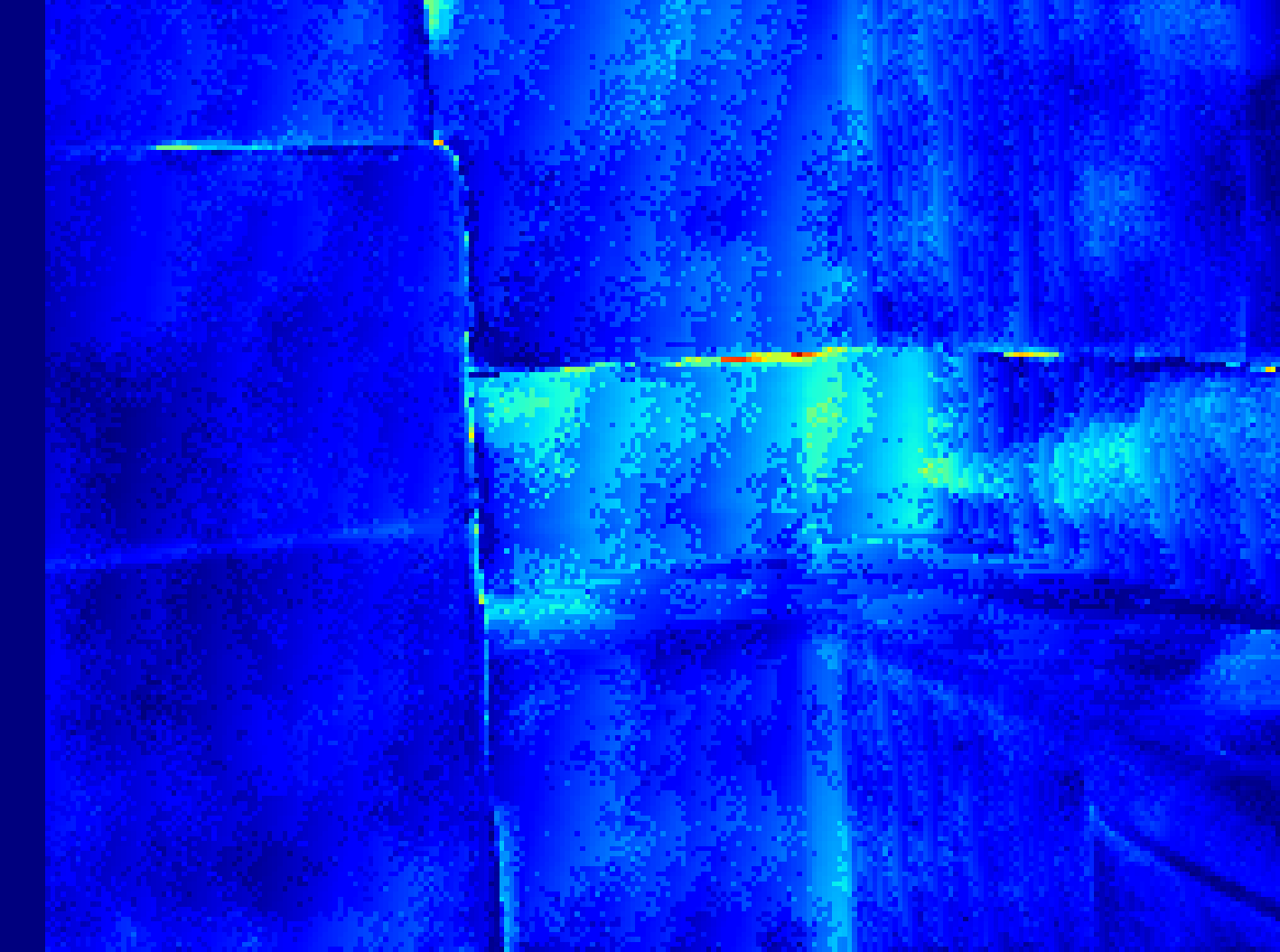}
    \caption{$|$GT - DPT$|$}
    \label{fig:4-5-gt_minus_dpt}
    \end{subfigure}
\centering
    \begin{subfigure}{0.355\linewidth}
    \includegraphics[width=1.0\linewidth]{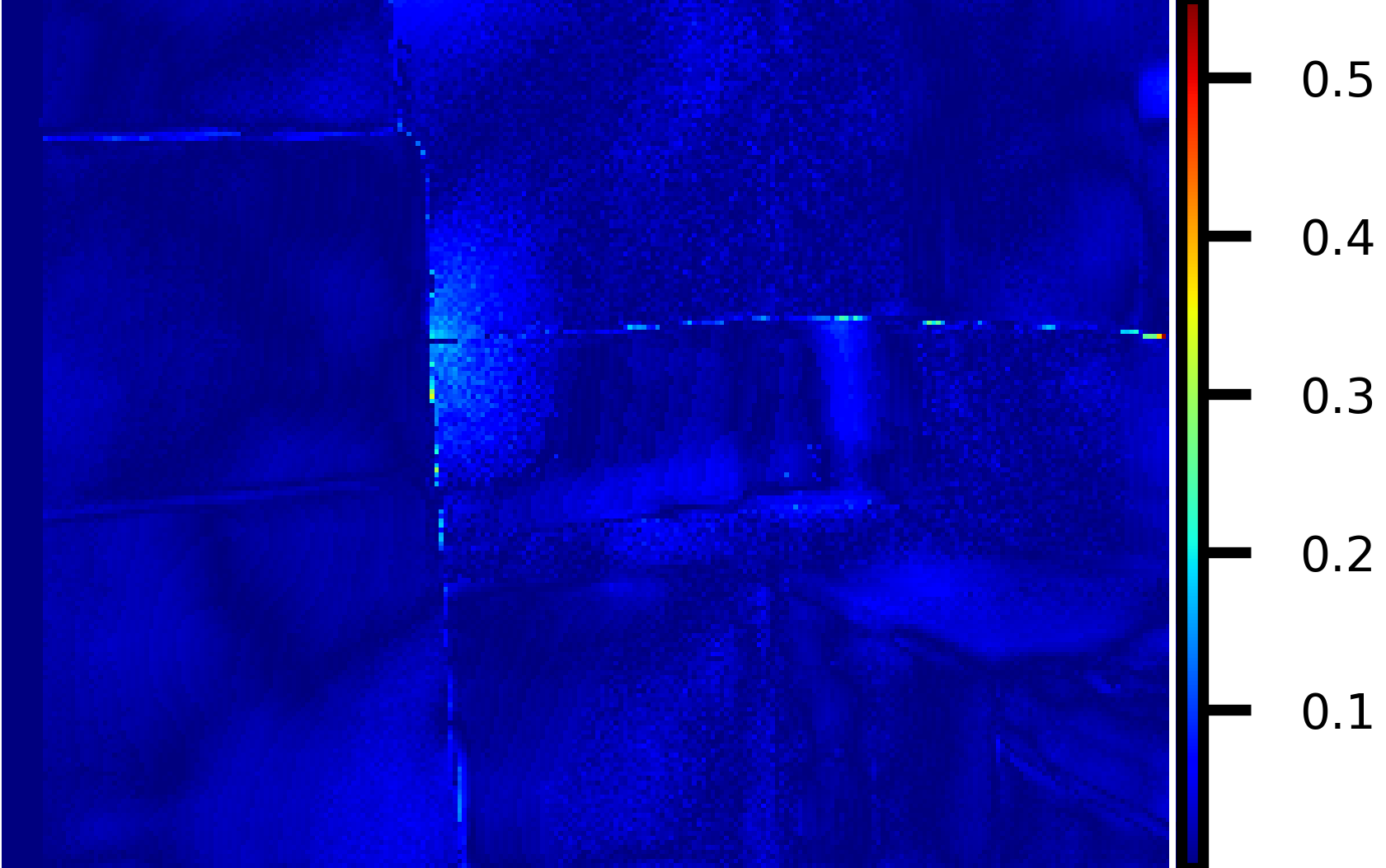}
    \caption{$|$GT - h3DPT$|$$\phantom{xxx}$}
    \label{fig:4-5-gt-h3dpt}
    \end{subfigure}
    \caption{Quantization errors of DPT and h3DPT, W8A8, DSP.}
    \label{fig:4-5-depth_errors}
\end{figure}

\begin{figure}
    \centering
    \begin{subfigure}{0.43\linewidth}
        \includegraphics[width=0.95\linewidth]{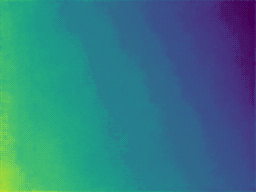}
        \caption{DispNet}
        \label{fig:DispNet_homo}
    \end{subfigure} \hfill
    \centering
    \begin{subfigure}{0.43\linewidth}
    \includegraphics[width=0.95\linewidth]{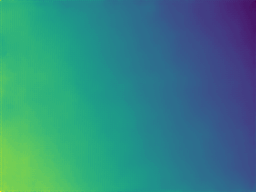}
        \caption{h2DispNet}
        \label{fig:h2DispNet_homo}
    \end{subfigure}
    \caption{Effect of increased bit-width on the quality of depth prediction for homogeneous areas on DSP.}
    \label{fig:depth_homo}
    \vspace*{-0.3cm}
\end{figure}

Qualitative results for the h3DPT model are illustrated in Fig.~\ref{fig:4-5-depth_errors}. Reduction of quantization error between the original model (Fig. \ref{fig:4-5-dptbase}) and the modified model (Fig.~\ref{fig:4-5-h3dpt}) is very significant as can be seen on error maps in Fig.~\ref{fig:4-5-gt_minus_dpt} and Fig.~\ref{fig:4-5-gt-h3dpt}.  We also observe from Fig.~\ref{fig:4-5-h3dpt} and Fig.~\ref{fig:4-5-dptbase} that h3DPT demonstrates better spatial details than the original model. This effect is caused by increase of effective number of bits for depth map coding by $\log_2 L$ bits (approximately by 2.85 bits for h3DPT resulting in INT10-INT11 precision). Indirectly, this effect is characterized by increase of $S_c$ value with $p$ (see Table \ref{tab:3-5-keras-dlc-metrics}). Visually, bit-precision increase is well seen for homogeneous planar areas as illustrated in Fig. \ref{fig:depth_homo}.

As one can see from these results, the proposed method not only solves the main problem of bit-width increase but also significantly reduces the level of quantization error of depth prediction. The latter is a positive side effect and we will study it more in detail in Section~\ref{sec:analysis_quat_reduction}. Effect of the improved quality of depth prediction on scene mesh fusion is discussed in Appendix~\ref{sec:mesh} and additional examples of predicted depth maps are given in Appendix~\ref{sec:depth}. Additional experiment on KITTI~2012 dataset is described in Appendix~\ref{sec:kitti}.

\subsection{Comparison of FP16, W8A16 and W8A8 Models}

Compared to the original model, the overhead of our method consists of three parts: (1) increased model complexity, (2) additional data transfer from DSP to CPU, (3) post-processing of Hilbert components. We found that this overhead ($\approx14\%$) is significantly smaller than the performance gain from using W8A8 models instead of W8A16 or FP16 models. Detailed profiling results are presented in Table \ref{tab:3-4-compare-fp16-w8a16-w8a8} for models with the curve order 3.

\begin{table}
\centering
\begin{scriptsize} 
\begin{tabular}{c @{\kern1em}c @{\kern1em}c @{\kern1em}c @{\kern1em}c @{\kern1em}c @{\kern1em}c@{\kern1em}}
    \toprule
    \rowcolor{white}
     Precision &  Abs Rel, \% &  EPE, px &  D1, \% &  $S_{C}$ & \textit{T}, 
 ms &  \textit{P}, {mW}$\cdot${s}/{infr.} \\
        \midrule
        \multicolumn{7}{c}{ \textbf{DispNet}} \\
        \midrule
    FP32 &  1.01 & 0.29 & 1.81 & 0.858  &  - & -  \\
    FP16 &  1.50 & 0.37 & 1.80 & 0.855  &  19.54 &  19.52 \\
    W8A16 & 1.78 & 0.63 & 5.22 & 0.798 & 18.7  & 12.3 \\
    \rowcolor{orange!20}
    W8A8 & 2.02 & 0.69 & 5.34 & 0.585 &  10.5 &  7.1 \\
    \rowcolor{green!20}
   Ours, W8A8 & 0.93 & 0.24 & 1.26 & 0.807 &  12.0 &  8.7 \\
 
        \midrule
        \rowcolor{white}
        \multicolumn{7}{c}{ \textbf{DPT}} \\
        \midrule
    FP32 & 0.75 & 0.21 & 0.95 & 0.889 & - & - \\ 
    FP16 & 1.14 & 0.27 & 0.97 & 0.884 & 54.1 & 110.5\\ 
    W8A16 & 4.03 & 0.97 & 5.58 & 0.825 & 46.2 & 64.5\\ 
    \rowcolor{orange!20}
    W8A8 & 4.16 & 1.03 & 5.76 & 0.520 & 26.7 & 28.3 \\
    \rowcolor{green!20}
   Ours, W8A8 & 1.35 & 0.32 & 1.28 & 0.697 & 30.4 & 29.7 \\ 

      \bottomrule
    \end{tabular}
    \end{scriptsize}
    \caption{On-device performance of the original and modified models including runtime (\textit{T}) and power consumption (\textit{P}). Measurements for our method include data transfer from DSP to CPU and Hilbert components post-processing on CPU. The overhead of our method is in runtime and power consumption increase between modified and original W8A8 models (compare lines marked by orange and green colors). Models in FP32 format are run on CPU; models in FP16, W8A16, W8A8 formats are run on DSP.}
    \label{tab:3-4-compare-fp16-w8a16-w8a8}
\end{table}

%
%
%

The modified DispNet model in W8A8 format shows better quality than the original model in W8A16 format while simultaneously reducing energy consumption by 35\% and latency by 30\%. Compared to the original model in W8A16 format, the modified DPT model in W8A8 format shows significantly better Abs Rel metric and only slightly worse $S_c$. It also lowers energy demands by 34\% and processing time by 54\%.

\subsection{Analysis of Quantization Errors Reduction}
\label{sec:analysis_quat_reduction}

\begin{figure}[t]
    \begin{subfigure}[t]{0.47\linewidth}
        \includegraphics[width=\linewidth]{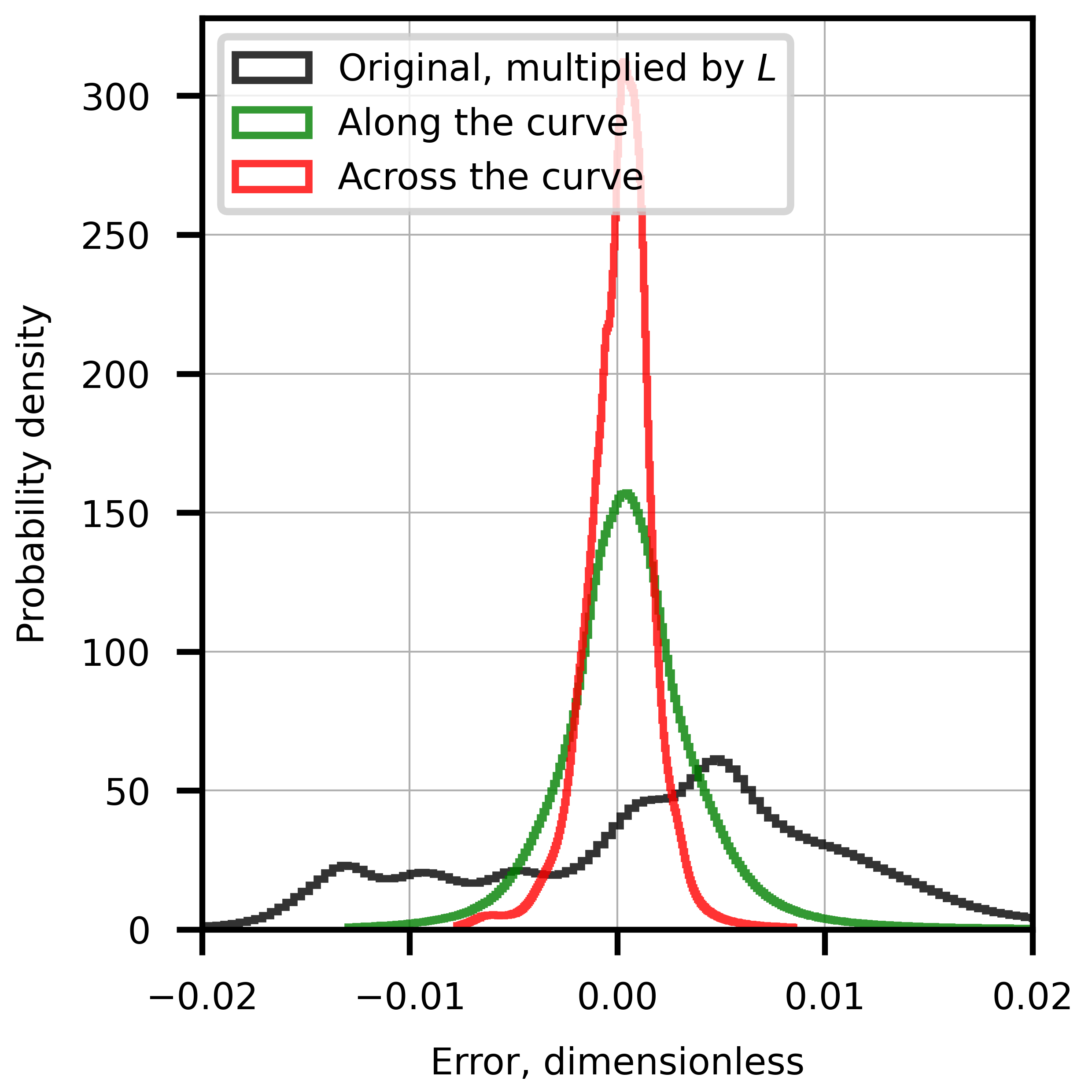}
        \caption{Distributions of errors across and along the Hilbert curve.}
        \label{fig:hilbert3-across-along-errors}
    \end{subfigure}
    \hfill
    \begin{subfigure}[t]{0.47\linewidth}
        \includegraphics[width=\linewidth]{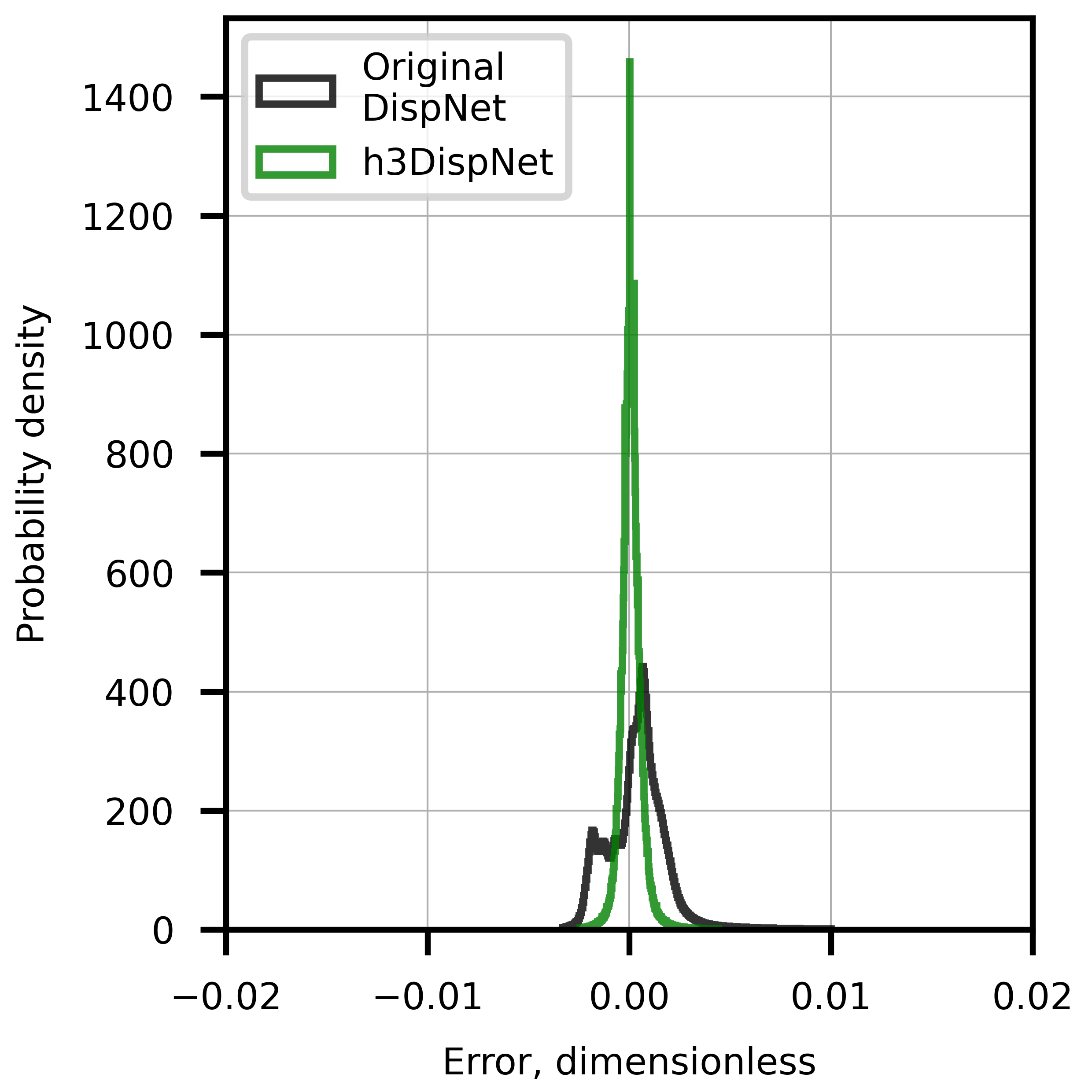}
        \caption{Disparity errors distribution for h3DispNet and DispNet models.}
        \label{fig:error-distr-dn-v4-p3}
    \end{subfigure}
    \caption{Quantization errors distributions for the h3DispNet model. Disparities $d$ are calculated from Hilbert components $x$ and $y$ and normalized to $[0,1]$ range.}
    \label{fig:error-historgams}
    \vspace*{-0.3cm}
\end{figure}

For the detailed analysis of quantization error reduction, we selected the h3DispNet model. To exclude precision loss influence, we perform analysis for W8A8 model run on CPU. The difference between the original and modified model lies in the number of channels utilized: the relationship between the quantization errors among two Hilbert components requires further examination. Our initial analysis in Section~\ref{sec:quatization_transformation} was made under the assumption of independent quantization errors for the Hilbert components. However, assessment using real data demonstrated that this hypothesis does not hold universally, as correlation has been discovered along the Hilbert curve. Quantitatively, the distribution of quantization errors along-the-curve is wider than that across-the-curve (Fig. \ref{fig:hilbert3-across-along-errors}).

Across-the-curve errors are mainly nullified in the post-processing, along-the-curve errors are compressed $L$ times and define the level of errors of the target quantity, that is disparity. To compare errors in the same scale, we added in Fig. \ref{fig:hilbert3-across-along-errors} disparity error of the original model multiplied by the curve length $L$. Quantization error in Hilbert curve-based output representation is significantly smaller than expected if the original model errors are scaled $L$ times. As a result, in disparity space, the quantization error of the modified model is significantly smaller than that of the original one (see Fig. \ref{fig:error-distr-dn-v4-p3}): $\hat{\sigma}$ is $5.65 \cdot 10^{-4}$ and $17.66 \cdot 10^{-4}$, respectively. Thus, we obtained quantization error reduction by $\approx$3.1 times on CPU. On DSP this gain is $\approx$4.6 times. Similar effect is observed for the DPT.

In order to bring further understanding, we applied Hilbert curve-based output representation to the Human Pose Estimation (HPE) task (see Appendix~\ref{sec:hpe_experiment}). For the ResNet-RS~\cite{Bello2021RevisitingRI} encoder and keypoints representation as heatmaps, the effect is similar to stereo matching but with a stronger along-the-curve correlation. Quantization error reduction on DSP is 2.69 times for $p=3$ curve. For direct keypoints regression along-the-curve correlation is so strong that quantization error reduction is not observed. For both stereo matching and HPE, we found that the magnitude of the across-the-curve quantization error is similar to the quantization error of the original model.



\section{Discussion and Limitations}
\label{sec:dis}

The proposed method is designed with the main goal of increasing the bit-precision of the predicted depth on devices with low-bit precision output. Apart from this, we observed a positive side effect consisting of a significant reduction of the quantization error. The effect reveals itself in different degrees for the depth prediction and human pose estimation with heatmaps; it is not observed for the human pose estimation with direct keypoints regression.

Our experiments show that the quantization error of Hilbert components can be correlated and have a large SD along the Hilbert curve. This effect manifests itself strongly for the human pose estimation with direct regression and only slightly for the stereo matching models. Such a correlation indicates that instead of learning independent paths for predicting Hilbert curve components, the modified model learns an internal representation of the target quantity (for example, disparity, depth or keypoints) and converts it into Hilbert components at the final model layers. In this case, the modified model predicts the target quantity with the same quantization error as the original model, this error undergoes forward transform to Hilbert representation (inside the model) and backward transform to the target quantity (at the post-processing step). As a result, the quantization error level remains the same, and the effect of our method application is only in bit-width precision increase. This is the case for the HPE with direct regression.

If both $x$ and $y$ are predicted by independent paths inside the model, the quantization errors of the Hilbert components are not correlated and have similar level as for the original model. This error is compressed $L$ times at the Hilbert components post-processing stage. As a result, our method allows both reducing quantization error and increasing bit-width precision as has been manifested by our stereo matching experiment. If the source of correlation in the along-the-curve direction can be found and eliminated, the proposed method could be used to reduce quantization error in cases when bit-precision is not an issue.

While our method improves the quality of predicted depth, it requires retraining of the full-precision model. In addition, our method requires that errors of the quantized depth prediction model are bounded, since large values of the quantization error violate one-to-one correspondence between the depth values and the points on the Hilbert curve. The next step involves joint usage of QAT and our method and moving to higher dimensions where a multidimensional Hilbert curve of the same length fills the space less densely.

We present results for stereo-matching models that predict disparity in one step. The recent models showing the best quality for stereo matching~\cite{xu2023iterative, SelectiveStereo} and monocular depth prediction~\cite{shao2023nddepth} incorporate iterative disparity refinement using GRU or similar recurrent units. In these architectures, the predicted disparity is refined over multiple iterations by adding a small correction signal to the initially predicted disparity map. Integration of the proposed idea of disparity representation as two Hilbert curve components into iterative models differs significantly from one-stage models like DispNet and DPT. While we experimented with modification of the output disparity representation, modification of iterative models requires integration of Hilbert components inside the model at multiple places. We leave this interesting problem for future work.

\section{Conclusions}
\label{sec:conclusion}

In this paper, we proposed a novel method for high dynamic range depth prediction on devices with low-precision arithmetic that exploits depth representation as points on a 2D Hilbert curve. This representation essentially codes the high dynamic range depth as two low dynamic range Hilbert curve components. The depth prediction model is trained to directly predict two Hilbert curve components that are calculated on-device in low-bit precision and used to reconstruct depth in high-bit precision. Apart from increasing bit-precision, our method reduces quantization error by a factor of up to 4.6. Experiments demonstrate that for the stereo matching task our method reconstructs depth in INT10-INT11 bits for a model quantized in W8A8 format and with quality similar to or even better than the original model quantized in W8A16 format. In this manner, depth can be predicted on-device 1.4-2 times faster and with 65\% of power consumption without sacrificing its quality. The proposed approach is beneficial for on-device application of different dense depth prediction methods including monocular and stereo depth prediction, Multi-View-Stereo, depth completion, depth quality enhancement, and depth inpainting. Future efforts need to be made to understand and fully explore the effect of quantization error reduction.

\section*{Impact Statement}

This paper proposed an innovative approach for increasing bit-precision of quantized depth prediction models on devices with low-precision arithmetic. This paper contributes to the advancement of quantization techniques for dense prediction tasks for devices with limited resources. While there are possible social implications that arise from our work, none of them are particularly relevant in this context.

\bibliography{main}
\bibliographystyle{icml2025}

\newpage
\appendix
\onecolumn

\section{Parametric curve selection}
\label{sec:curves}

\begin{wrapfigure}{rt}{0.40\textwidth}
    \begin{subfigure}[b]{0.34\textwidth}
    \centering
        \includegraphics[width=\textwidth]{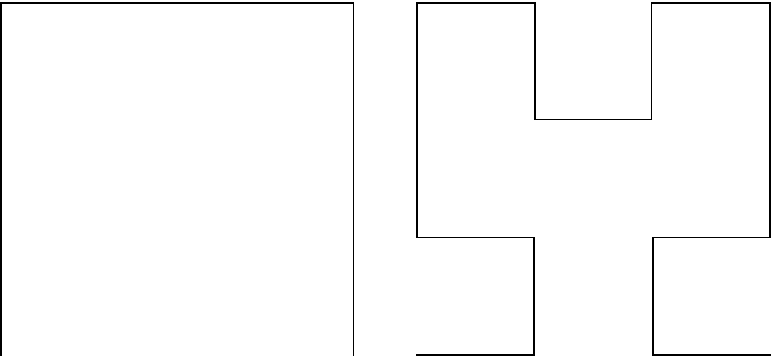}
        \caption{}
        \label{fig:sup-hilbert-curve}
    \end{subfigure}
    
    \vspace{0.5em} 
    
    \begin{subfigure}[b]{0.34\textwidth}
    \centering
        \includegraphics[width=\textwidth]{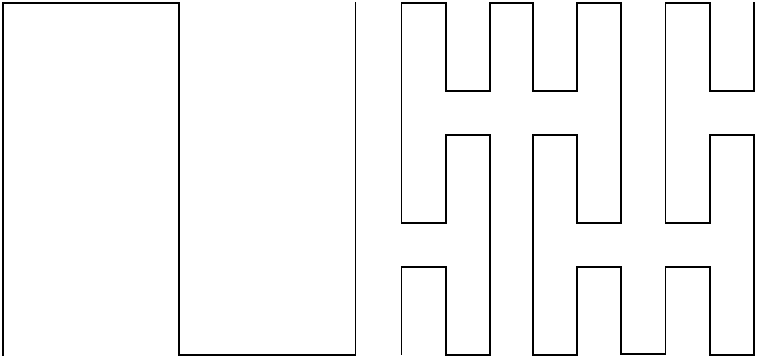}
       \caption{}
        \label{fig:sup-peano-curve}
    \end{subfigure}
    
    \vspace{0.5em} 
    
    \begin{subfigure}[b]{0.34\textwidth}
    \centering
        \includegraphics[width=\textwidth]{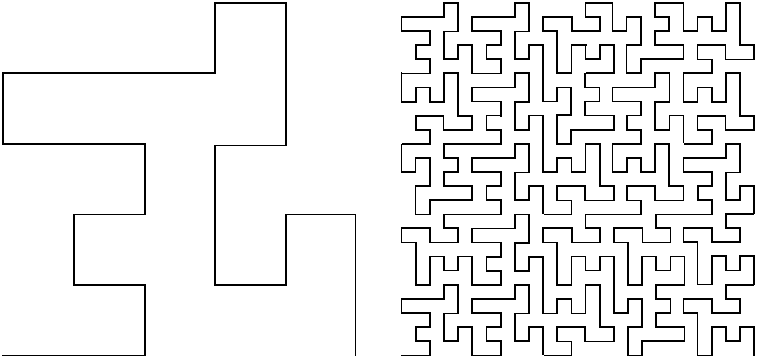}
       \caption{}
        \label{fig:sup-gosper-curve}
    \end{subfigure}
\caption{Space-filling curves filling unit square. The first (left column) and the second (right column) order $p$ curves for: (a)~Hilbert curve, (b)~Peano curve, and (c)~Quadratic Gosper curve.}
    \label{fig:sup-space-filling-curves}
     \vspace*{-0.3cm}
\end{wrapfigure}


The Hilbert curve is one representative of the wide class of space-filling curves~\cite{ventrella2012brainfilling}. Let us provide additional arguments in favor of Hilbert curve selection for high-precision depth prediction. In our analysis, we follow the terminology of \cite{ventrella2012brainfilling} that classifies curves on square and triangular grids, and divides them into families $\sqrt{N}$. In family $\sqrt{N}$, the distance between starting and ending points of the curve generator equals to $\sqrt{N}$.

In order to fully utilize depth representation as two components, it is desirable that a space-filling curve uniformly covers the unit square. This requirement eliminates all curves on triangular grids.

Curves with non-orthogonal generators have the drawback of filling square not uniformly as for example Z-order curves. Among curves with orthogonal generation curves defined on square grid our choice is limited to the Hilbert curve ($\sqrt{4}$ family) (Fig.~\ref{fig:sup-hilbert-curve}), the Peano curve ($\sqrt{9}$ family)(Fig.~\ref{fig:sup-peano-curve}) and the Quadratic Gosper curve ($\sqrt{25}$ family)(Fig.~\ref{fig:sup-gosper-curve}). For Hilbert and Peano curves different generators are possible (e.g., the Moore curve is a variant of the Hilbert curve) but they are different only in the order the space is filled.

The value of N defines how fast the curve length $L_p$ increases and curve edge size $h_p$ decreases with the curve order. Our experiments show that it is desirable to have the ability of fine-tuning the curve length depending on the DNN quantization error magnitude. From this point of view, the Hilbert curve is the most flexible as it has the lowest N value. For the Hilbert curve, the number of nodes grows as 1, 4, 16, 64, 256 with the order $p$. For Peano curve, the number of nodes grows as 1, 9, 81, 729, 6561, and for Quadratic Gosper curve as 1, 25, 625, 15625, 390625. If we limit the number of nodes to a reasonable value of 256, the Hilbert curve provides 4 usable low-order curves (that we experiment with in the paper), Peano~--~2, Quadratic Gosper Curve~--~1.

The Hilbert curve is the simplest and the most flexible curve that satisfies all requirements essential for coding depth values. To construct even more flexible list of curves, it is possible to use Hilbert, Peano and Quadratic Gosper curve of different orders to create a sequence of curves with the number of nodes 4, 9, 16, 25, 64, 81, 256.

Provided the main requirements for the parametric curve (self-avoidance, uniform filling of unit square, continuity) are satisfied, the detailed structure of the curve is not important. For example, we can use arbitrary non-self-similar curves that fill unit square with a given number of nodes, curves with smoothed corners, curves that stretch different parts of 1D value in a different degree (to emphasize the most probable range of target quantity variation).

\section{DPT model modification}
\label{sec:DPT}

\begin{figure}[t]
   \centering
   \includegraphics[width=0.9\linewidth]{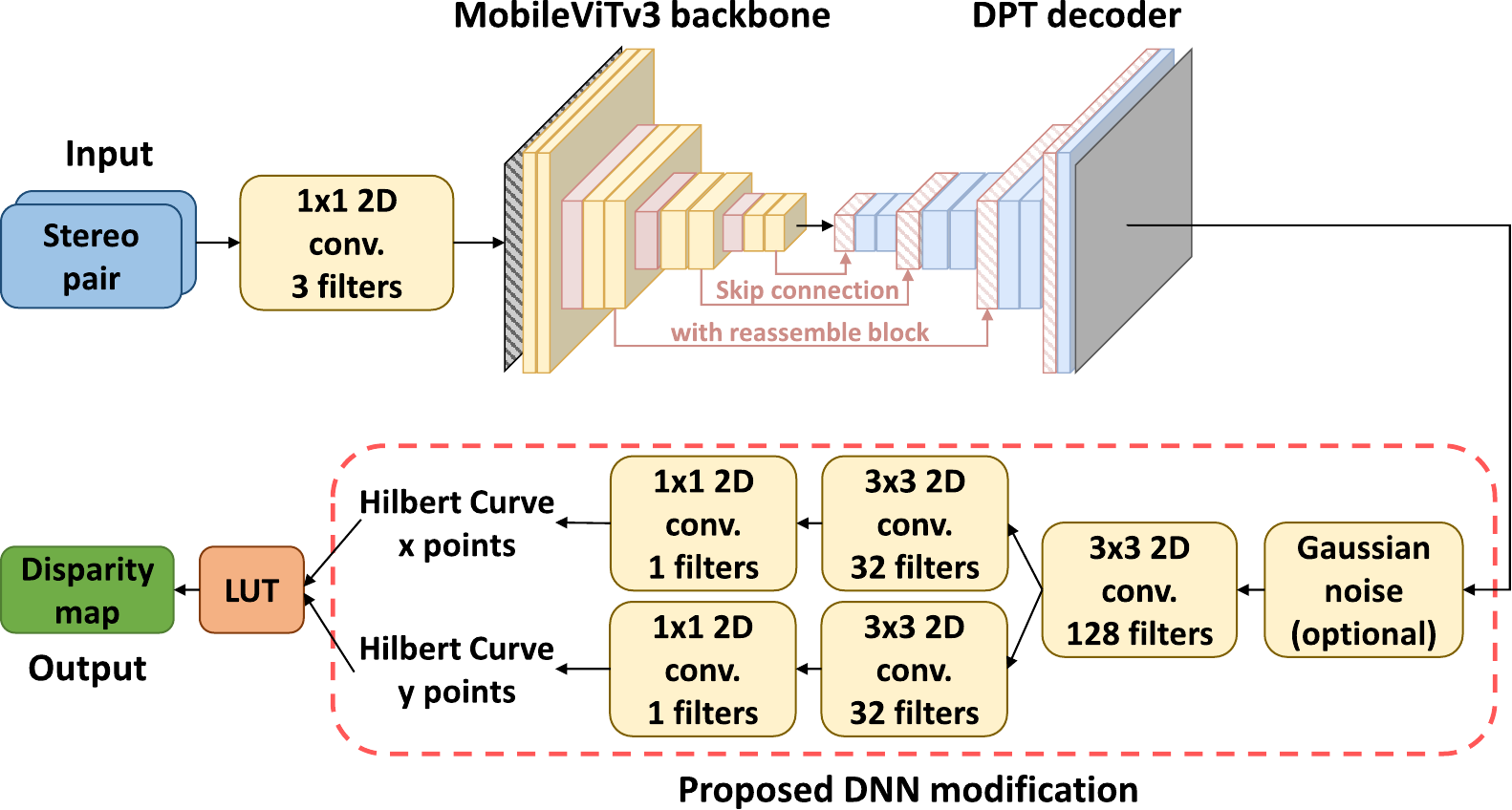}
   \caption{The hpDPT model architecture. The input RGB stereo pair is processed by an encoder which is MobileViTv3-S backbone and the decoder proposed for depth prediction in DPT. Features from the decoder are fed to an optional Gaussian noise layer and $3\times3$ 2D convolution layer followed by two heads for Hilbert curve components. They consist of one $3\times3$ and one $1\times1$ 2D convolution layers with a decreasing number of filters: 32 and 1 respectively. At the post-processing stage, Hilbert components are converted to the final disparity map.}
   \label{fig:hpDPT-architecture}
\end{figure}

As it mentioned in the paper, one of the models chosen for the experiments is Dense Prediction Transformer (DPT)~\cite{DenseTransformer}. All modifications of the model architecture are illustrated in Fig.~\ref{fig:hpDPT-architecture}. An additional $1\times1$ 2D convolution layer was used for proper integration of the input RGB stereo pair into the MobileViTv3-S~\cite{MobileViTv3} backbone.

Also, MobileNet blocks in the encoder are modified for better quantization as described by Sheng et al.~\cite{QuantizationFriendlyMN}. For disparity prediction, the original DPT head was used. The Hilbert curve head architecture for this model includes an additional up-sample layer after the first $3\times3$ 2D convolution layer.

The DPT model includes a MobileViTv3-S encoder with skip connections before each MobileViT block. Each skip connection integrates into the decoder part using a reassemble block proposed by Ranftl et al.~\cite{DenseTransformer}.

During analysis of the network architecture, we found that layers in MobileNet blocks have large kurtosis values of their weights' distributions. It is suggested in~\cite{RobustQuantization} that large kurtosis values might lead to the model quantization quality degradation because of outliers clipping. We added kurtosis regularization proposed in~\cite{RobustQuantization} to all $1\times1$ convolutions in MobileNet blocks in MobileViTv3-S to reduce quantization error in both experiments with depth and Hilbert components output.

\section{Experiment on KITTI 2012 dataset}
\label{sec:kitti}
\begin{wraptable}{r}{0.4\textwidth} 
\centering
\begin{scriptsize} 
\begin{tabular}{c @{\kern1em}c @{\kern1em}c @{\kern1em}c }
    \toprule
    \rowcolor{white}
     Precision &  Abs Rel, \% $\downarrow$ &  EPE, px $\downarrow$ &  D1, \% $\downarrow$\\
        \midrule
    FP32 &  3.88 & 1.38 & 9.13  \\
    W8A8 & 5.01 & 1.63 & 9.95 \\
   Ours, FP32 & 3.24 & 1.07 & 5.22 \\
   Ours, W8A8 & \textbf{3.30} & \textbf{1.09} & \textbf{5.36} \\

      \bottomrule
    \end{tabular}
    \end{scriptsize}
    \caption{Results on KITTI 2012 for the DispNet and h2DispNet models. Models in FP32 format are run on CPU; models in W8A8 formats are run on DSP. Best metrics for W8A8 models are shown in bold.}
    \label{tab:kitti}
\end{wraptable}

We evaluate our approach on KITTI~2012~\cite{geiger2012we} dataset. KITTI~2012 is a real-world dataset for autonomous driving domain with sparse ground-truth disparities collected with a LiDAR system. Our evaluation is based on 194 images in the training part of KITTI~2012. Training dataset is composed of ScanNet and Virtual KITTI~2~\cite{cabon2020virtual} datasets with 25/75\% balancing. The DispNet and h2DispNet models were trained in 256 by 1152~px input resolution and 128 by 576~px output disparity resolution. Training settings are the same as for the ScanNet experiment. 

\begin{figure*}
    \centering
    \begin{subfigure}{0.45\linewidth}
        \includegraphics[width=1.0\linewidth]{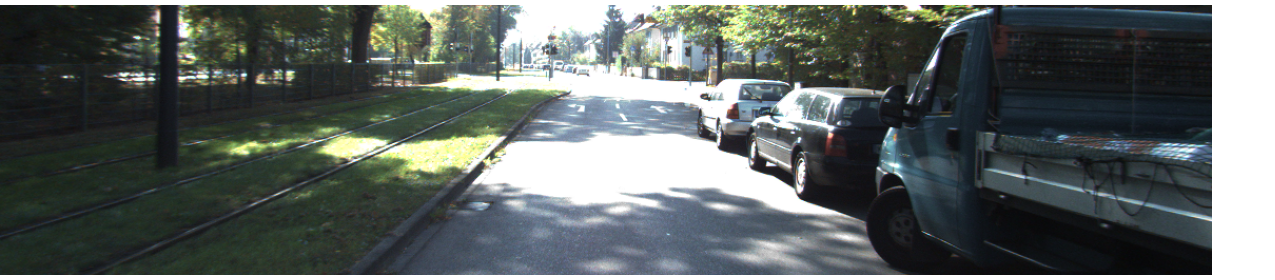}
        \caption{Image}
        \label{fig:4.1-image}
    \end{subfigure}
        \centering
        \begin{subfigure}{0.45\linewidth}
        \includegraphics[width=1.0\linewidth]{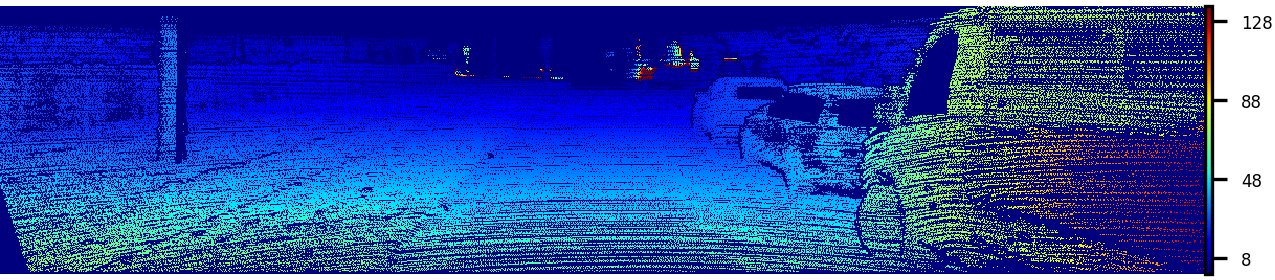}
        \caption{GT disparity}
        \label{fig:4.1-gt_disparity}
    \end{subfigure}
    \centering
    \begin{subfigure}{0.45\linewidth}
    \includegraphics[width=1.0\linewidth]{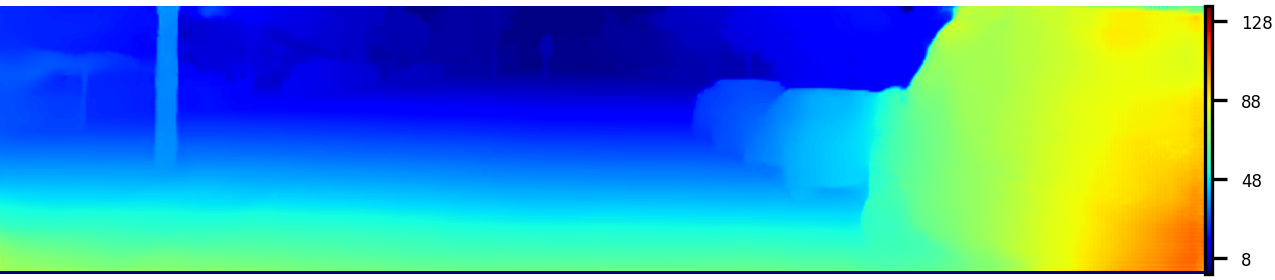}
        \caption{DispNet disparity, FP32}
        \label{fig:4.1-dispnet_fp32}
    \end{subfigure}
   \centering
    \begin{subfigure}{0.45\linewidth}
    \includegraphics[width=1.0\linewidth]{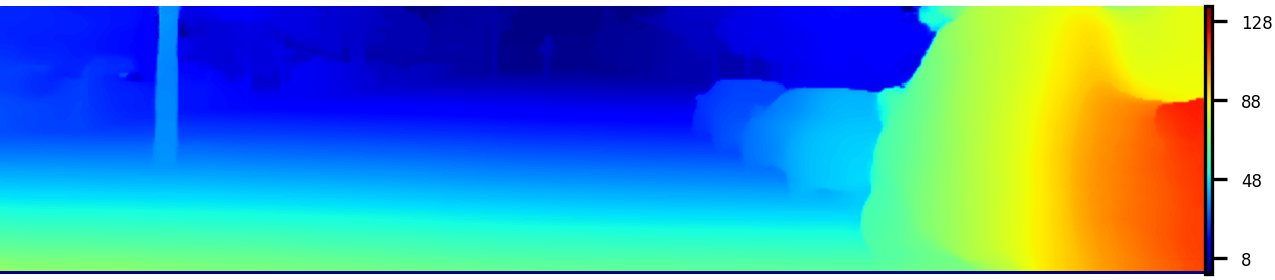}
        \caption{h2DispNet disparity, FP32}
        \label{fig:4.1-h2dispnet_fp32_dsp}
    \end{subfigure}
   \centering
       \begin{subfigure}{0.45\linewidth}
    \includegraphics[width=1.0\linewidth]{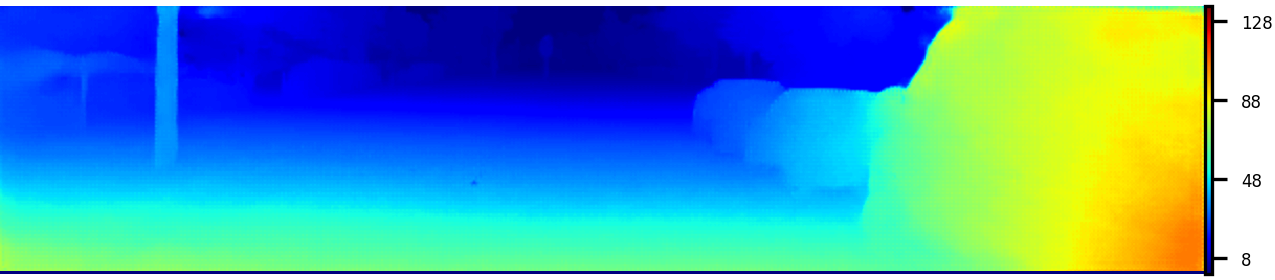}
        \caption{DispNet disparity, W8A8}
        \label{fig:4.1-dispnet_int8}
    \end{subfigure}
   \centering
    \begin{subfigure}{0.45\linewidth}
    \includegraphics[width=1.0\linewidth]{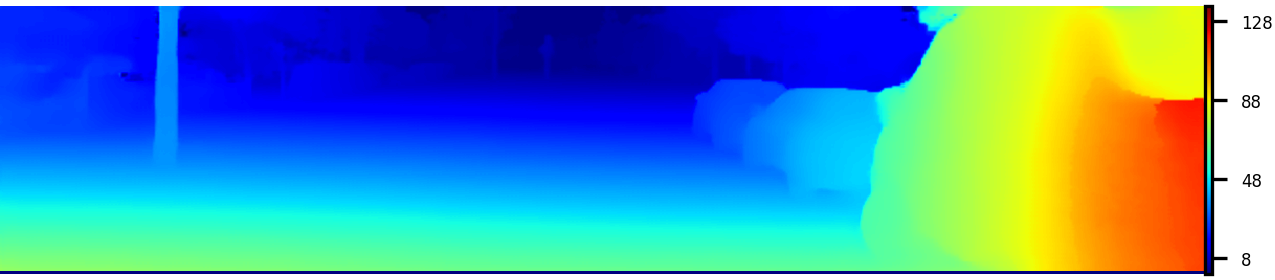}
        \caption{h2DispNet disparity, W8A8}
        \label{fig:4.1-h2dispnet_int8_dsp}
    \end{subfigure}
   \centering
    \begin{subfigure}{0.45\linewidth}
    \includegraphics[width=1.0\linewidth]{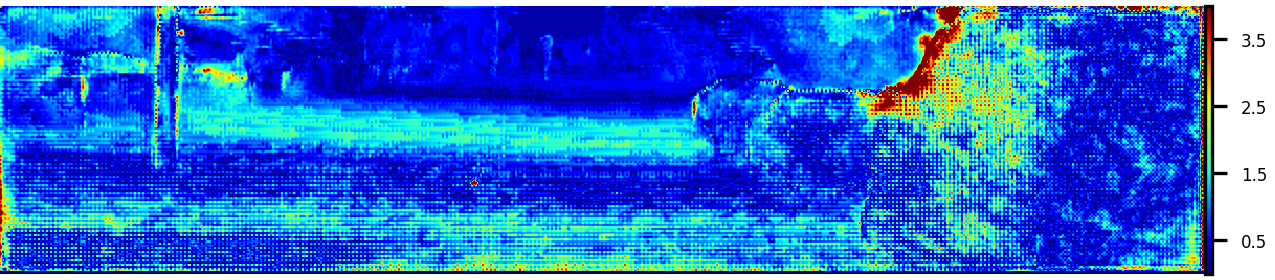}
        \caption{$|\text{FP32 - W8A8}|$, DispNet}
        \label{fig:4.1-dispnet_base_int8_cpu}
    \end{subfigure}
\centering
     \begin{subfigure}{0.45\linewidth}
    \includegraphics[width=1.0\linewidth]{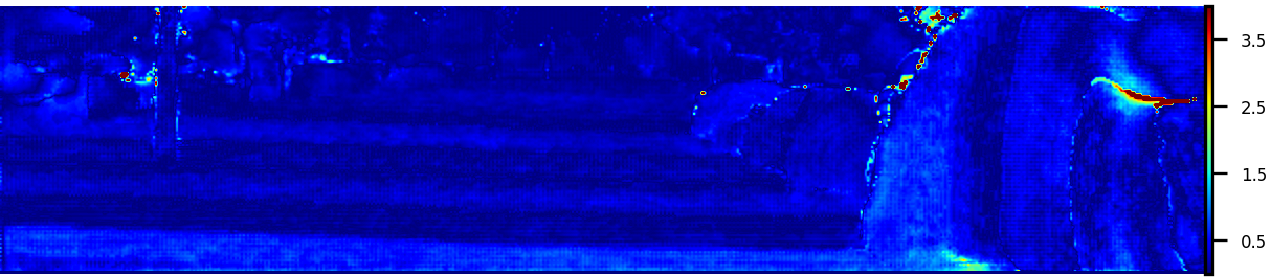}
       \caption{$|\text{FP32 - W8A8}|$, h2DispNet}
        \label{fig:4.1-h2dispnet_int8_cpu}
    \end{subfigure}
   \centering
    \begin{subfigure}{0.45\linewidth}
    \includegraphics[width=1.0\linewidth]{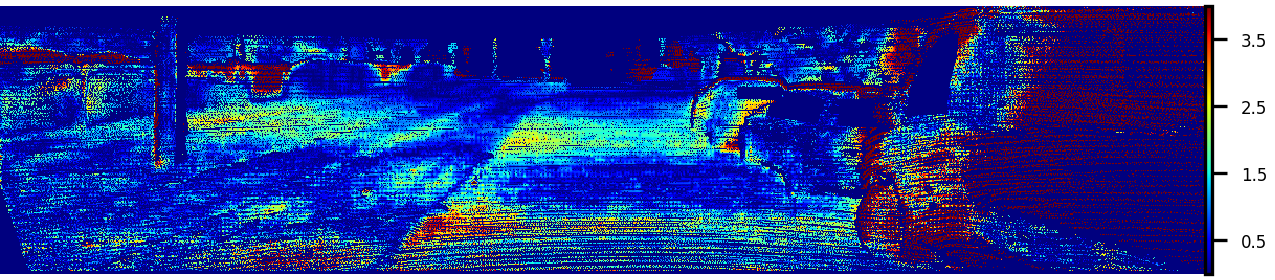}
       \caption{$|\text{GT - W8A8}|$, DispNet}
        \label{fig:4.1-dispnet_gt_int8_cpu}
    \end{subfigure}
\centering
     \begin{subfigure}{0.45\linewidth}
    \includegraphics[width=1.0\linewidth]{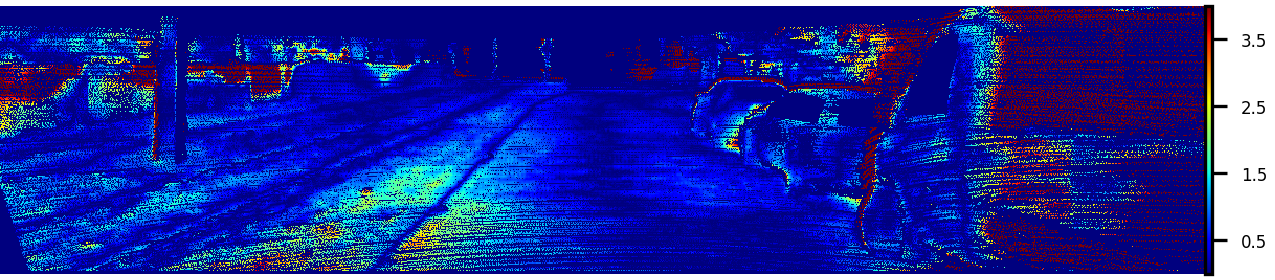}
        \caption{$|\text{GT - W8A8}|$, h2DispNet}
        \label{fig:kitti_illustration}
    \end{subfigure} 

    \caption{Comparison of DispNet and h2DispNet on KITTI 2012 dataset. Differences between disparity maps predicted by FP32 (CPU) (c, d) and W8A8 (DSP) (e, f) models are shown in (g) and (h); absolute disparity prediction errors for W8A8 (DSP) models are shown in (i) and (j). All values are presented in pixels.}
    \label{fig:4.1}
\end{figure*}

 As shown in Table~\ref{tab:kitti}, for the original DispNet model we reached EPE 1.38~px and D1 9.13\%. Interestingly, h2DispNet shows better results with EPE 1.07~px and D1 5.22\%. We exclude $S_c$ from the analysis, because it is not applicable to sparse ground-truth disparities of KITTI~2012 dataset. The DispNet model quantized to W8A8 format and run on DSP shows degradation of both EPE (to 1.63~px) and D1 (to 9.95\%). At the same time the h2DispNet in W8A8 format on DSP shows very minor quality degradation. Measured as EPE between disparity predicted by FP32 and W8A8 models, quantization error is 1.01~px for the DispNet and 0.38~px for the h2DispNet. It corresponds to gain in disparity prediction quality on device of 2.6 times. Overall, on DSP the h2DispNet improves D1 by approximately 4.6\% and EPE by about 33\% compared to the original DispNet model. Examples of predicted disparity are given in Fig.~\ref{fig:4.1}. This experiment demonstrates that the proposed approach can be successfully applied to real datasets.

\section{Quantization quality influence on mesh fusion}
\label{sec:mesh}

\begin{figure}
    \centering
    \begin{subfigure}{0.45\linewidth}
        \includegraphics[width=1.0\linewidth]{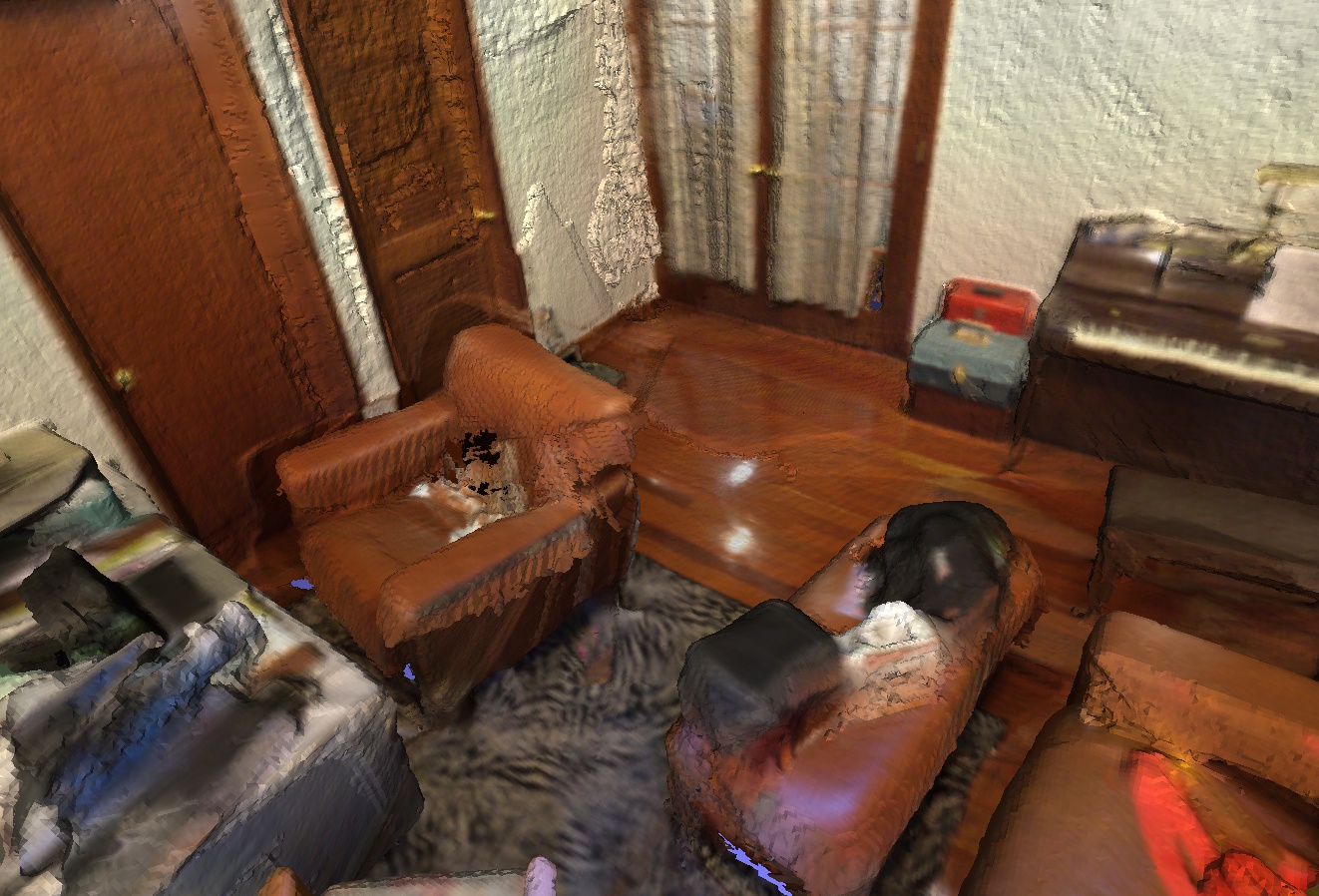}
        \caption{GT mesh with texture}
        \label{fig:3-gt_mesh_color}
    \end{subfigure}
    \centering
    \begin{subfigure}{0.45\linewidth}
    \includegraphics[width=1.0\linewidth]{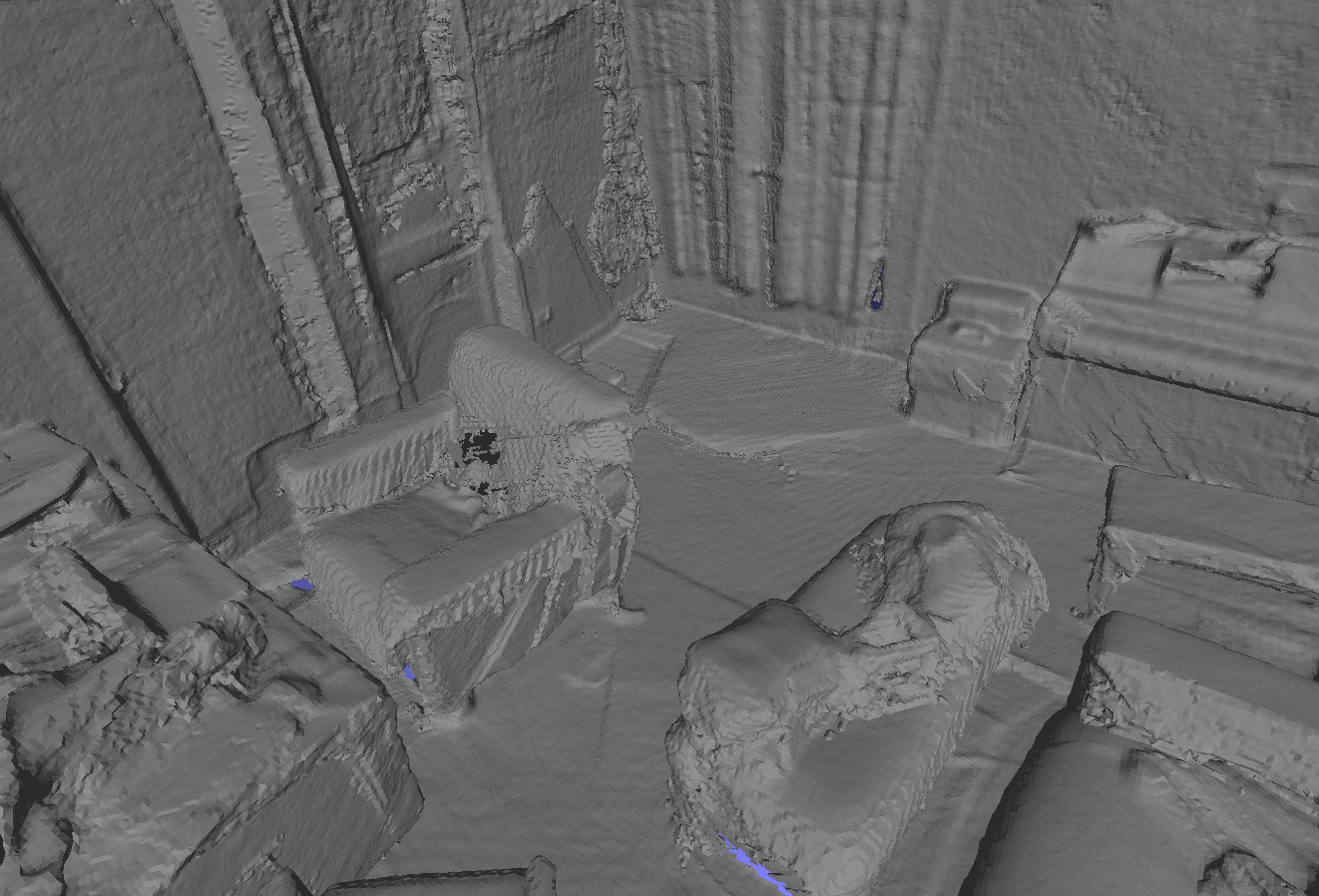}
        \caption{GT mesh without texture}
        \label{fig:3-gt_mesh_gray}
    \end{subfigure}
    \caption{A view of 3D mesh fused with GT depth maps for ScanNet scene scene0050\_02.}
    \label{fig:3}
\end{figure}

We provide additional experiment to understand how quantization artifacts in depth maps affect a scene mesh reconstruction. For a mesh fusion we utilize truncated signed distance function (TSDF)~\cite{tsdf} approach as implemented in Python library Open3D~\cite{open3d}. We used scalable TSDF volume with parameters $voxel\_length = 0.01 \, m$, $sdf\_trunc = 0.15$. For experiments we chose ScanNet scene scene0050\_02 comprising $4379$ frames. Each 40th frame was used to reconstruct mesh. Quality of the mesh fused from GT depth maps is illustrated in Fig.~\ref{fig:3}.

\begin{figure}
    \centering
    \begin{subfigure}{0.45\linewidth}
        \includegraphics[width=1.0\linewidth]{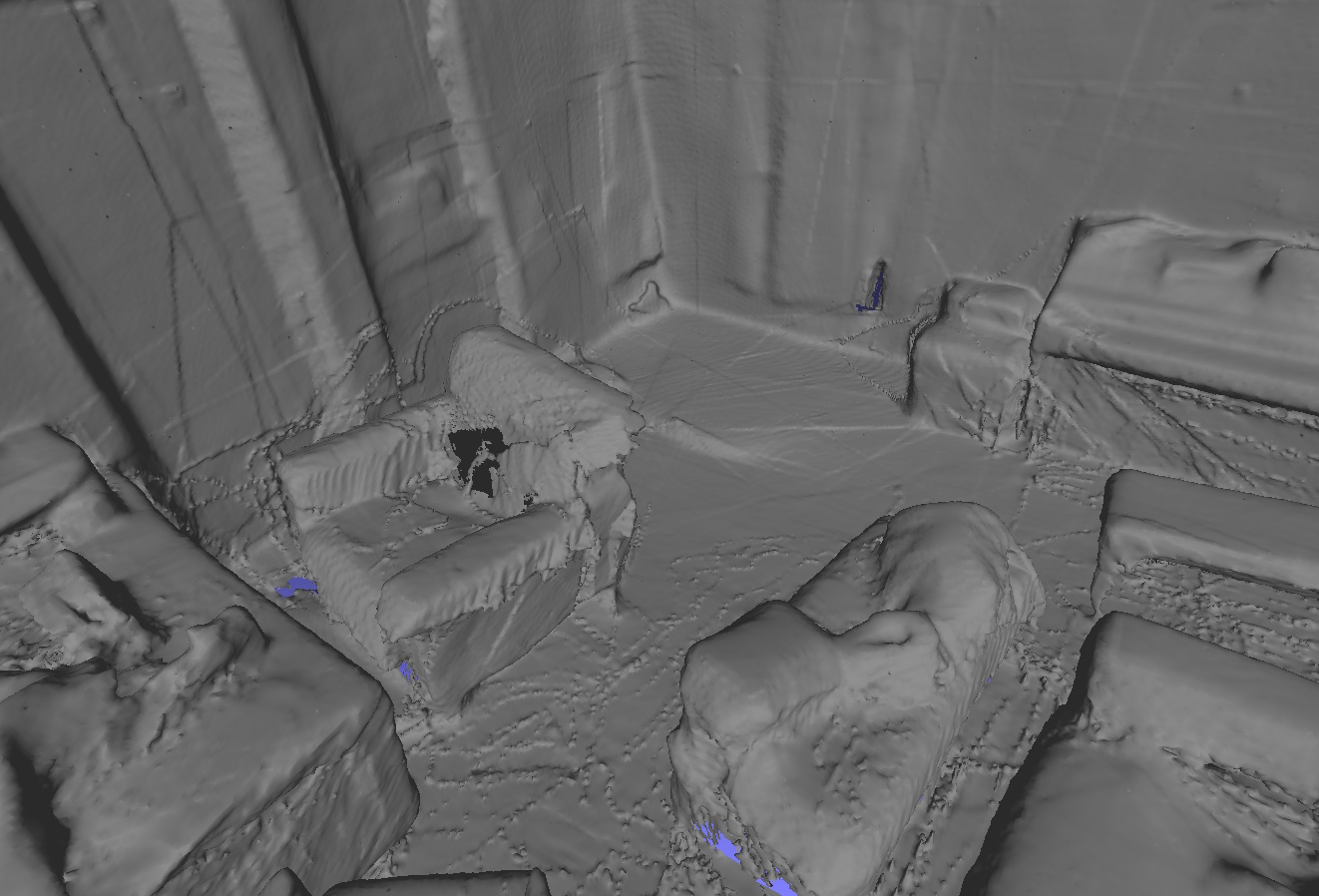}
        \caption{DispNet, FP32}
        \label{fig:4-dispnet_base_fp3212}
    \end{subfigure}
        \centering
        \begin{subfigure}{0.45\linewidth}
        \includegraphics[width=1.0\linewidth]{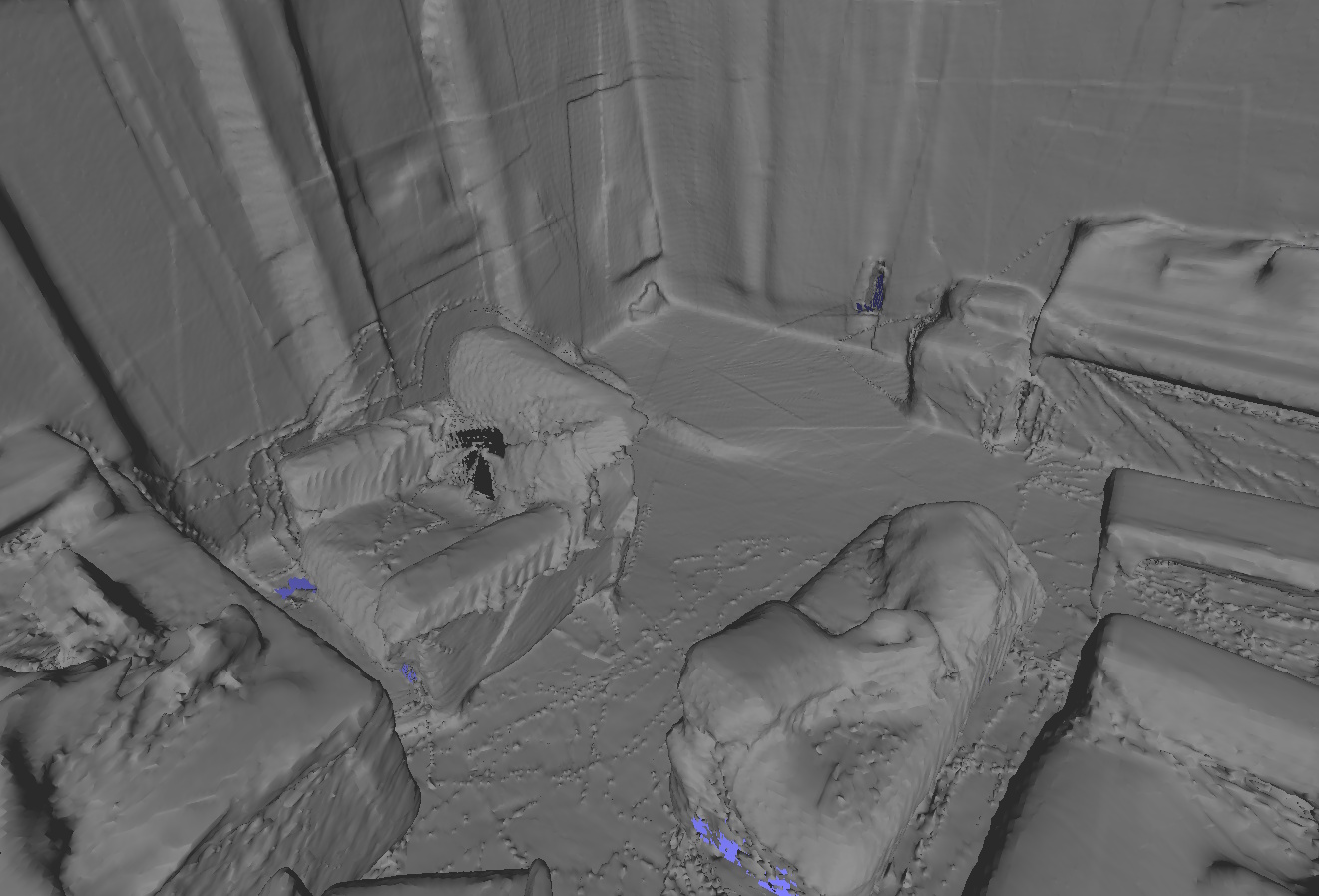}
        \caption{h2DispNet, FP32}
        \label{fig:4-h2dispnet_fp32}
    \end{subfigure}
    \centering
    \begin{subfigure}{0.45\linewidth}
    \includegraphics[width=1.0\linewidth]{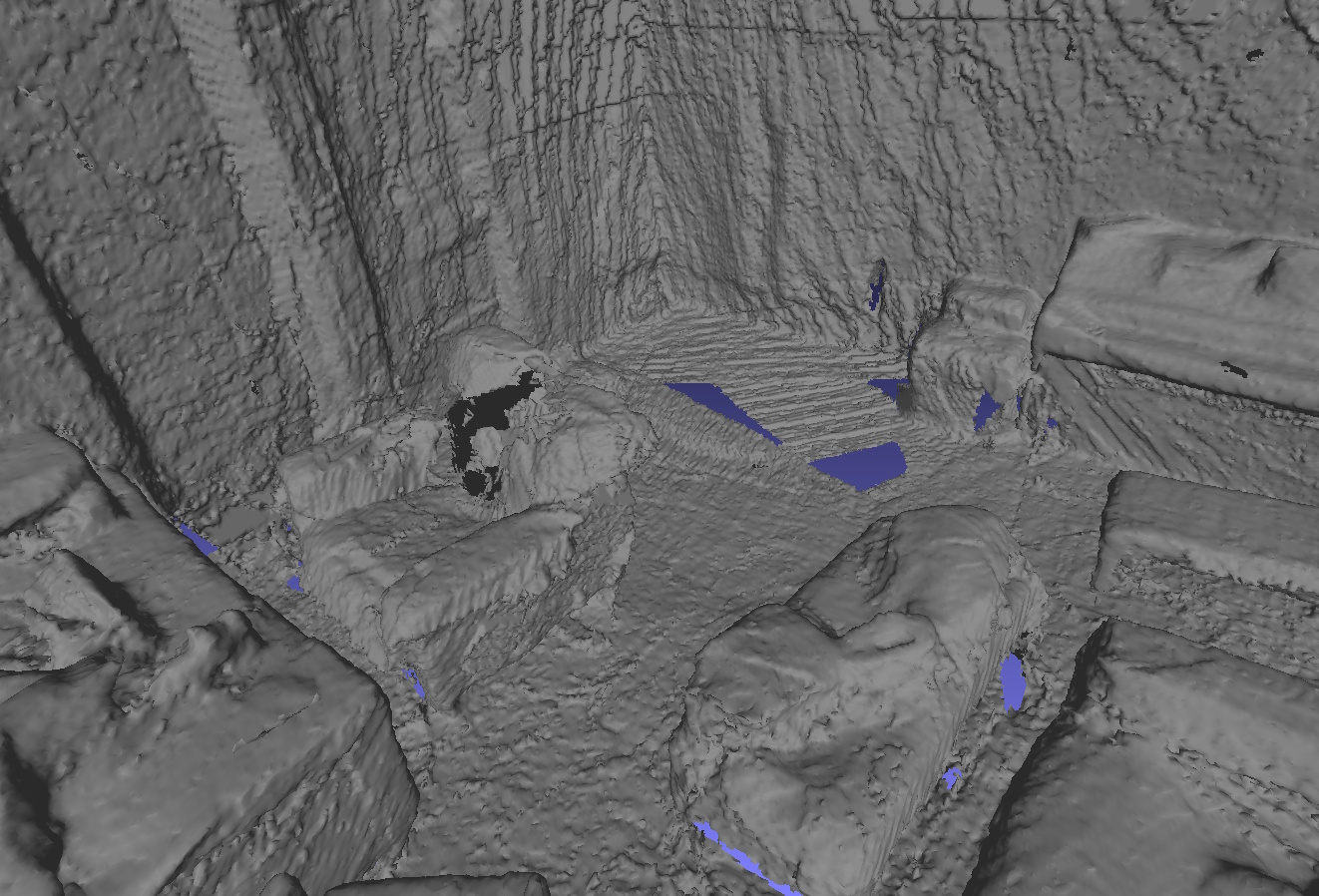}
        \caption{DispNet, W8A8(DSP)}
        \label{fig:4-dispnet_base_int8_dsp}
    \end{subfigure}
   \centering
    \begin{subfigure}{0.45\linewidth}
    \includegraphics[width=1.0\linewidth]{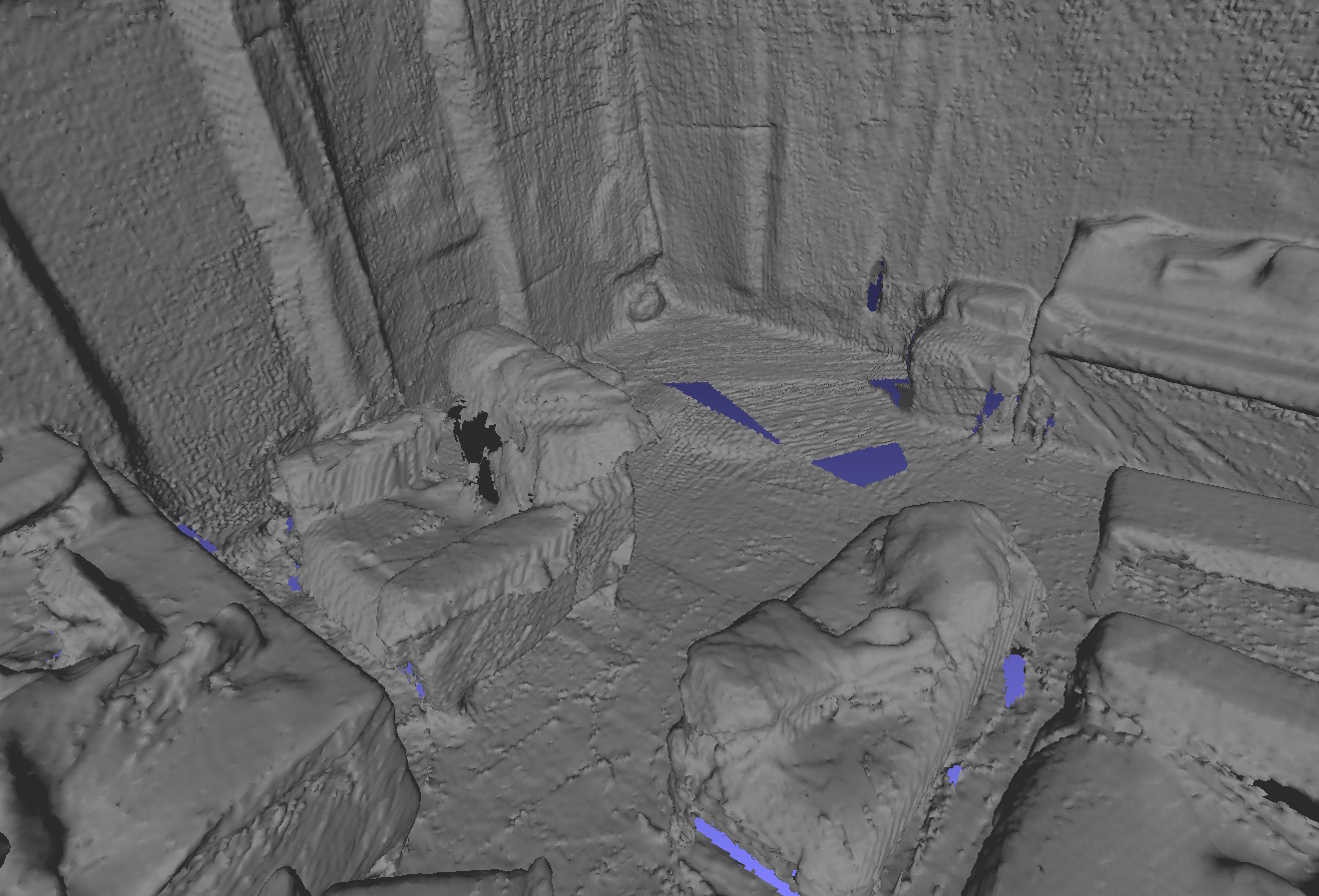}
        \caption{h2DispNet, W8A8(DSP)}
        \label{fig:4-h2dispnet_int8_dsp}
    \end{subfigure}
   \centering
    \begin{subfigure}{0.45\linewidth}
    \includegraphics[width=1.0\linewidth]{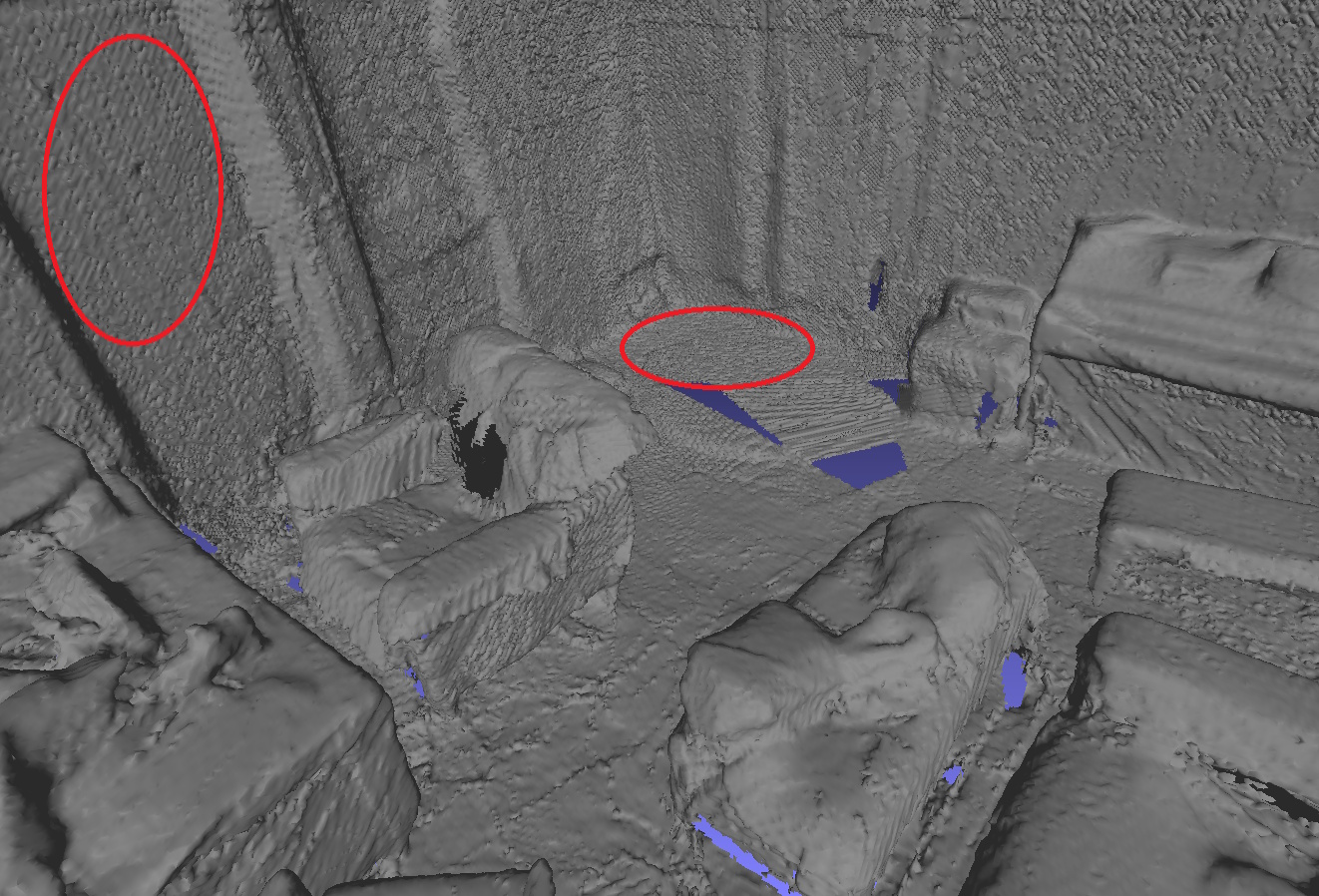}
        \caption{DispNe, W8A8(CPU)}
        \label{fig:4-dispnet_base_int8_cpu}
    \end{subfigure}
\centering
     \begin{subfigure}{0.45\linewidth}
    \includegraphics[width=1.0\linewidth]{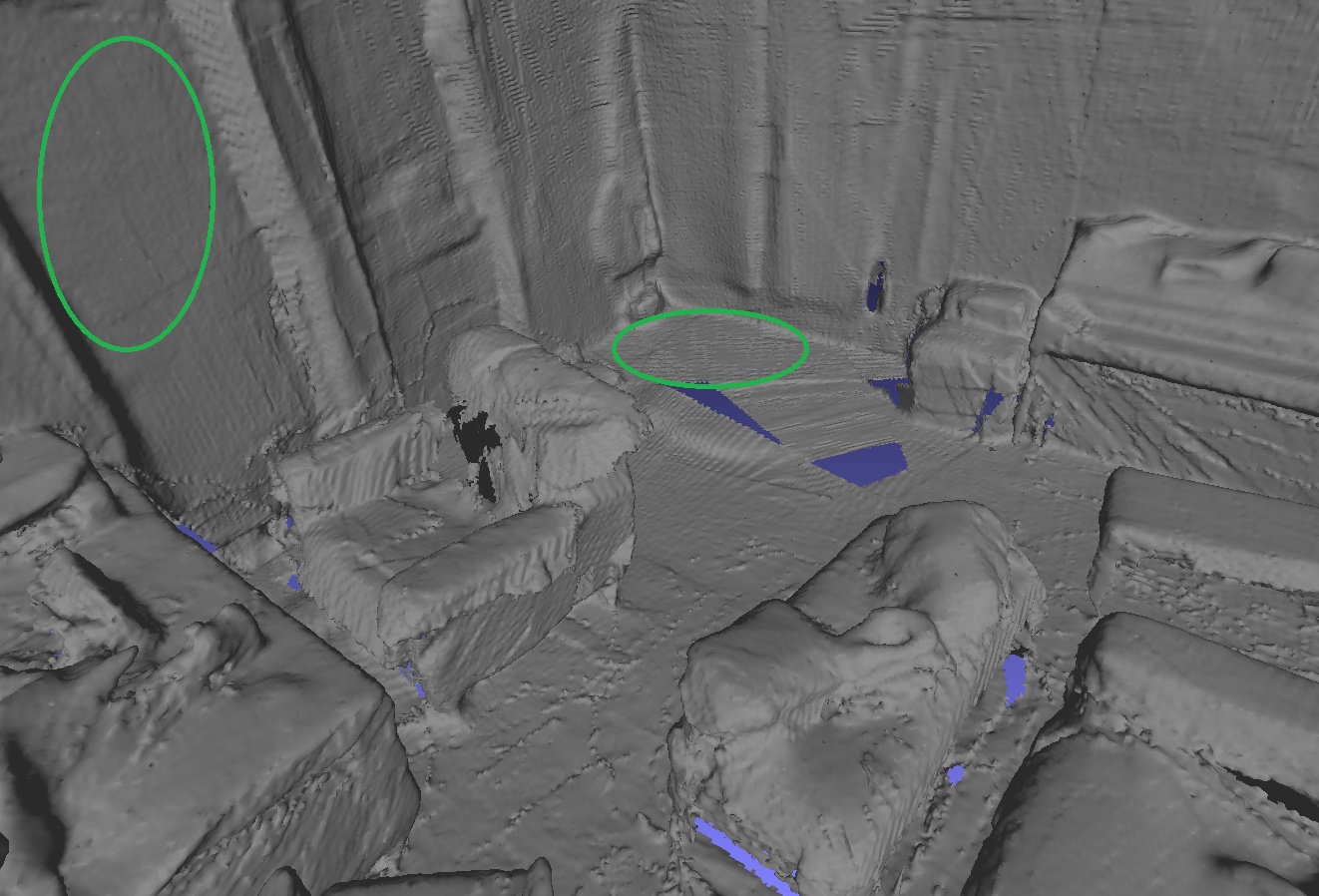}
        \caption{h2DispNet, W8A8(CPU)}
        \label{fig:4-h2dispnet_int8_cpu}
    \end{subfigure}   
    \caption{A view of 3D mesh fused with predicted depth maps by DispNet and h2DispNet for ScanNet scene scene0050\_02. Some reconstruction errors are highlighted by red and improved structures are marked by green.}
    \label{fig:4}
\end{figure}

\begin{figure}
    \centering
    \begin{subfigure}{0.45\linewidth}
        \includegraphics[width=1.0\linewidth]{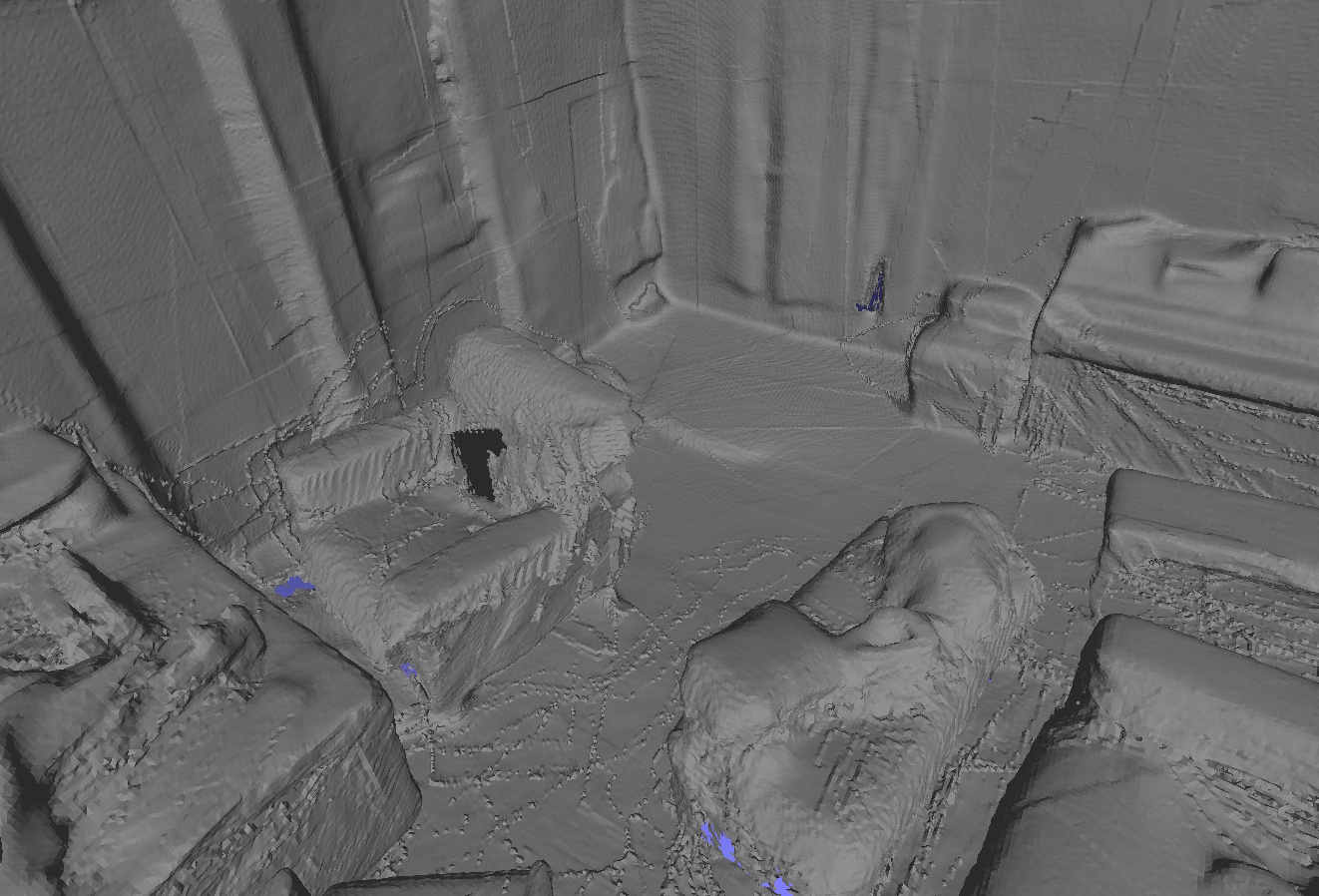}
        \caption{DPT, FP32}
        \label{fig:5-dpt_base_fp32}
    \end{subfigure}
        \centering
        \begin{subfigure}{0.45\linewidth}
        \includegraphics[width=1.0\linewidth]{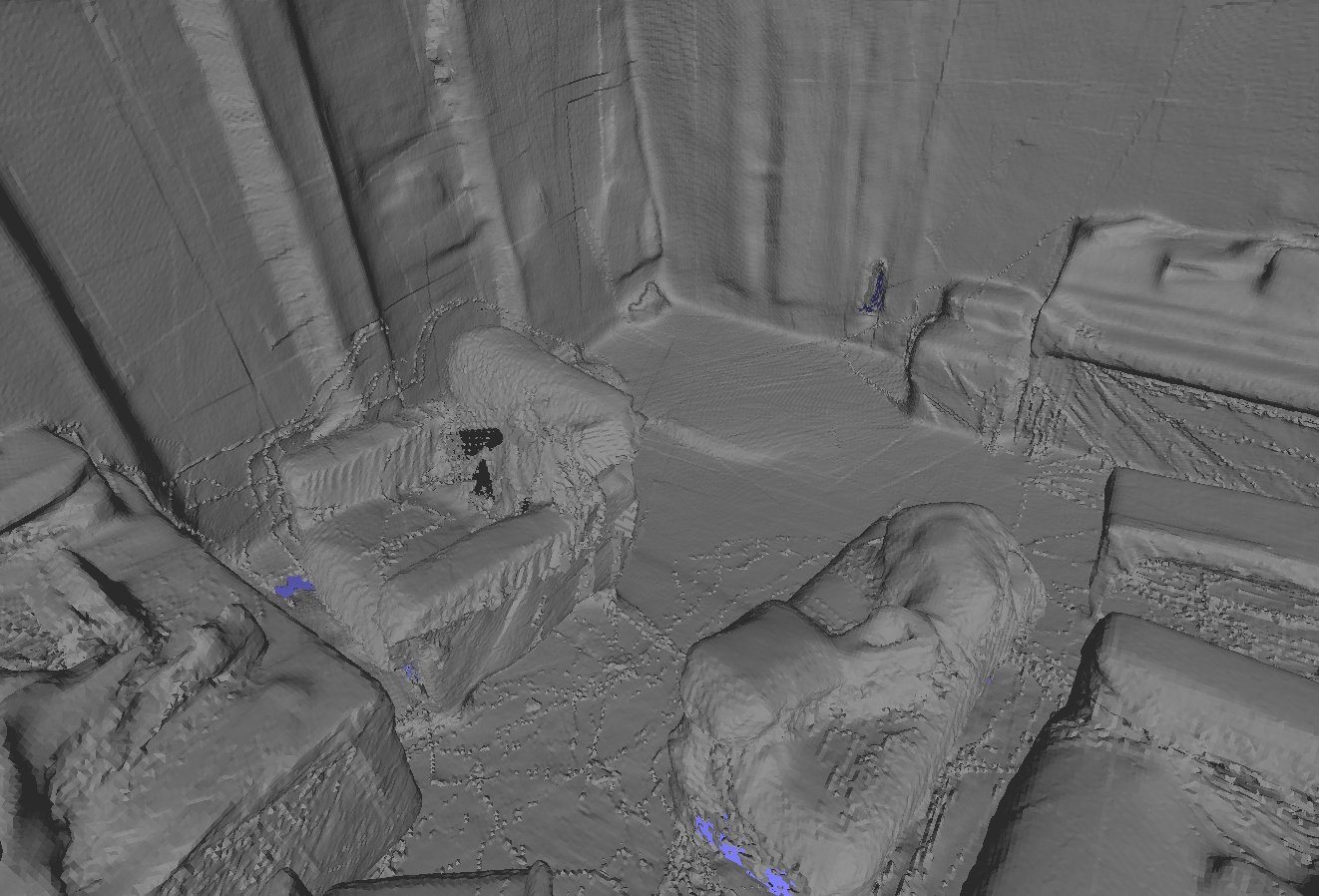}
        \caption{h2DPT, FP32}
        \label{fig:5-h2dpt_fp32}
    \end{subfigure}
    \centering
    \begin{subfigure}{0.45\linewidth}
    \includegraphics[width=1.0\linewidth]{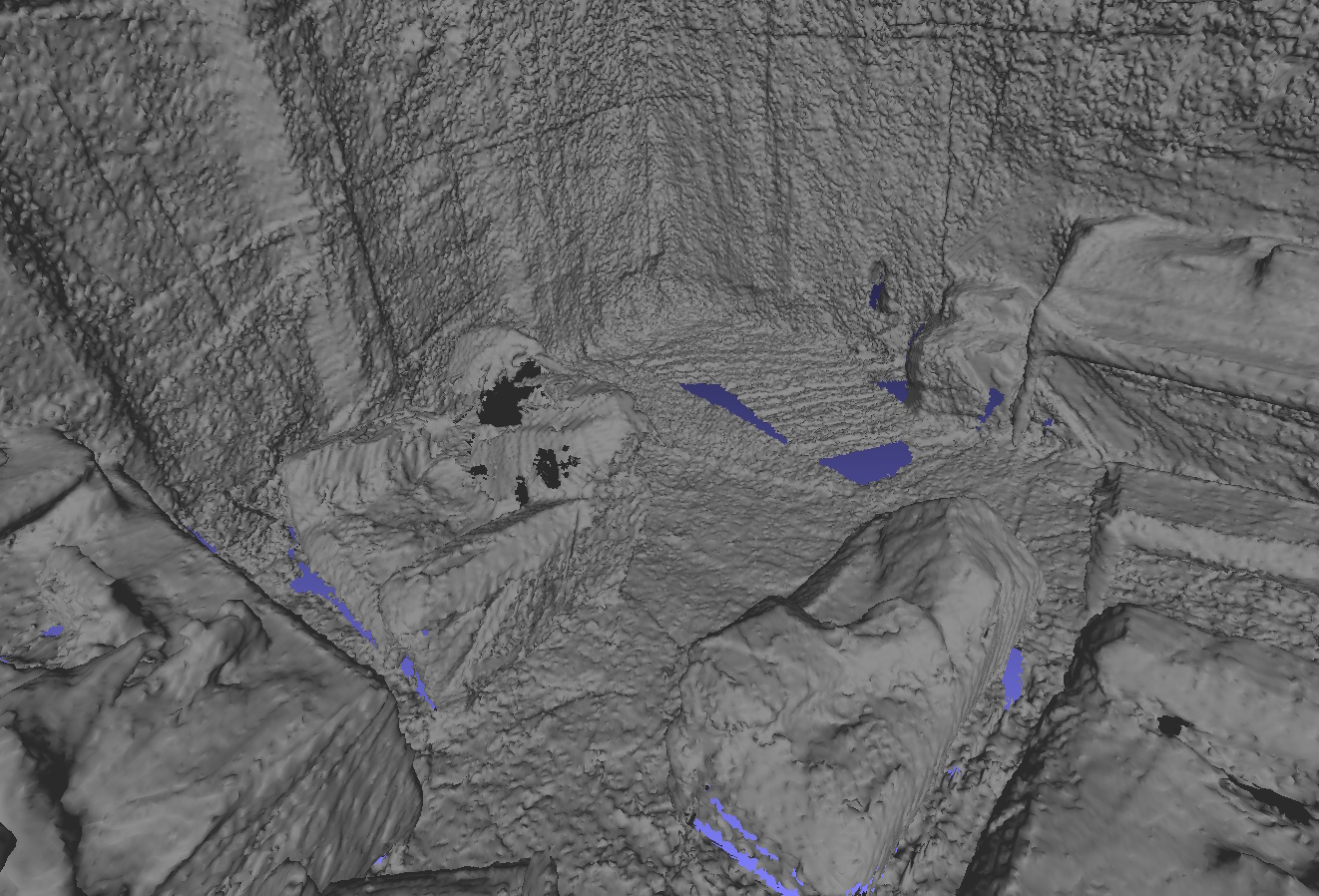}
        \caption{DPT, W8A8(DSP)}
        \label{fig:5-dpt_base_int8_dsp}
    \end{subfigure}
   \centering
    \begin{subfigure}{0.45\linewidth}
    \includegraphics[width=1.0\linewidth]{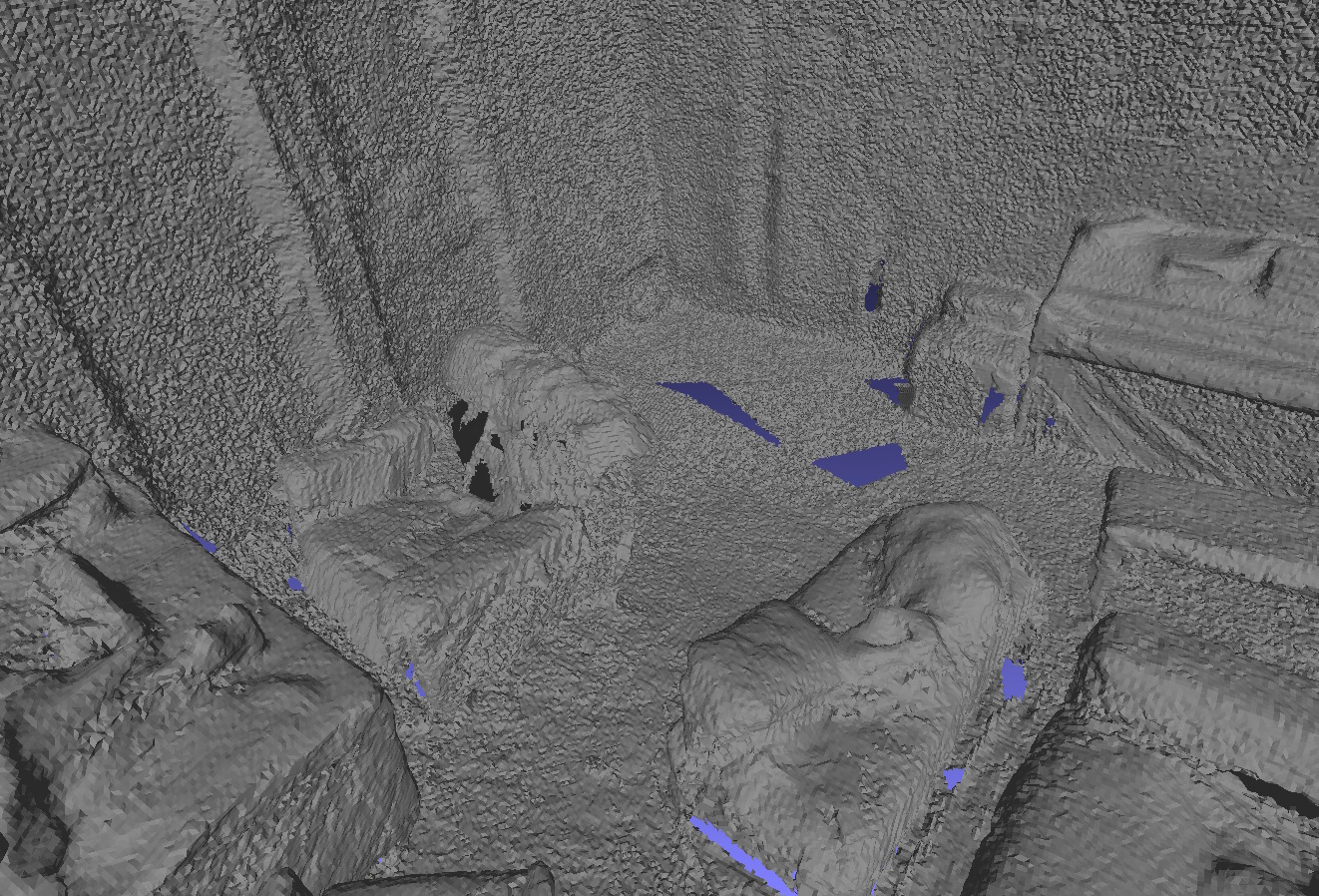}
        \caption{h2DPT, W8A8(DSP)}
        \label{fig:5-h2dpt_int8_dsp}
    \end{subfigure}
   \centering
    \begin{subfigure}{0.45\linewidth}
    \includegraphics[width=1.0\linewidth]{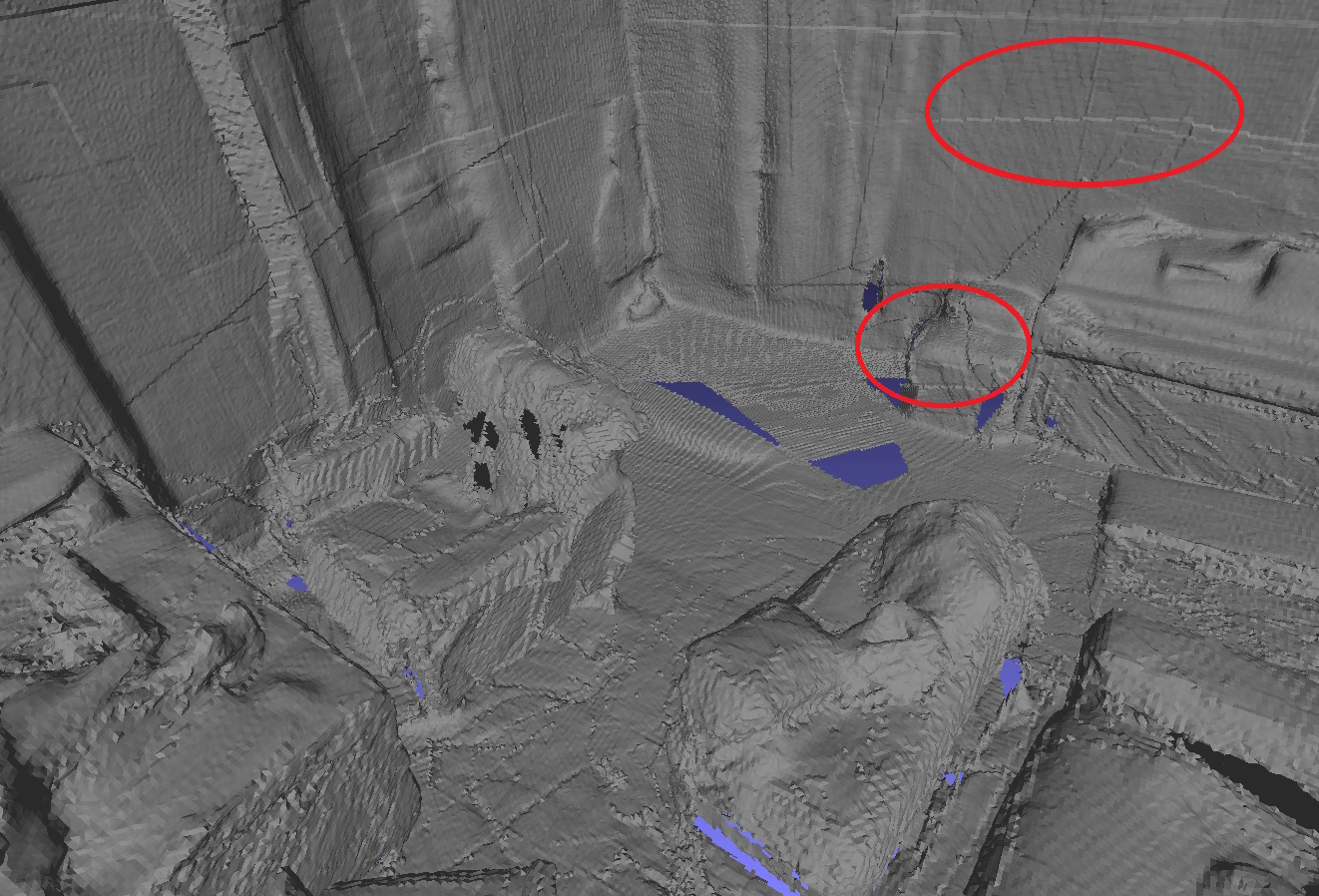}
        \caption{DPT, W8A8(CPU)}
        \label{fig:5-dpt_base_int8_cpu}
    \end{subfigure}
\centering
     \begin{subfigure}{0.45\linewidth}
    \includegraphics[width=1.0\linewidth]{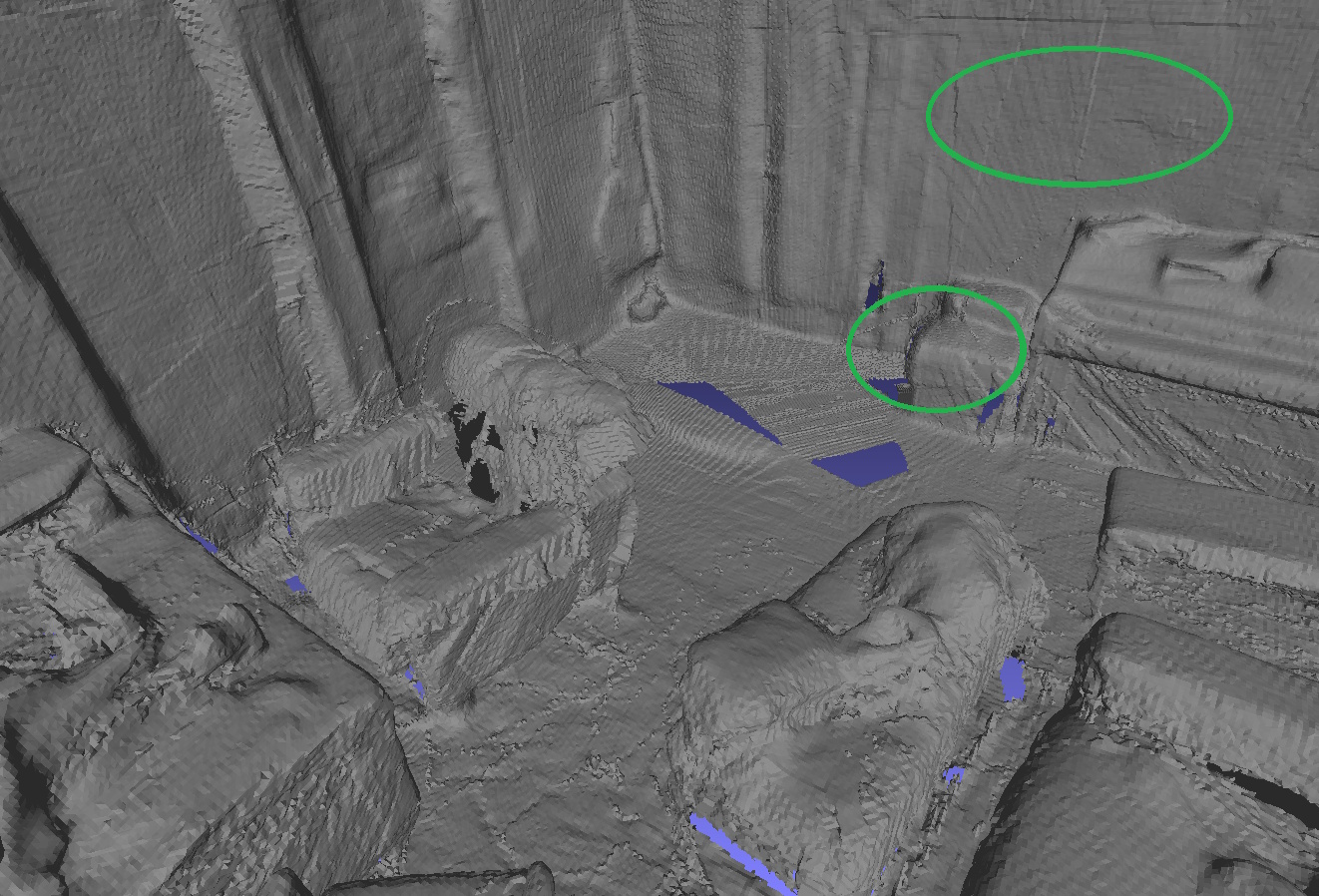}
        \caption{h2DPT, W8A8(CPU)}
        \label{fig:5-h2dpt_int8_cpu}
    \end{subfigure}    
    \caption{A view of 3D mesh fused with predicted depth maps by DPT and h2DPT for ScanNet scene scene0050\_02. Some reconstruction errors are highlighted by red and improved structures are marked by green.}
    \label{fig:5}
\end{figure}

In Fig.~\ref{fig:4} (Fig.~\ref{fig:5}), we show qualitative results of h2DispNet (h2DPT) model compared to the corresponding baseline variant. 3D meshes fused from depth maps predicted by FP32 h2DispNet (Fig.~\ref{fig:4-h2dispnet_fp32}) and FP32 h2DPT (Fig.~\ref{fig:5-h2dpt_fp32}) models have very similar structure and depth smoothness compared to the FP32 DispNet (Fig.~\ref{fig:4-dispnet_base_fp3212}) and FP32 DPT (Fig.~\ref{fig:5-dpt_base_fp32}). Both baseline and modified FP32 models' variants produce quality of reconstructed 3D mesh comparable to GT (Fig.~\ref{fig:3}) but with slightly smoother structure.

We observe two types of quantization artifacts present in fused meshes for models run on CPU delegate. The first is the noise on flat surfaces for original DispNet in W8A8 format (Fig.~\ref{fig:4-dispnet_base_int8_cpu}); the second is presence of visible edges of different frames' depth maps (step-like structures) in the mesh for original DPT in W8A8 format (Fig.~\ref{fig:5-dpt_base_int8_cpu}). We attribute the second type of artifacts to the systematic errors in depth prediction leading to errors in depth scale. Models modified according to the proposed method lead to mesh reconstruction with much reduced noise level (Figs.~\ref{fig:4-h2dispnet_int8_cpu} and \ref{fig:5-h2dpt_int8_cpu}). 

For the baseline models run on DSP delegate, we observe the same artifacts but more pronounced (Figs.~\ref{fig:4-dispnet_base_int8_dsp}--\ref{fig:5-dpt_base_int8_dsp}). The h2DispNet model in W8A8 format almost eliminates quantization artifacts (Fig.~\ref{fig:4-h2dispnet_int8_dsp}) and restores mesh spatial details. The h2DPT in W8A8 format model removes step-like artifacts and reduces noise for flat surfaces.

\section{Additional details of depth maps quality}
\label{sec:depth}

\begin{figure}[!htb]
    \centering
    \begin{subfigure}{0.3\linewidth}
        \includegraphics[width=1.0\linewidth]{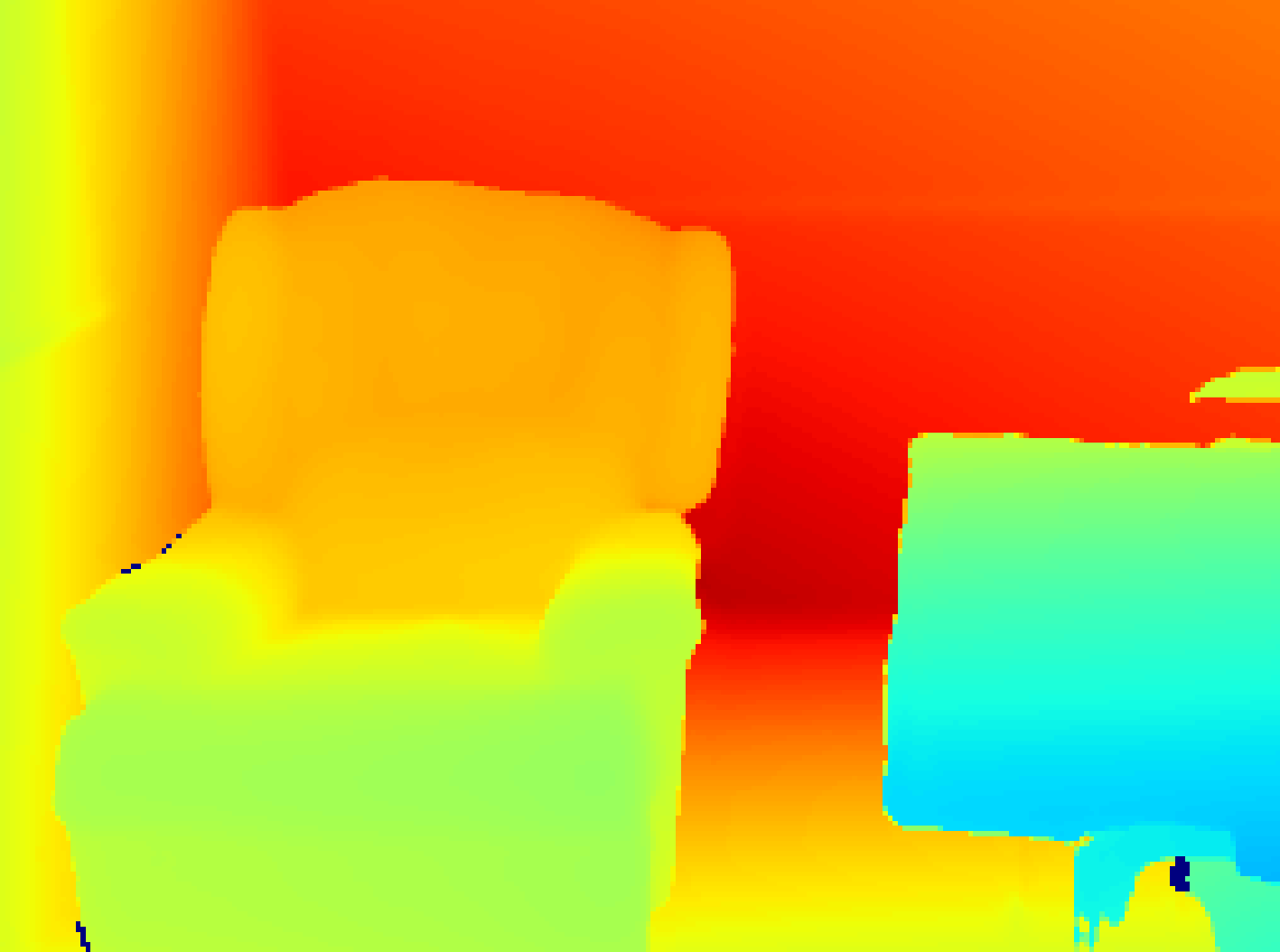}
        \caption{GT depth}
        \label{fig:gt_depth0}
    \end{subfigure}
    \centering
    \begin{subfigure}{0.3\linewidth}
    \includegraphics[width=1.0\linewidth]{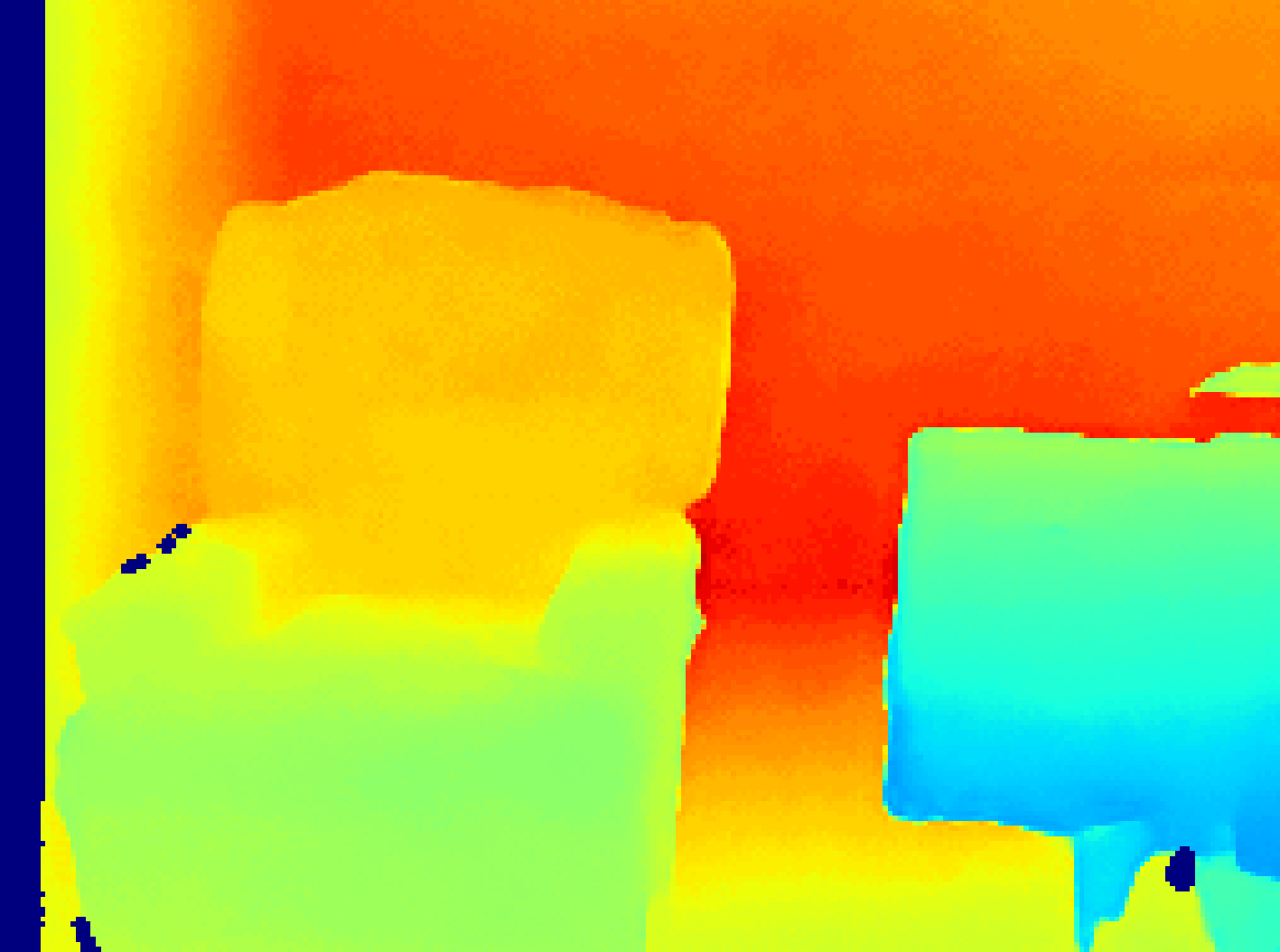}
        \caption{DPT depth}
        \label{fig:dptbase0}
    \end{subfigure}
    \centering
    \begin{subfigure}{0.355\linewidth}
    \includegraphics[width=1.0\linewidth]{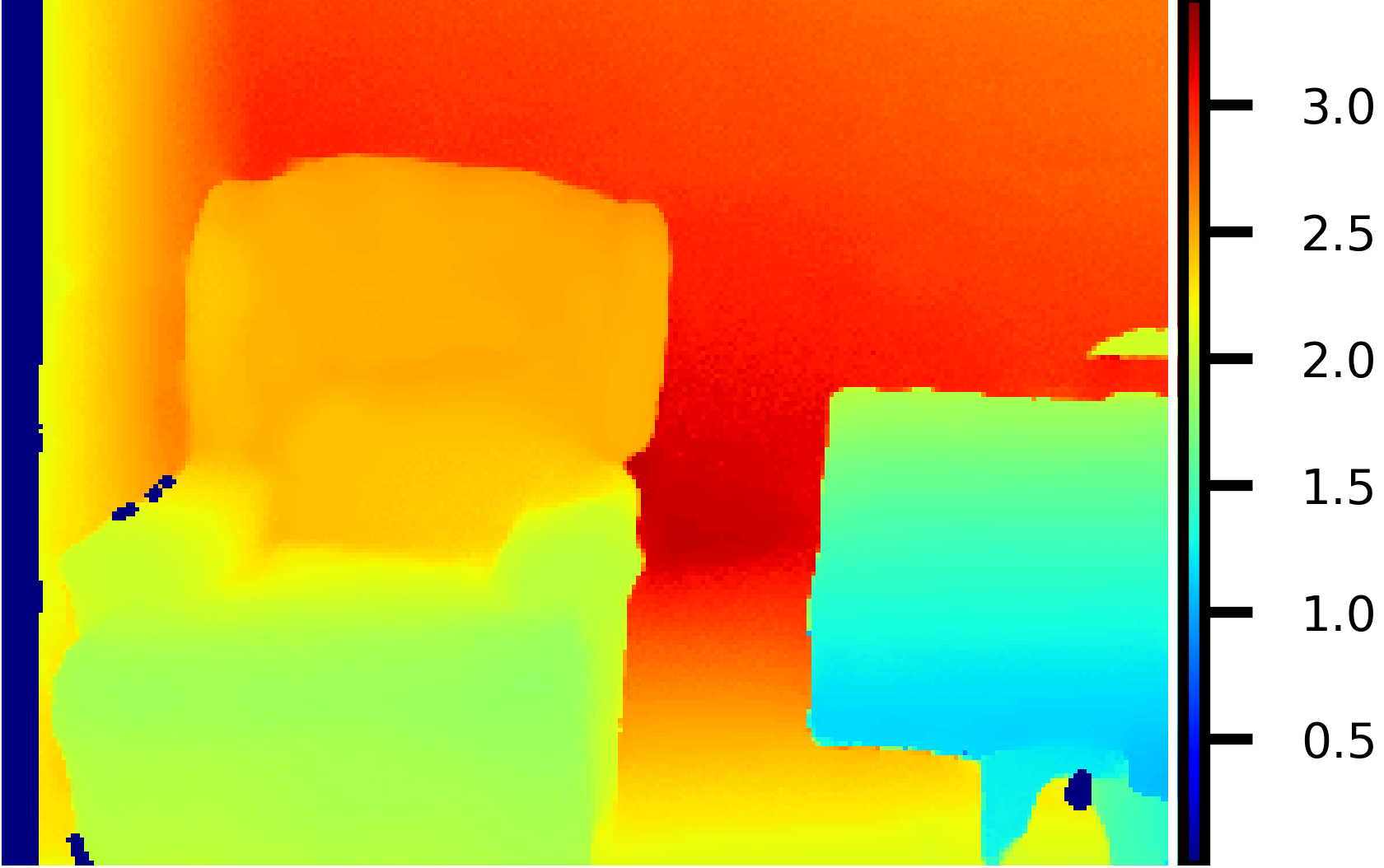}
        \caption{h3DPT depth$\phantom{xxx}$}
        \label{fig:h3dpt0}
    \end{subfigure}
    \centering 
   \begin{subfigure}{0.30\linewidth}
        \includegraphics[width=1.0\linewidth]{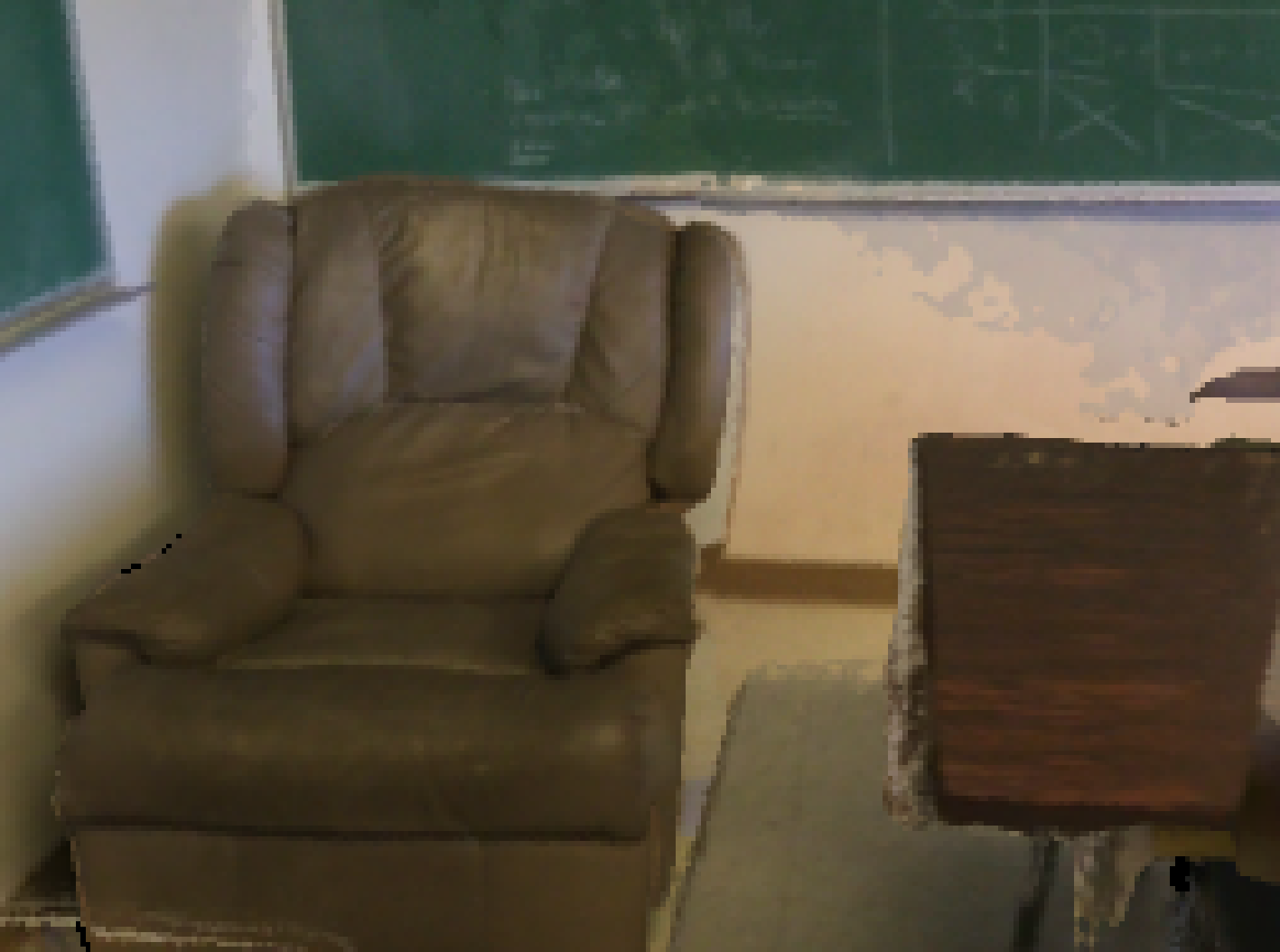}
        \caption{Image}
        \label{image0}
    \end{subfigure}
    \centering
    \begin{subfigure}{0.3\linewidth}
    \includegraphics[width=1.0\linewidth]{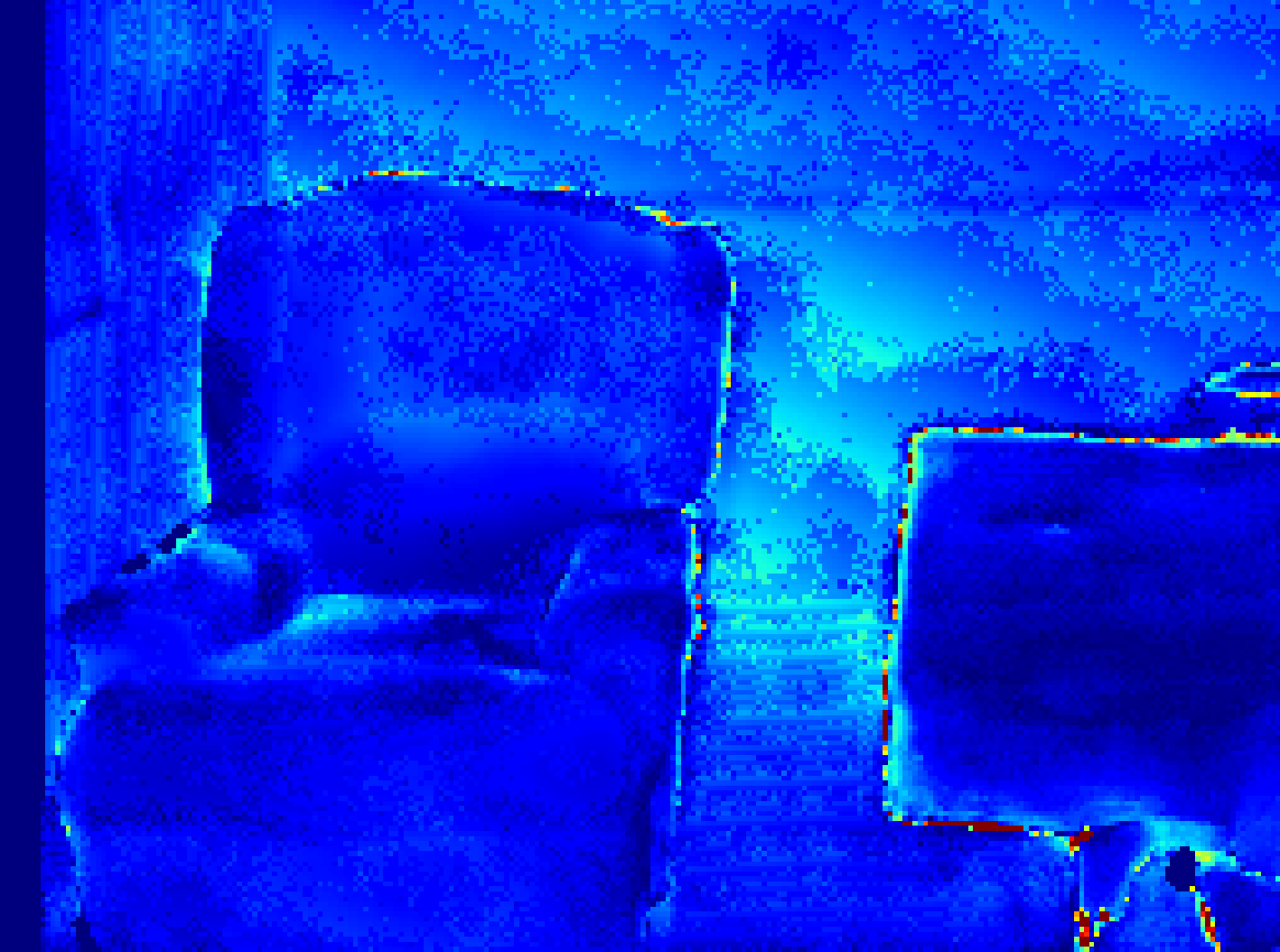}
    \caption{$|\text{GT - DPT}|$}
    \label{fig:gt_minus_dpt0}
    \end{subfigure}
\centering
    \begin{subfigure}{0.355\linewidth}
    \includegraphics[width=1.0\linewidth]{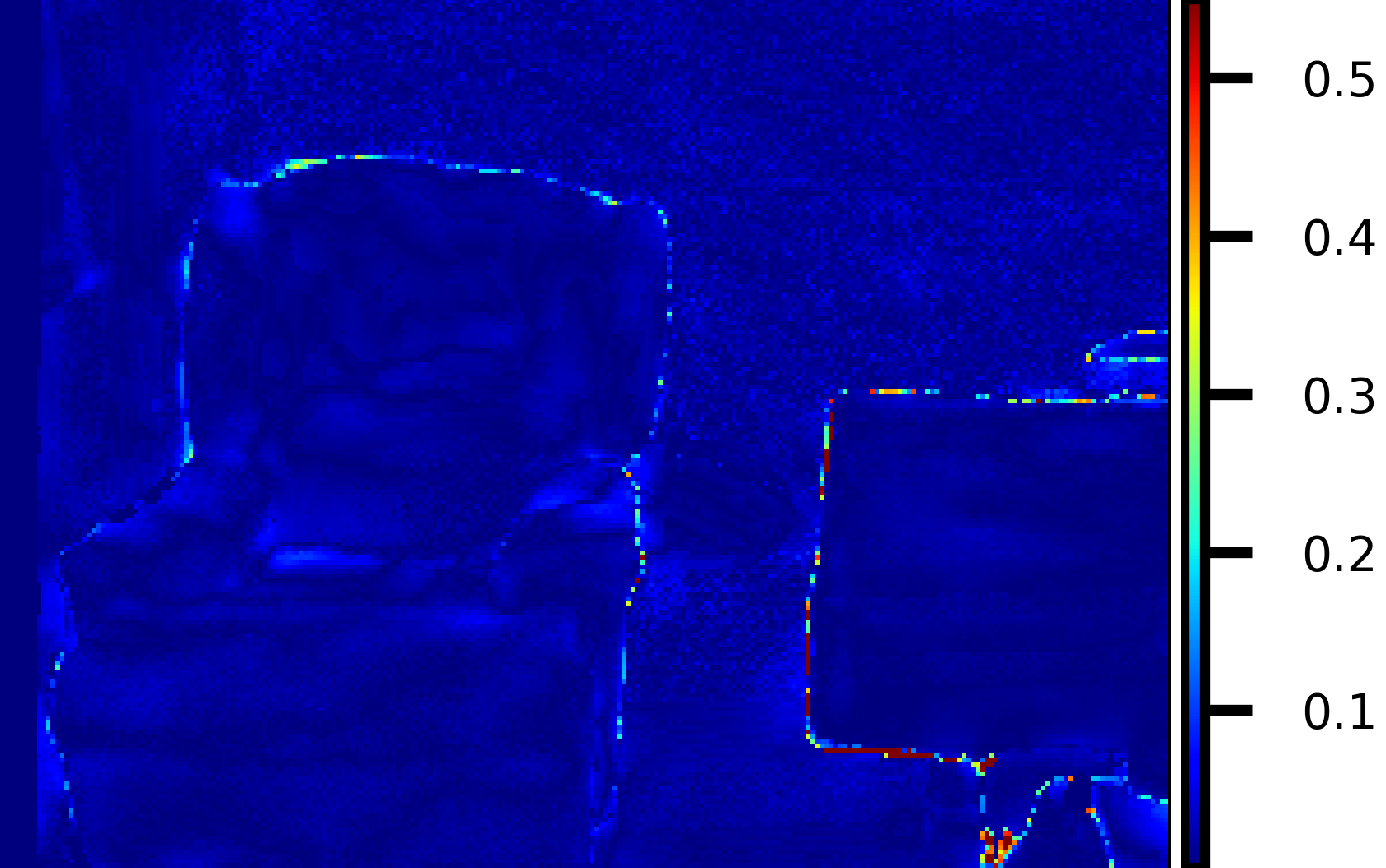}
    \caption{$|\text{GT - h3DPT}|\phantom{xxx}$}
    \label{fig:gt-h3dpt0}
    \end{subfigure}
    \caption{Depth errors of DPT and h3DPT models on DSP. ScanNet scene scene0030\_02. All values are presented in meters.}
    \label{fig:depth_errors0}
\end{figure}

\begin{figure}[!htb]
    \centering
    \begin{subfigure}{0.3\linewidth}
        \includegraphics[width=1.0\linewidth]{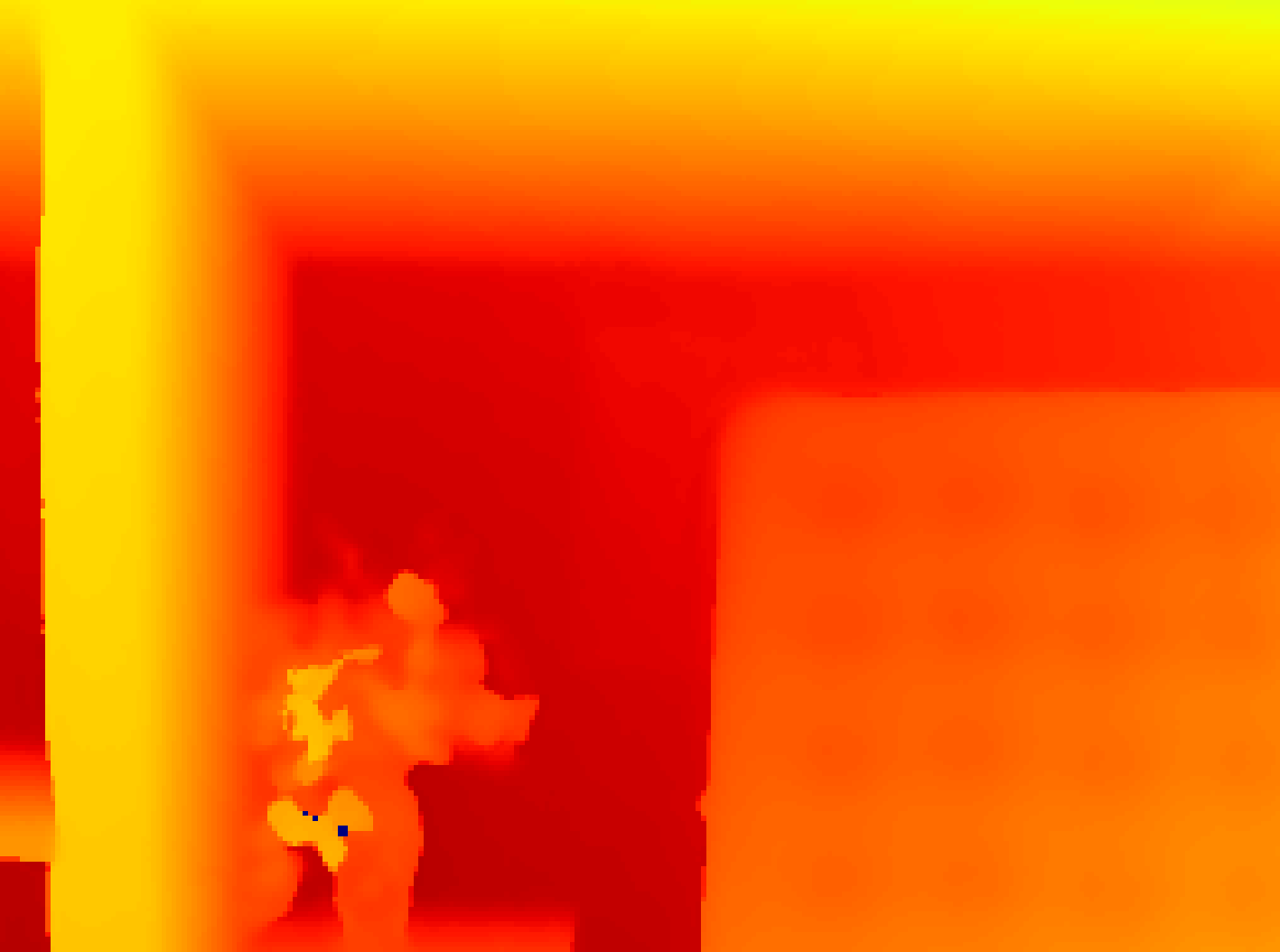}
        \caption{GT depth}
        \label{fig:gt_depth2}
    \end{subfigure}
    \centering
    \begin{subfigure}{0.3\linewidth}
    \includegraphics[width=1.0\linewidth]{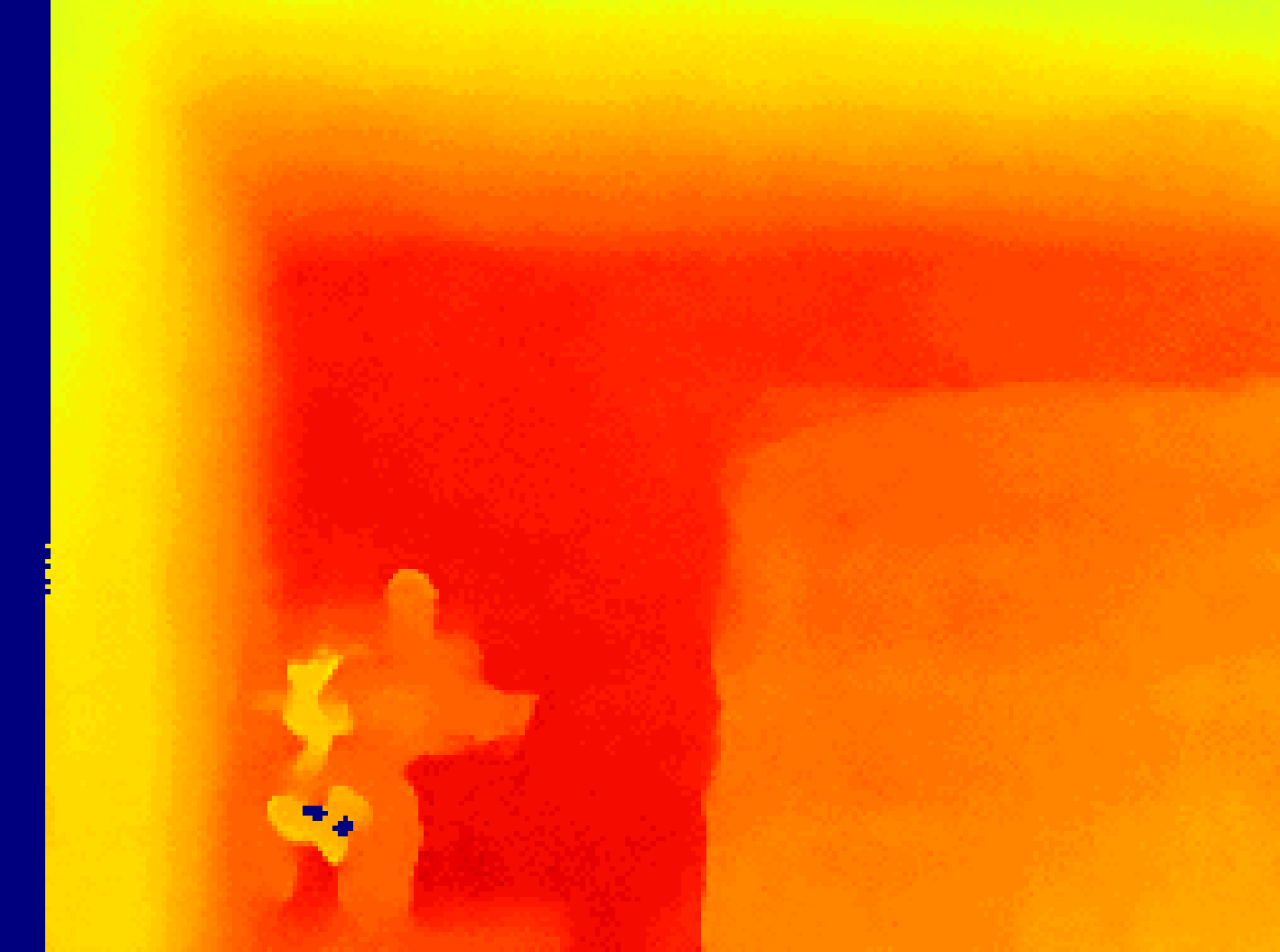}
        \caption{DPT depth}
        \label{fig:dptbase2}
    \end{subfigure}
    \centering
    \begin{subfigure}{0.355\linewidth}
    \includegraphics[width=1.0\linewidth]{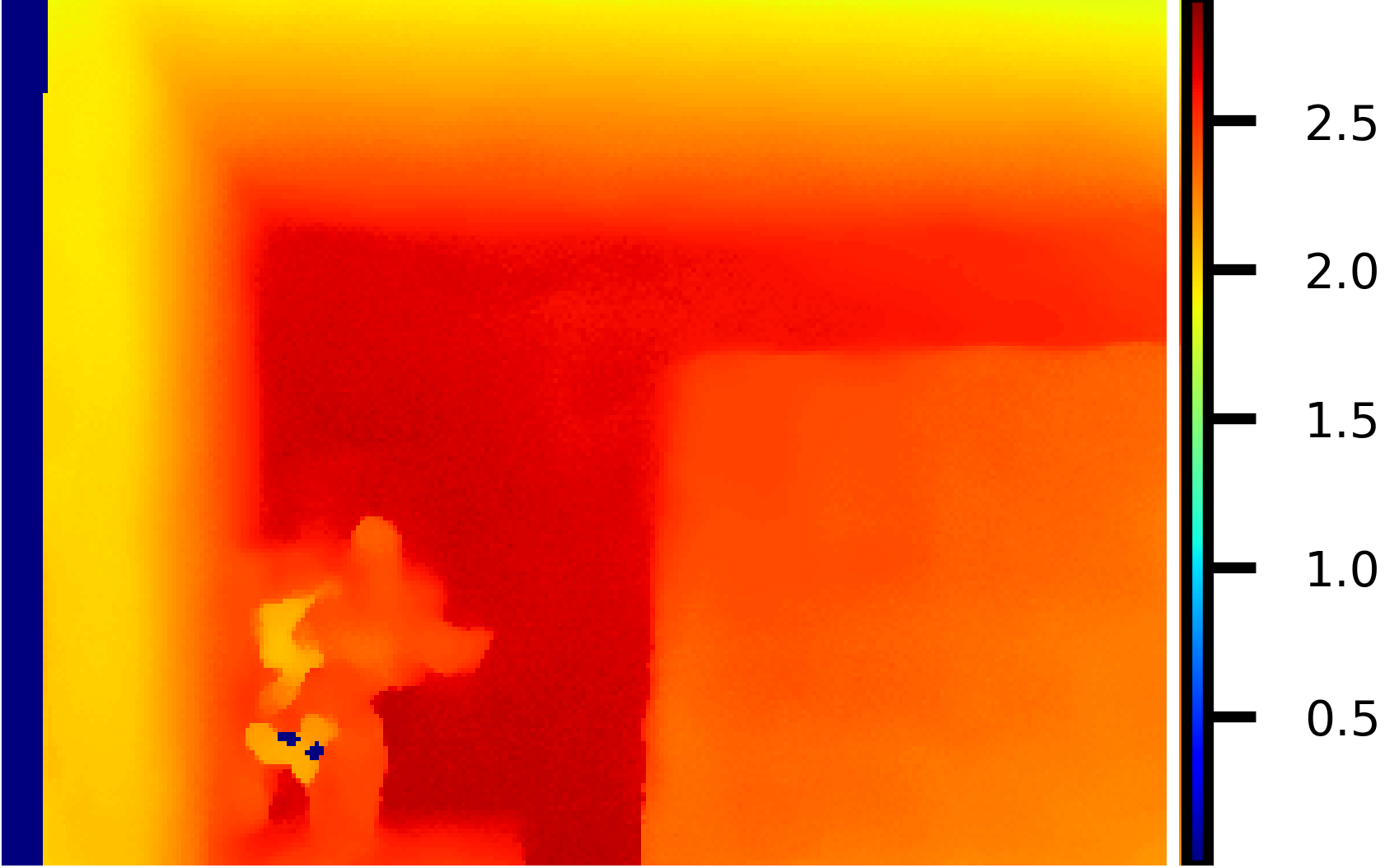}
        \caption{h3DPT depth$\phantom{xxx}$}
        \label{fig:h3dpt2}
    \end{subfigure}
    \centering 
   \begin{subfigure}{0.30\linewidth}
        \includegraphics[width=1.0\linewidth]{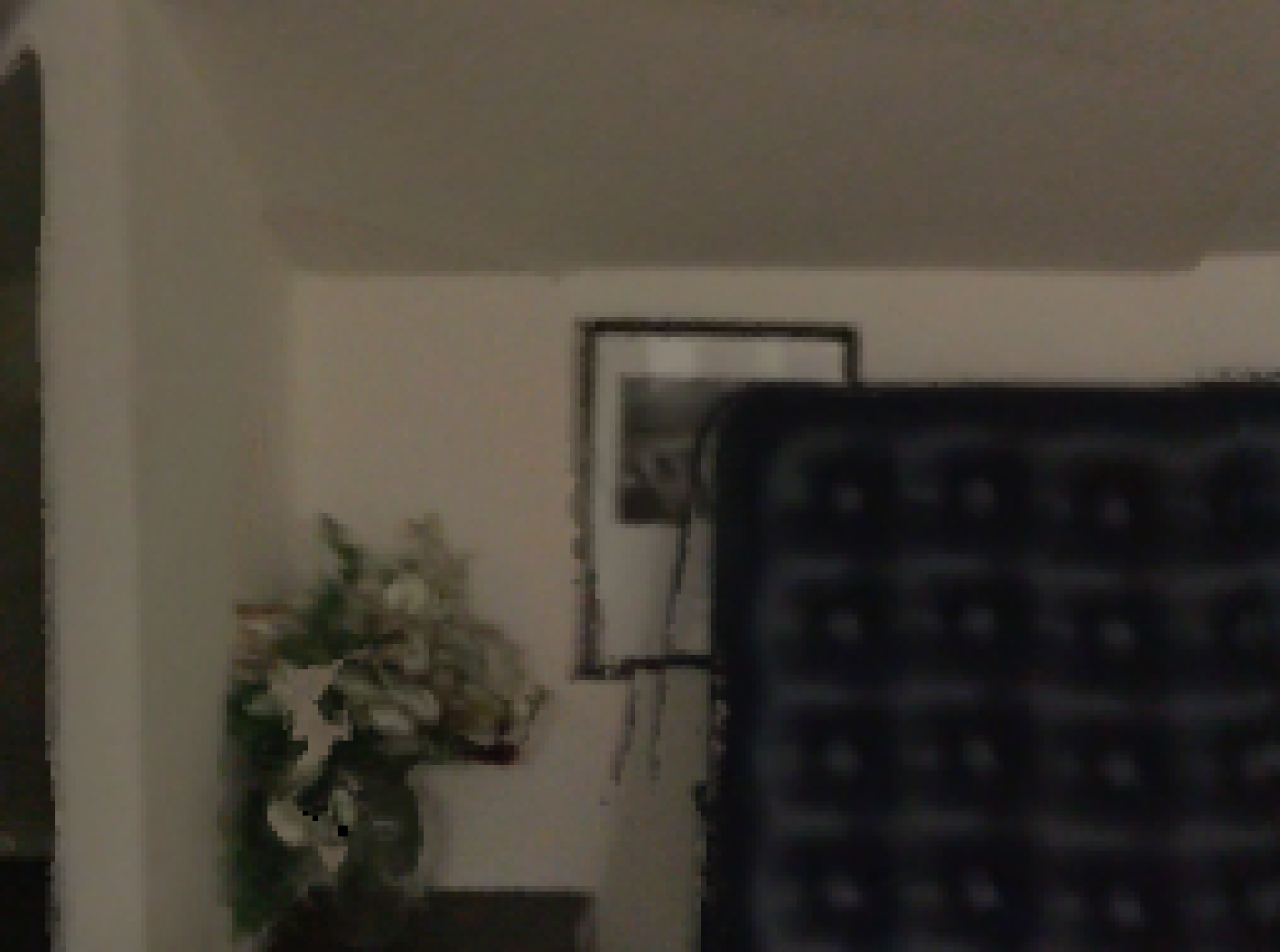}
        \caption{Image}
        \label{image2}
    \end{subfigure}
    \centering
    \begin{subfigure}{0.3\linewidth}
    \includegraphics[width=1.0\linewidth]{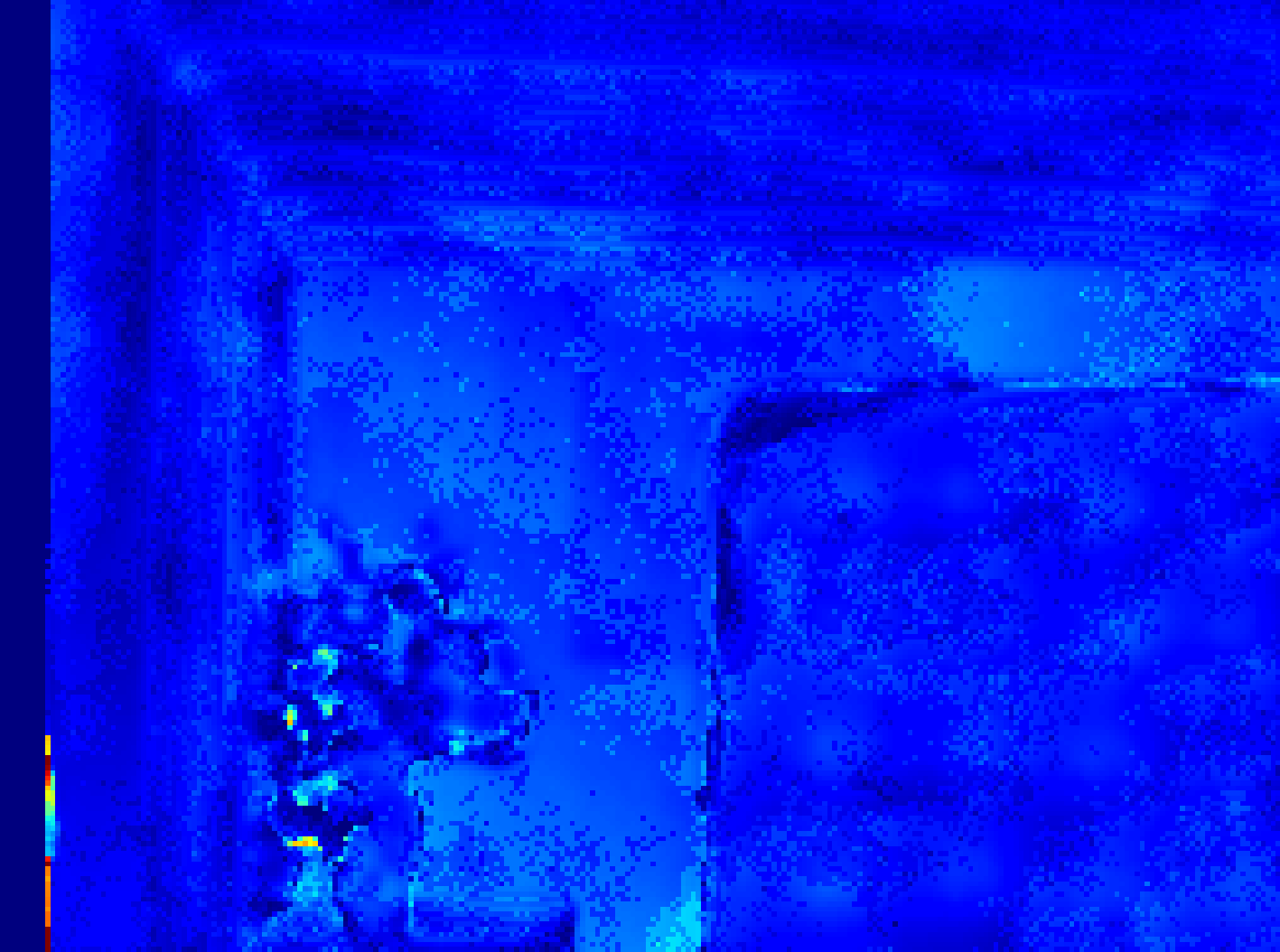}
    \caption{$|\text{GT - DPT}|$}
    \label{fig:gt_minus_dpt2}
    \end{subfigure}
\centering
    \begin{subfigure}{0.355\linewidth}
    \includegraphics[width=1.0\linewidth]{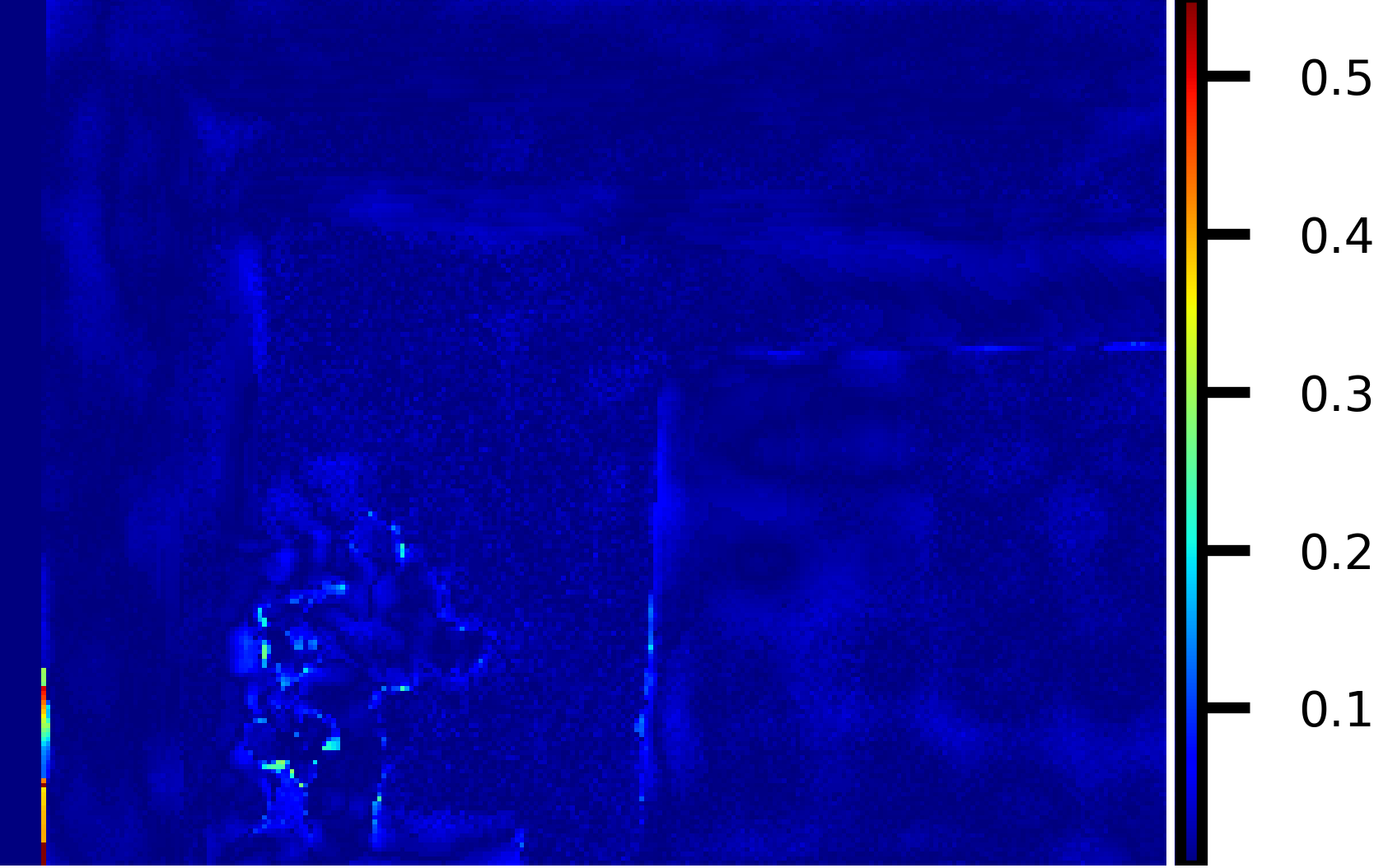}
    \caption{$|\text{GT - h3DPT}|\phantom{xxx}$}
    \label{fig:gt-h3dpt2}
    \end{subfigure}
    \caption{Depth errors of DPT and h3DPT models on DSP. ScanNet scene scene0629\_00. All values are presented in meters.}
    \label{fig:depth_errors2}
\end{figure}

\begin{figure}[!htb]
    \centering
    \begin{subfigure}{0.3\linewidth}
        \includegraphics[width=1.0\linewidth]{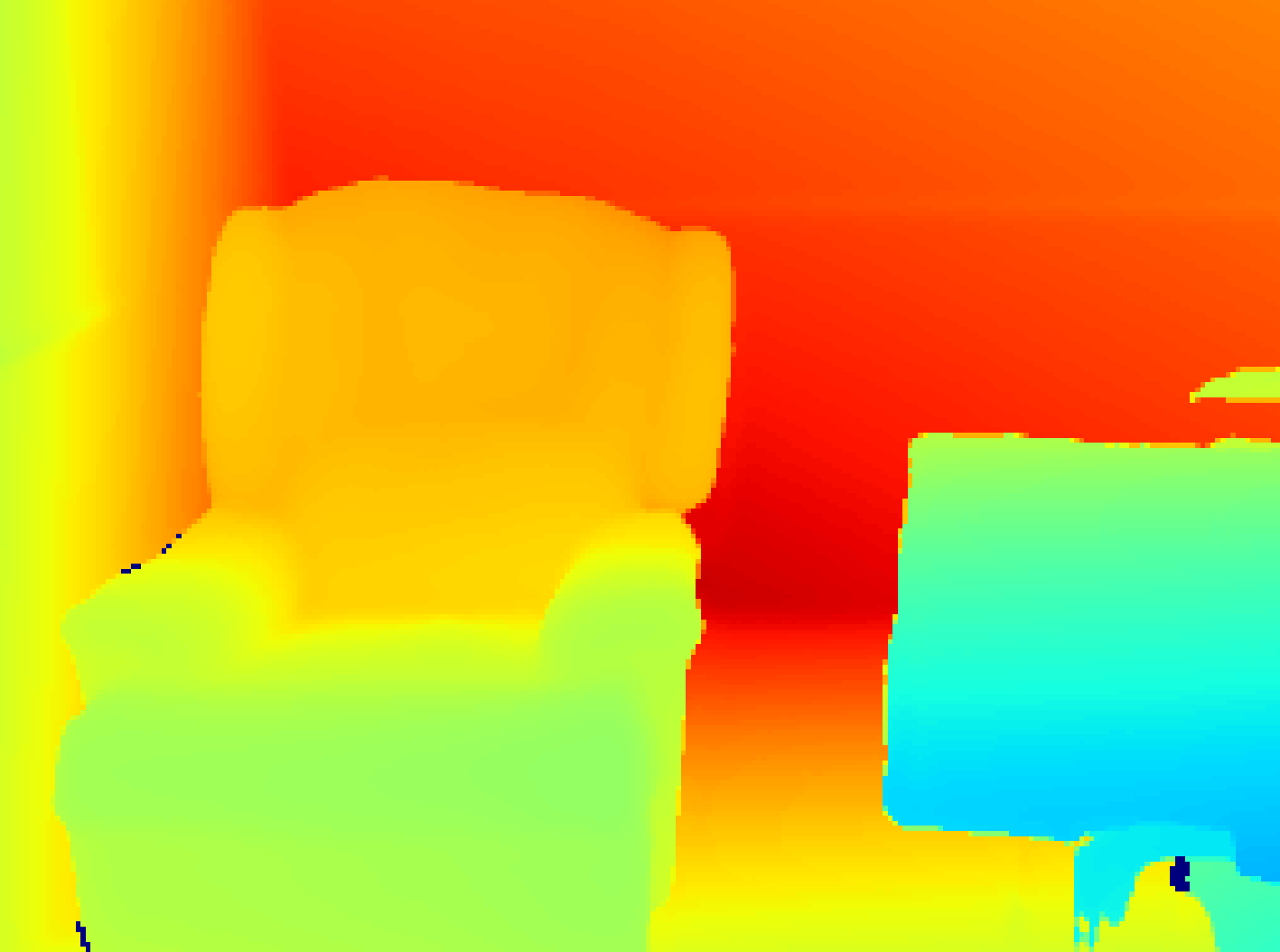}
        \caption{GT depth}
        \label{fig:gt_depth0_30}
    \end{subfigure}
    \centering
    \begin{subfigure}{0.3\linewidth}
    \includegraphics[width=1.0\linewidth]{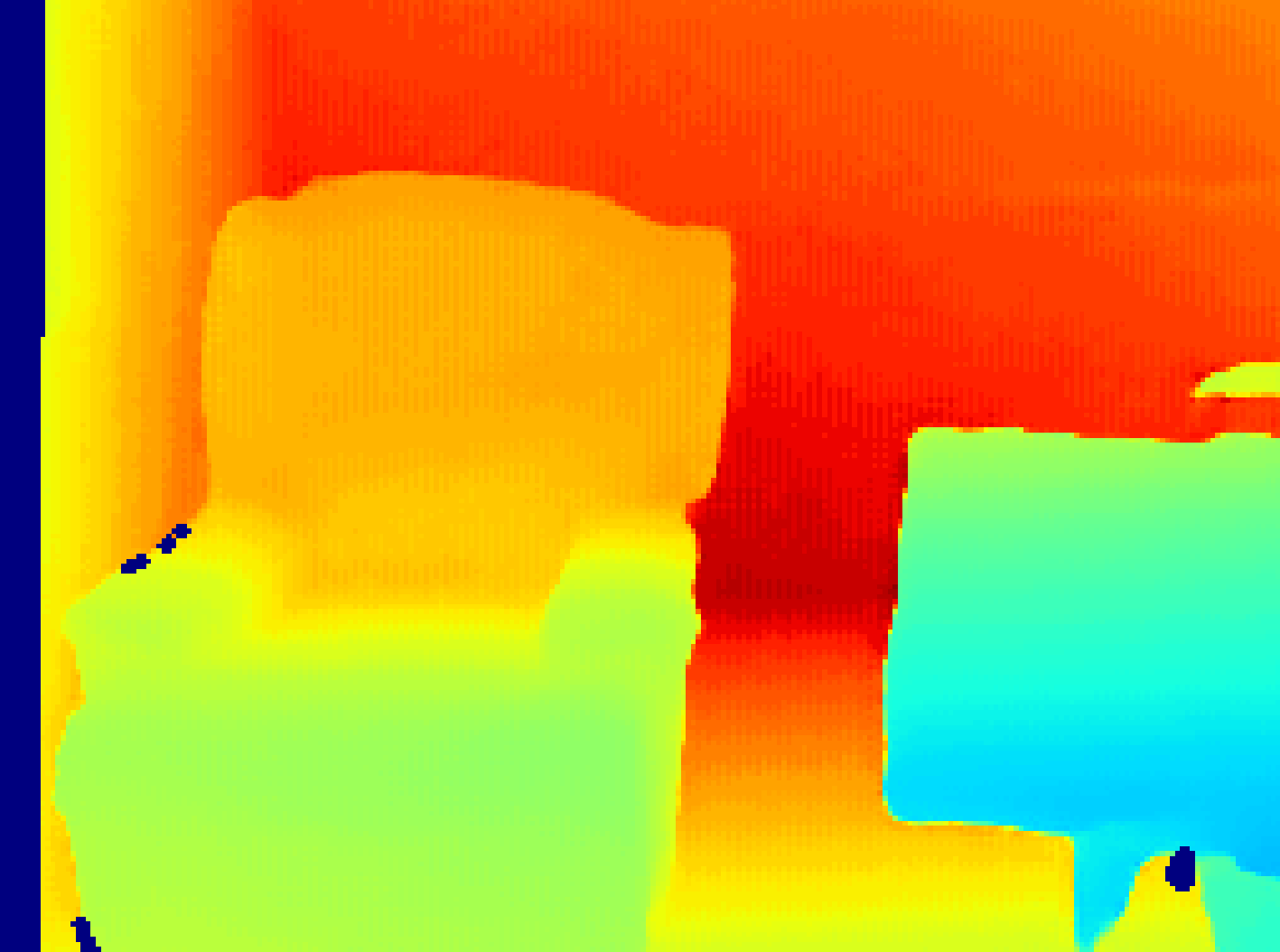}
        \caption{DispNet depth}
        \label{fig:dispnetbase0}
    \end{subfigure}
    \centering
    \begin{subfigure}{0.355\linewidth}
    \includegraphics[width=1.0\linewidth]{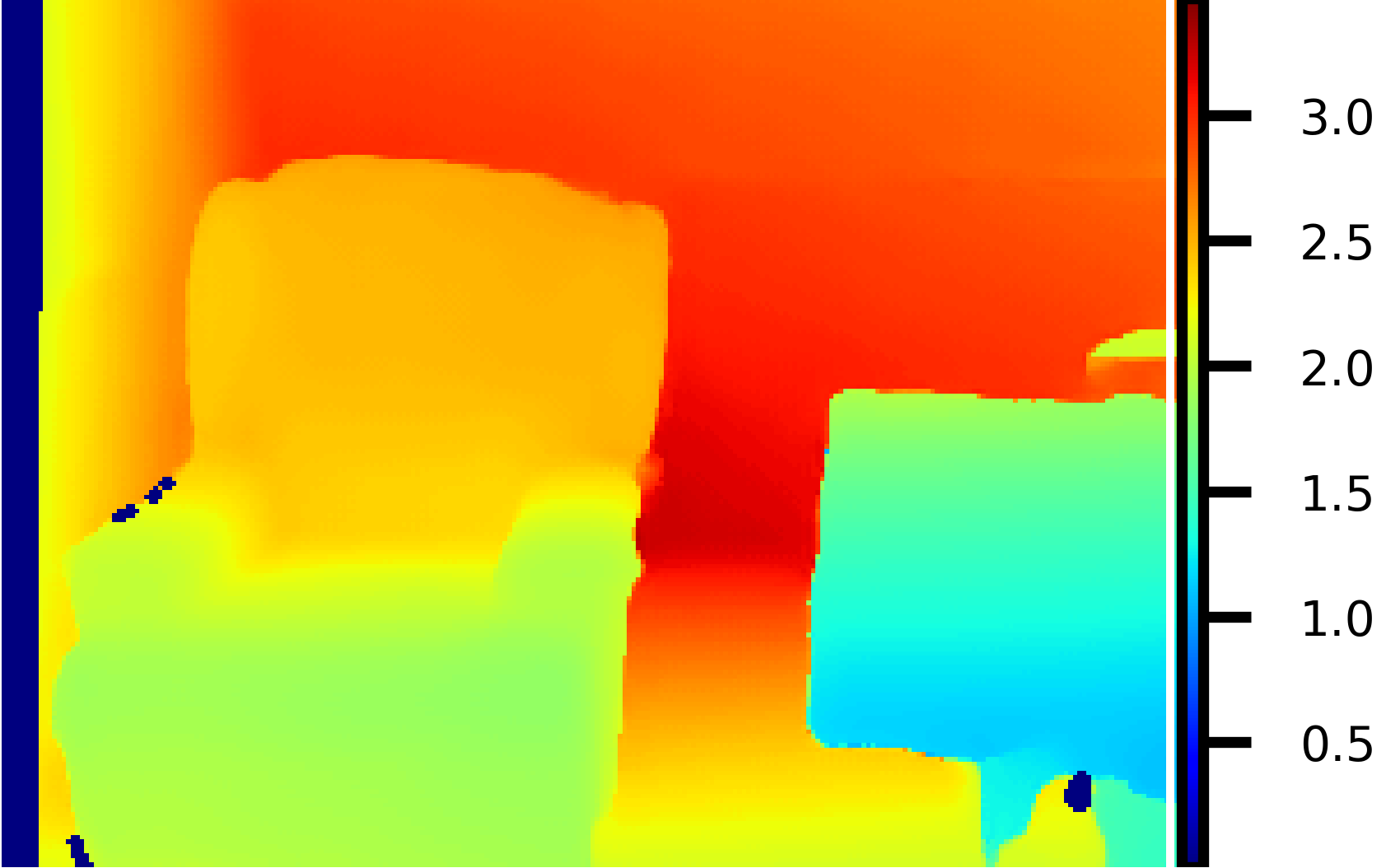}
        \caption{h3DispNet depth$\phantom{xxx}$}
        \label{fig:h3dfs0}
    \end{subfigure}
    \centering 
   \begin{subfigure}{0.30\linewidth}
        \includegraphics[width=1.0\linewidth]{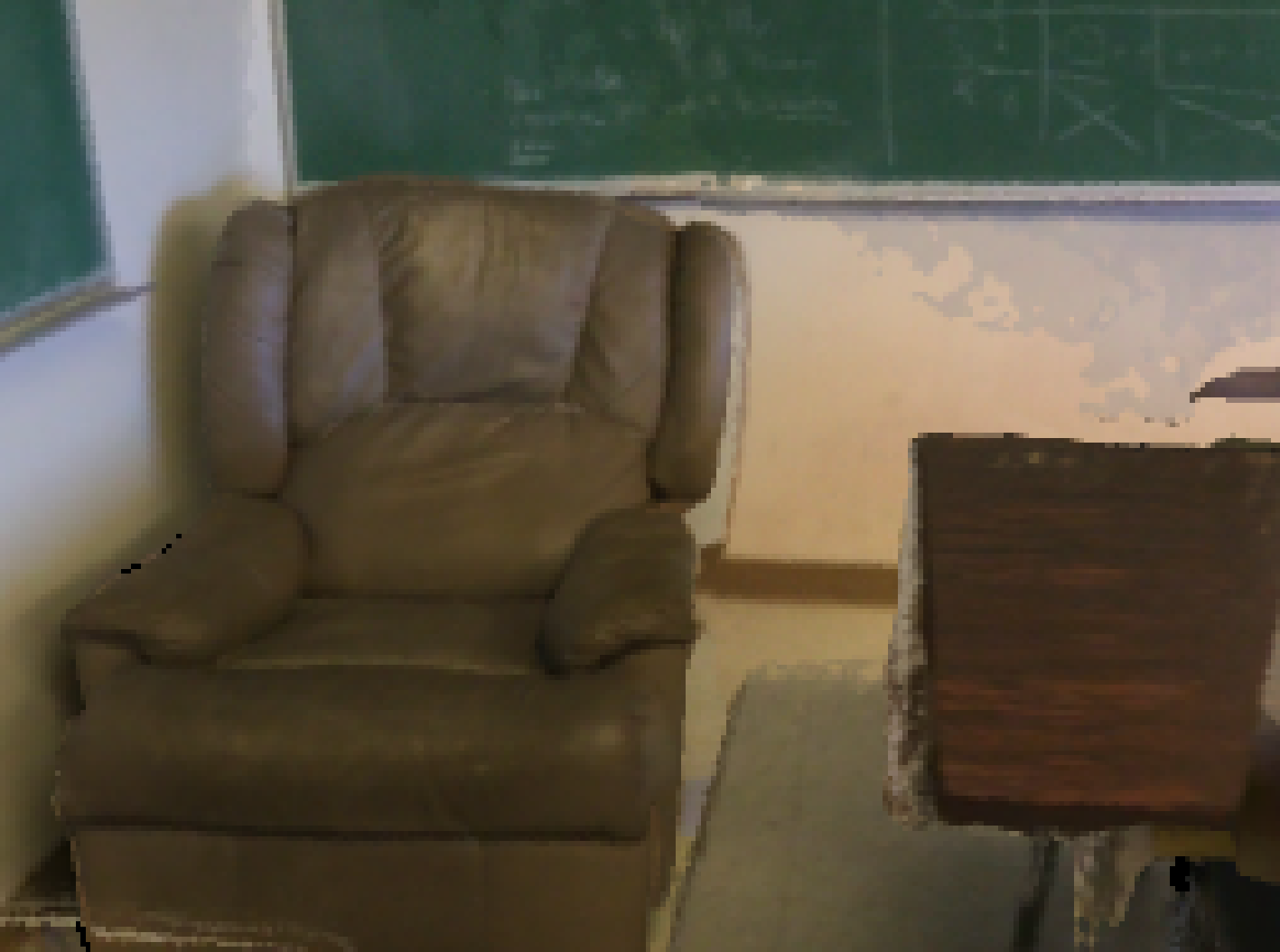}
        \caption{Image}
        \label{image0_v2}
    \end{subfigure}
    \centering
    \begin{subfigure}{0.3\linewidth}
    \includegraphics[width=1.0\linewidth]{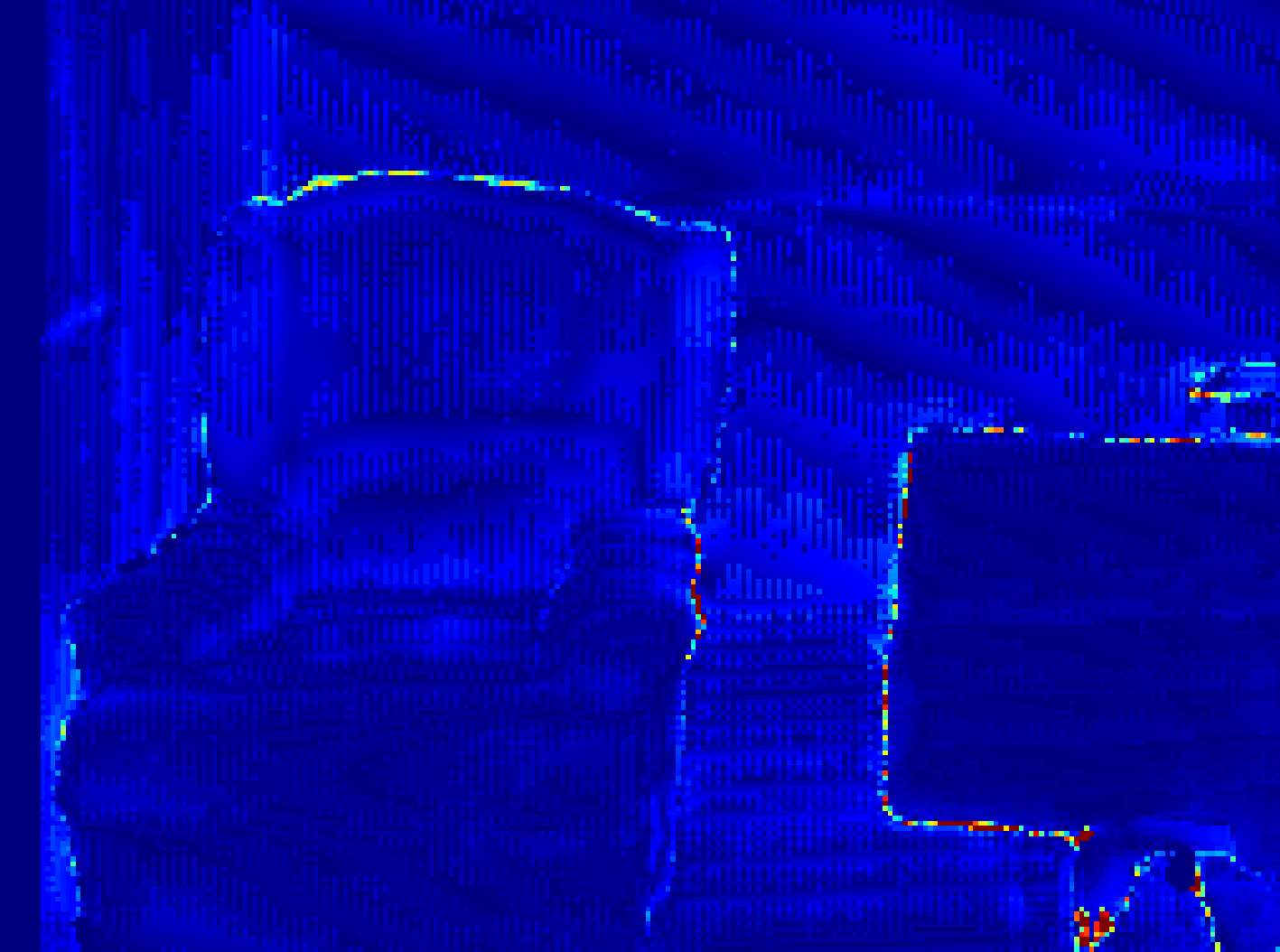}
    \caption{$|\text{GT - DispNet}|$}
    \label{fig:gt_minus_dfs0}
    \end{subfigure}
\centering
    \begin{subfigure}{0.355\linewidth}
    \includegraphics[width=1.0\linewidth]{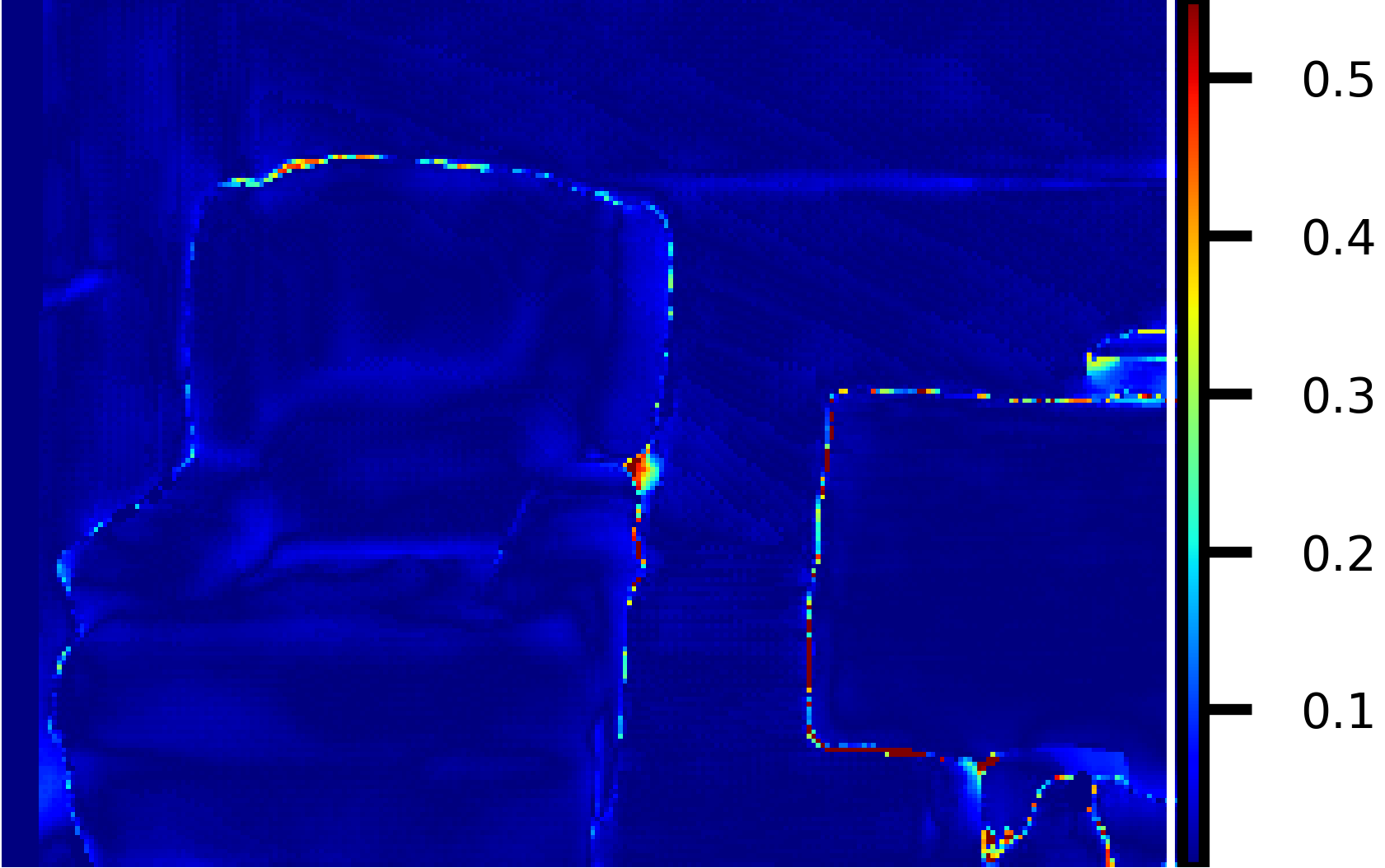}
    \caption{$|\text{GT - h3DispNet}|\phantom{xxx}$}
    \label{fig:gt-h3dfs0}
    \end{subfigure}
    \caption{Depth errors of DispNet and h3DispNet models on DSP. ScanNet scene scene0030\_02. All values are presented in meters.}
    \label{fig:depth_errorsdfs0}
\end{figure}

\begin{figure}[!htb]
    \centering
    \begin{subfigure}{0.3\linewidth}
        \includegraphics[width=1.0\linewidth]{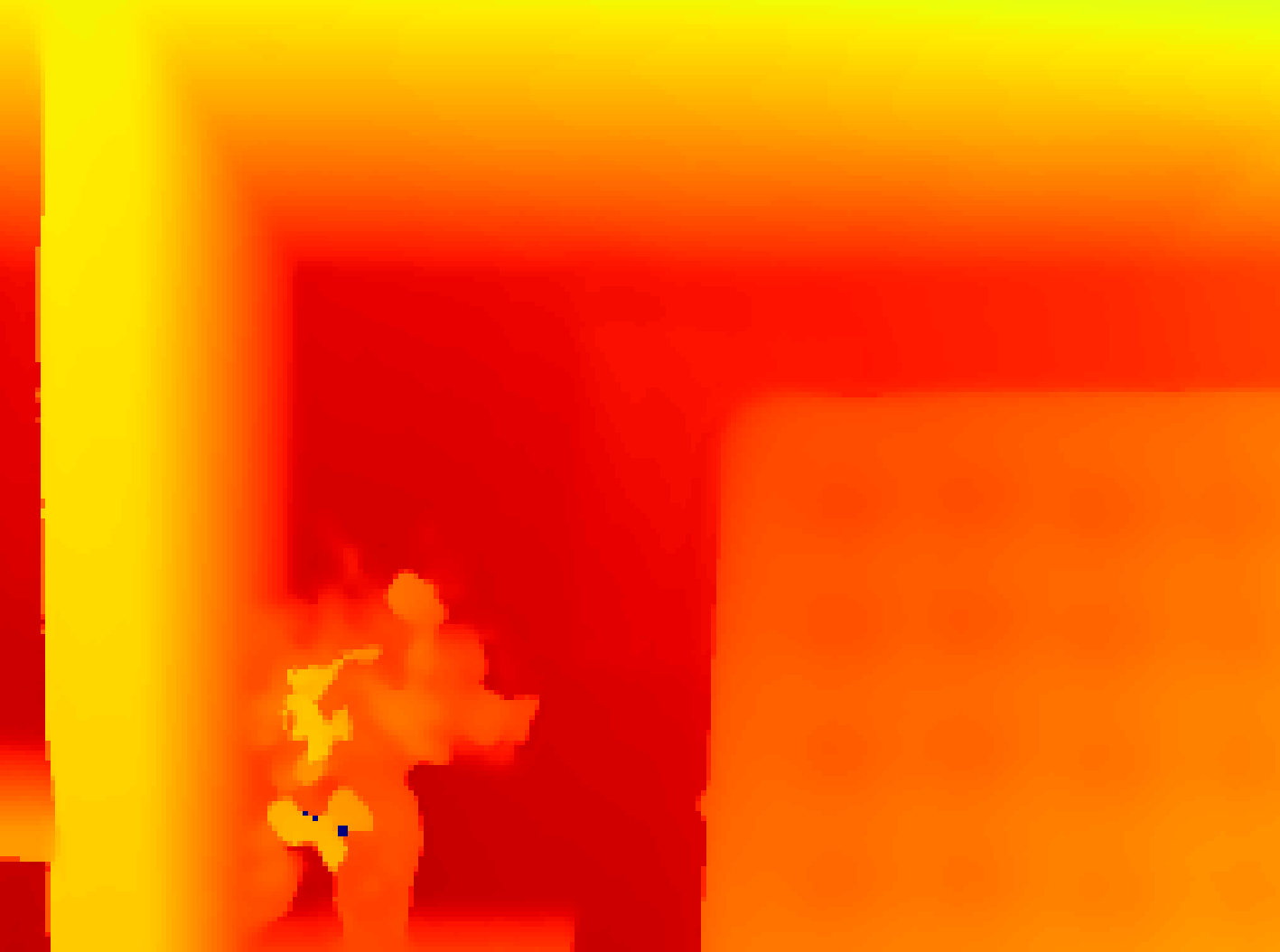}
        \caption{GT depth}
        \label{fig:gt_depth2v2}
    \end{subfigure}
    \centering
    \begin{subfigure}{0.3\linewidth}
    \includegraphics[width=1.0\linewidth]{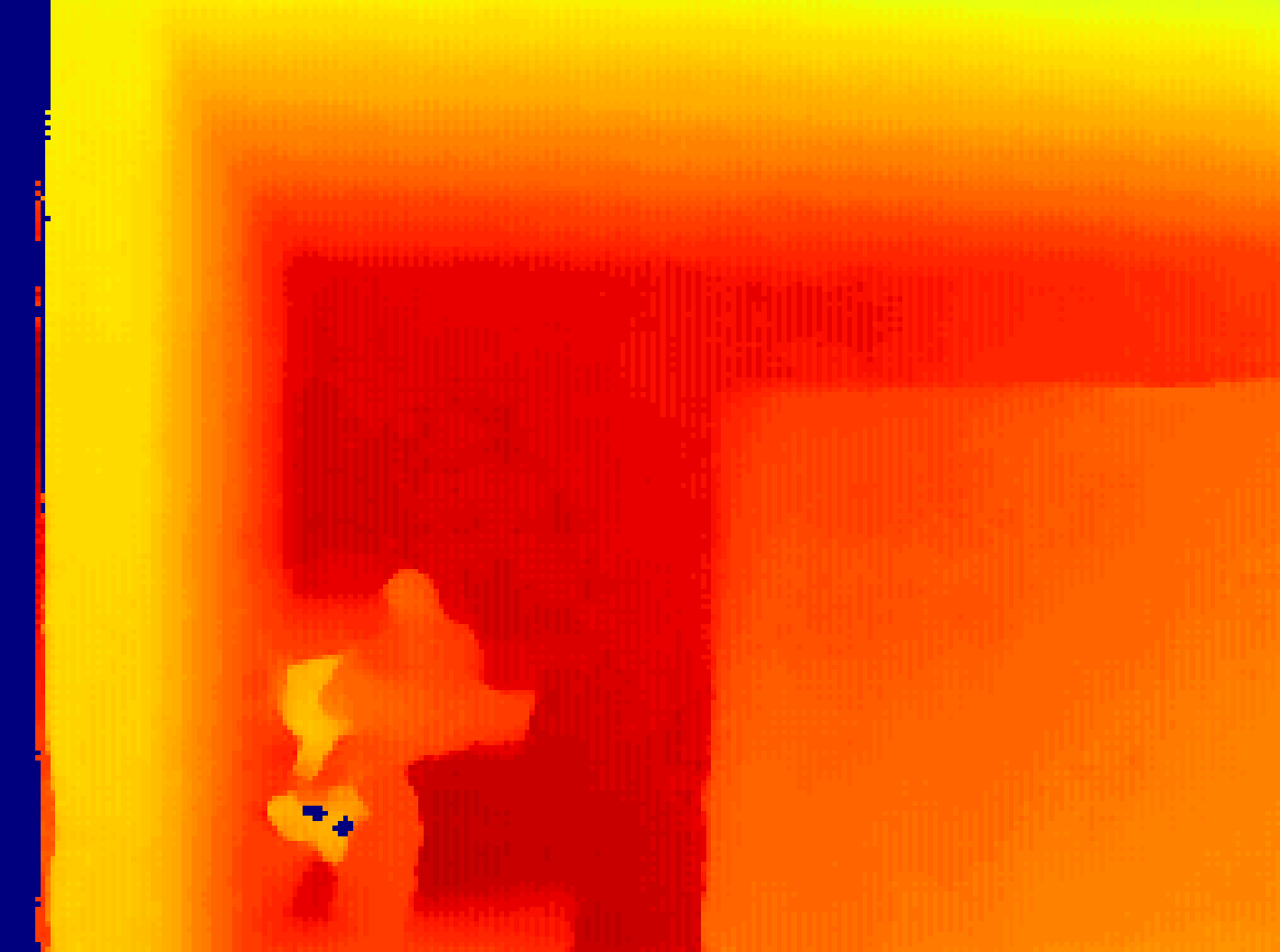}
        \caption{DispNet depth}
        \label{fig:dispnetbase2}
    \end{subfigure}
    \centering
    \begin{subfigure}{0.355\linewidth}
    \includegraphics[width=1.0\linewidth]{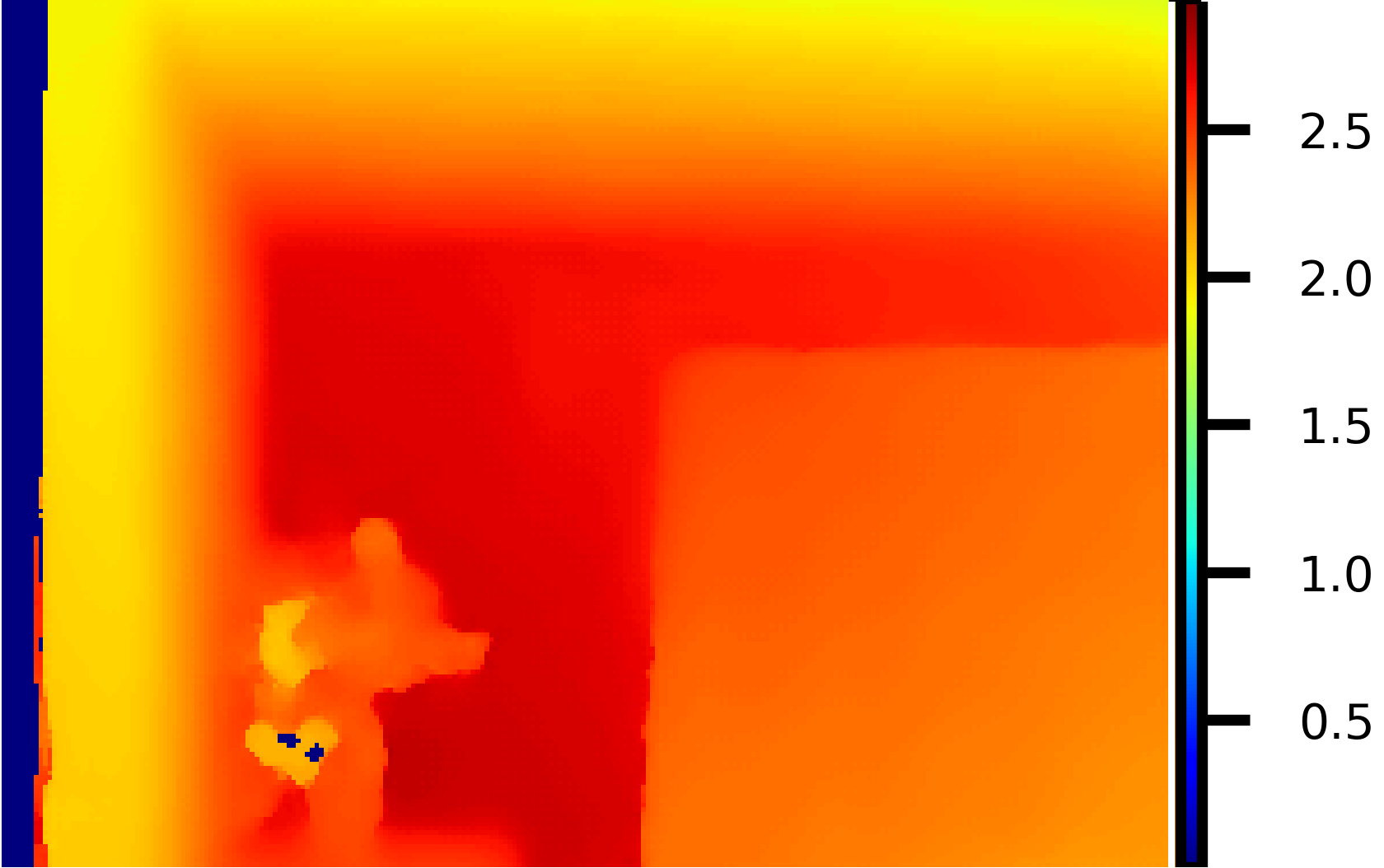}
        \caption{h3DispNet depth$\phantom{xxx}$}
        \label{fig:h3dfs2}
    \end{subfigure}
    \centering 
   \begin{subfigure}{0.30\linewidth}
        \includegraphics[width=1.0\linewidth]{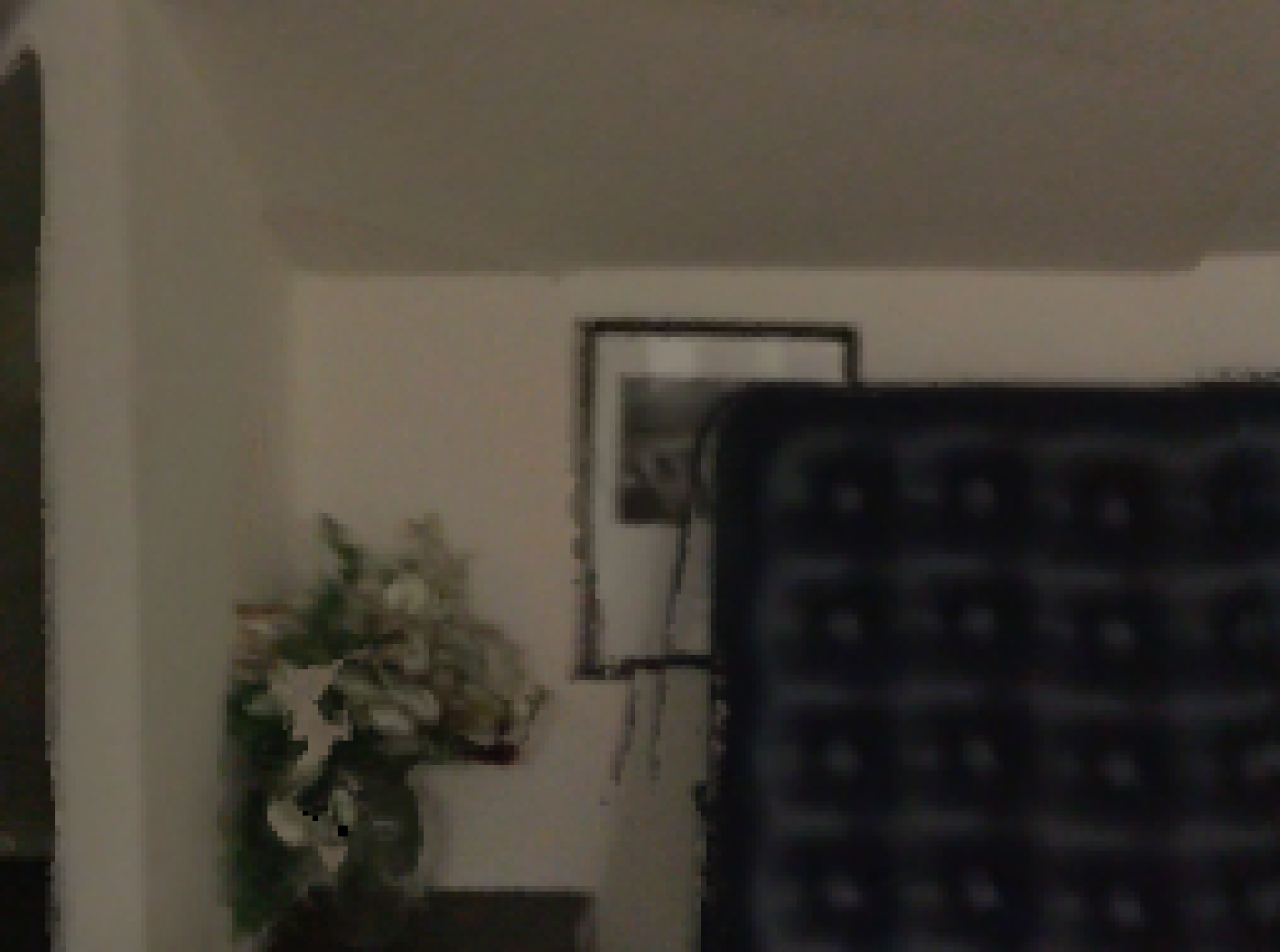}
        \caption{Image}
        \label{image0_30}
    \end{subfigure}
    \centering
    \begin{subfigure}{0.3\linewidth}
    \includegraphics[width=1.0\linewidth]{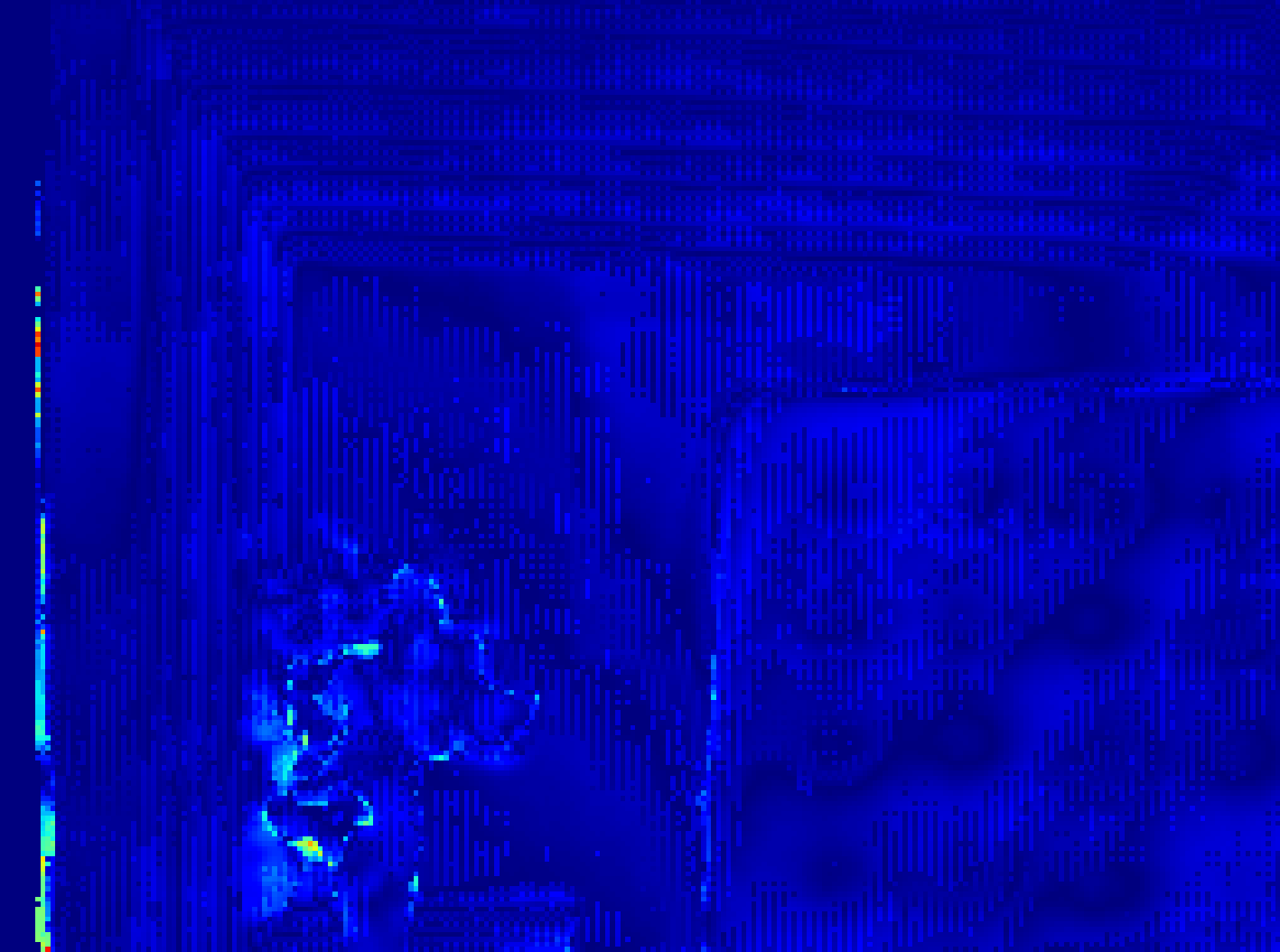}
    \caption{$|\text{GT - DispNet}|$}
    \label{fig:gt_minus_dfs2}
    \end{subfigure}
\centering
    \begin{subfigure}{0.355\linewidth}
    \includegraphics[width=1.0\linewidth]{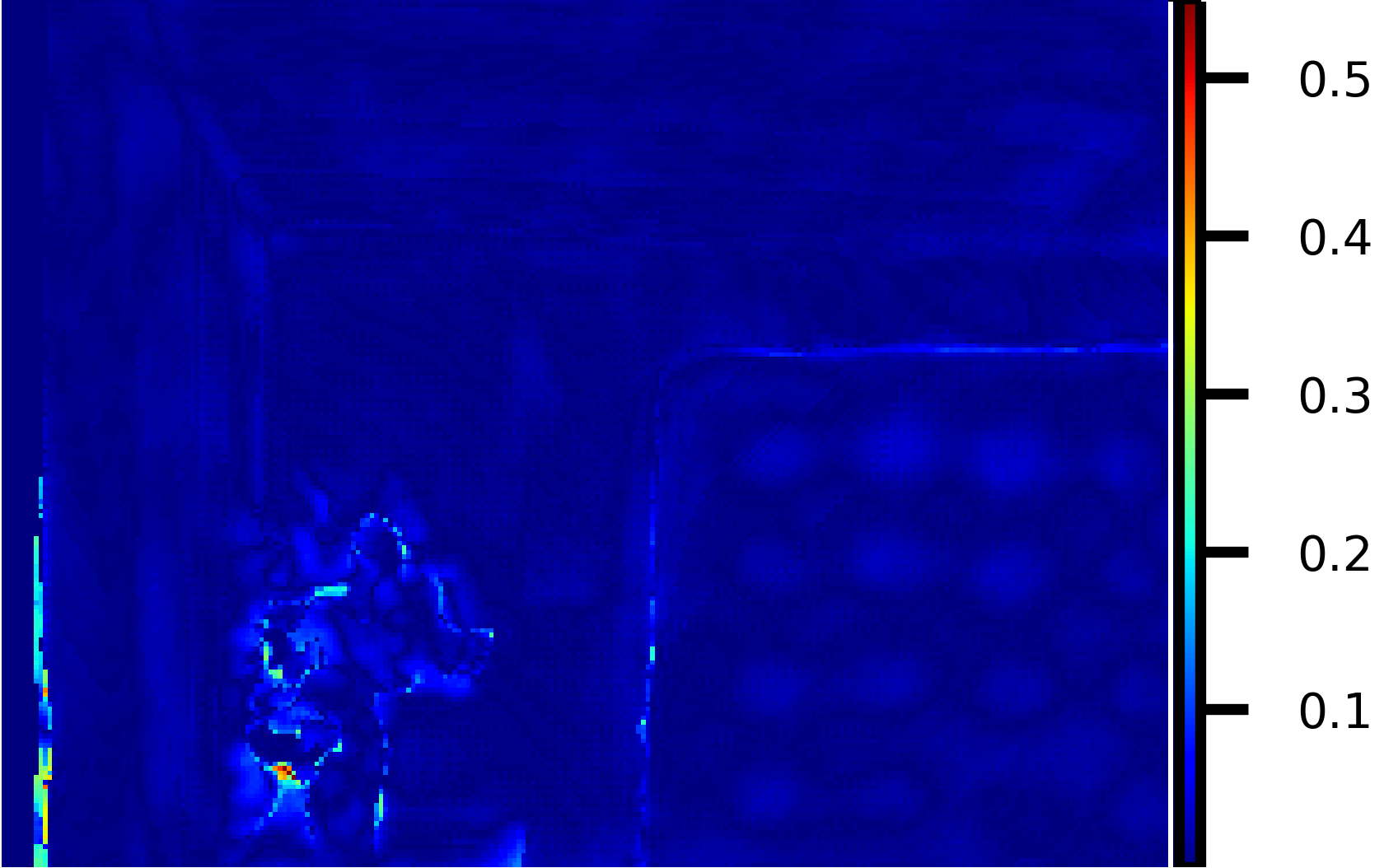}
    \caption{$|\text{GT - h3DispNet}|\phantom{xxx}$}
    \label{fig:gt-h3dfs2}
    \end{subfigure}
    \caption{Depth errors of DispNet and h3DispNet models on DSP. ScanNet scene scene0629\_00. All values are presented in meters.}
    \label{fig:depth_errorsdfs2}
\end{figure}

\begin{figure}[!htb]
    \centering
    \begin{subfigure}{0.3\linewidth}
        \includegraphics[width=1.0\linewidth]{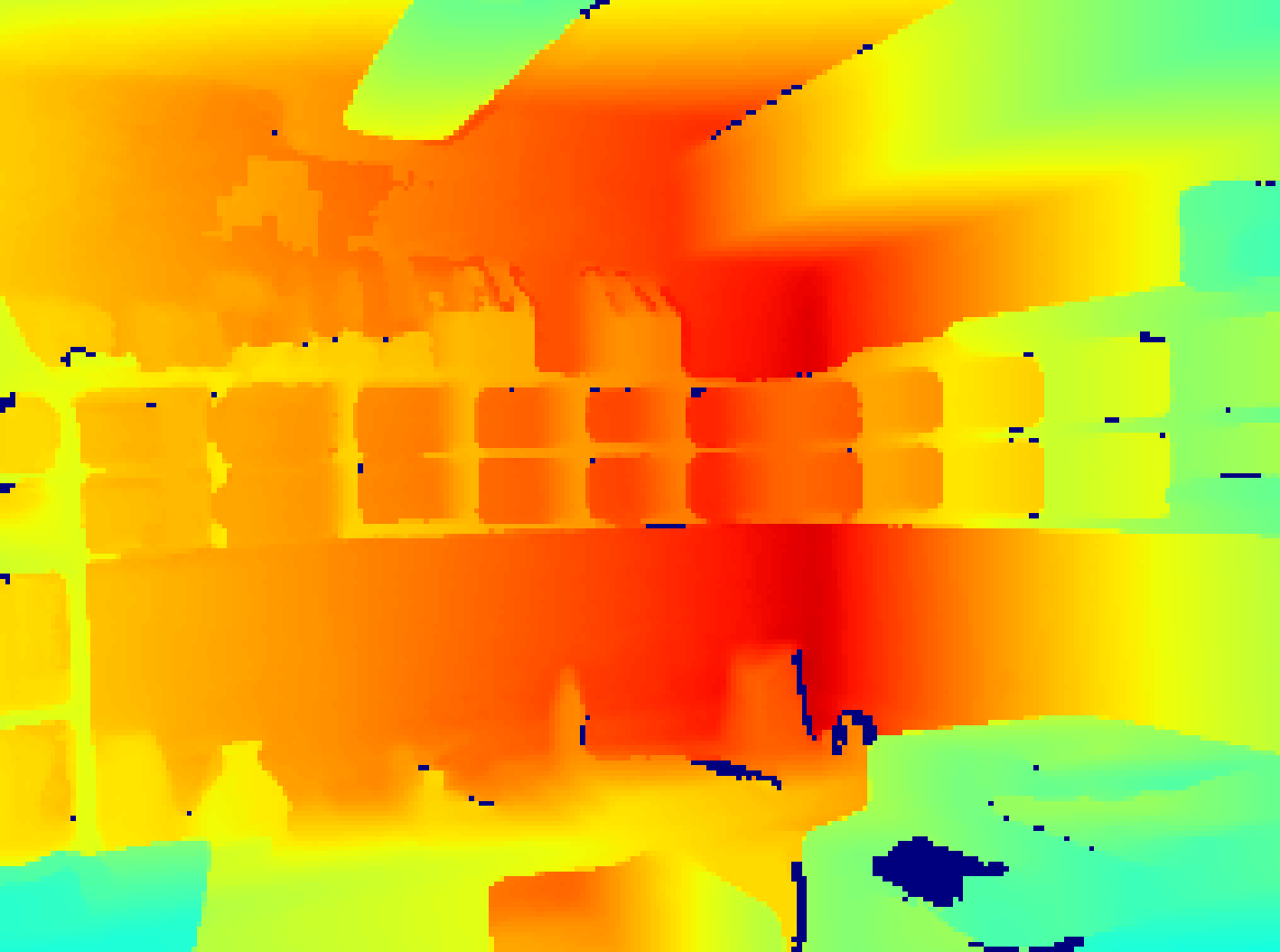}
        \caption{GT depth}
        \label{fig:gt_depth2v3}
    \end{subfigure}
    \centering
    \begin{subfigure}{0.3\linewidth}
    \includegraphics[width=1.0\linewidth]{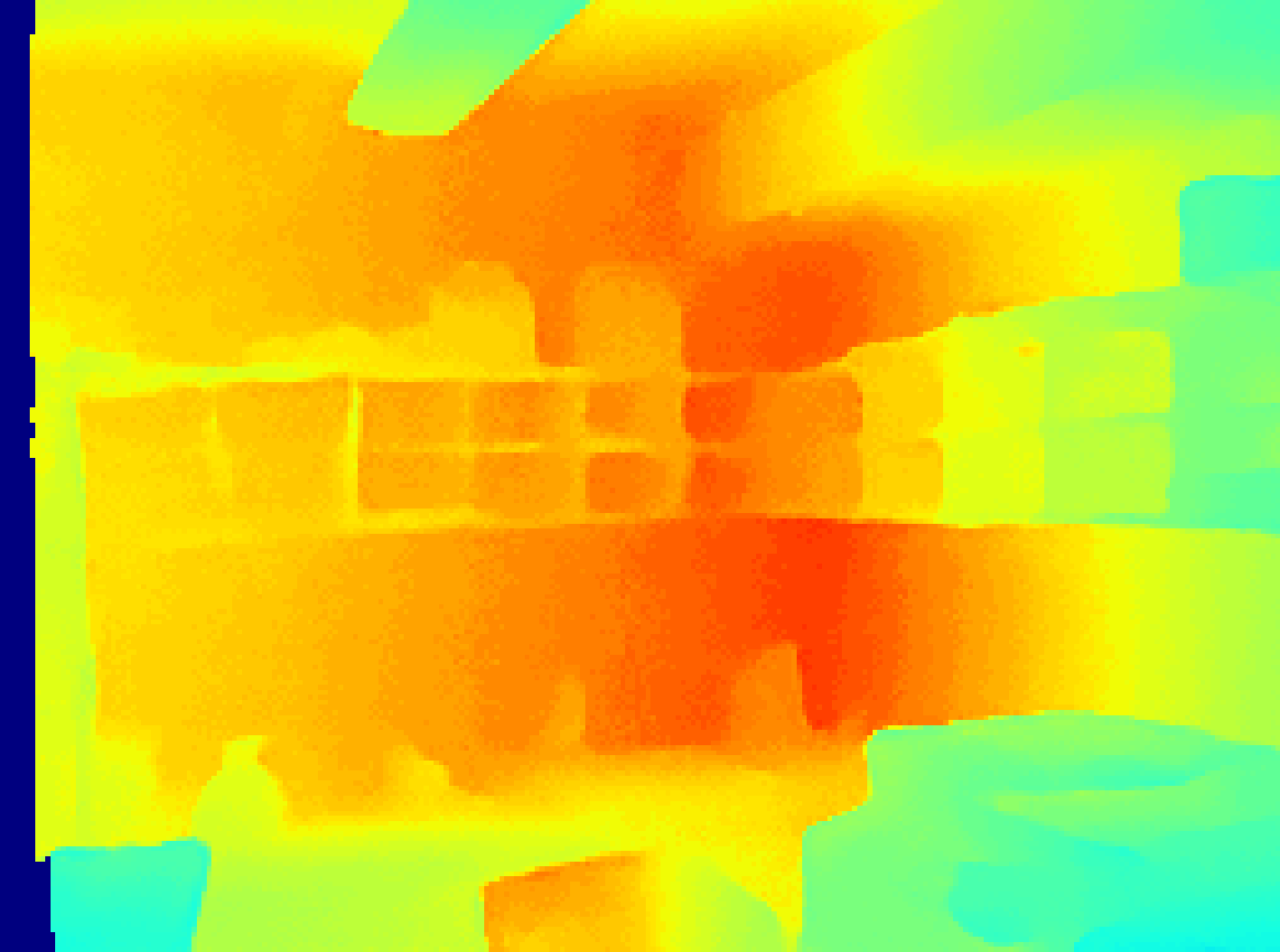}
        \caption{DPT depth}
        \label{fig:dispnetbase3}
    \end{subfigure}
    \centering
    \begin{subfigure}{0.355\linewidth}
    \includegraphics[width=1.0\linewidth]{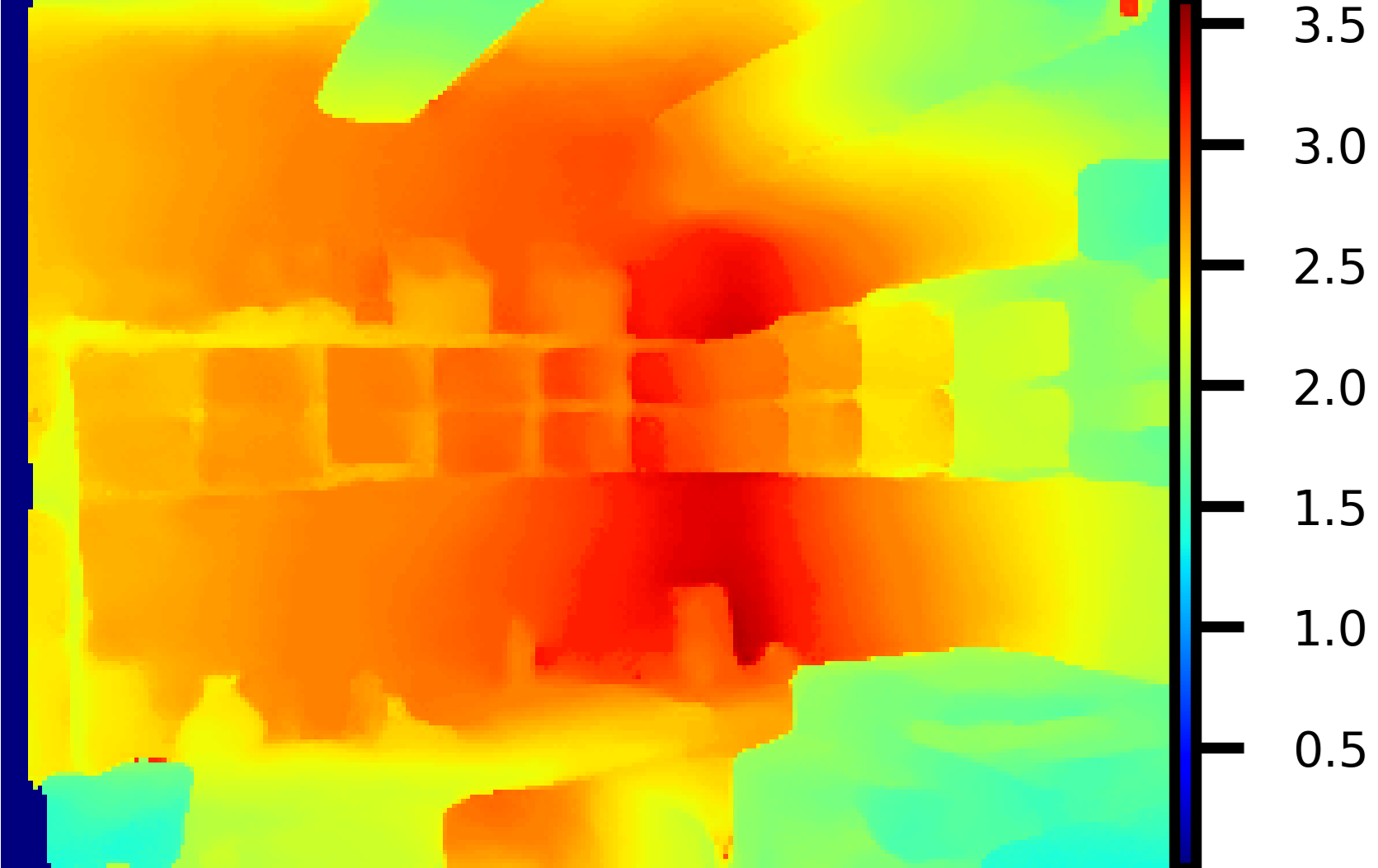}
        \caption{h3DPT depth$\phantom{xxx}$}
        \label{fig:h3dfs3}
    \end{subfigure}
    \centering 
   \begin{subfigure}{0.30\linewidth}
        \includegraphics[width=1.0\linewidth]{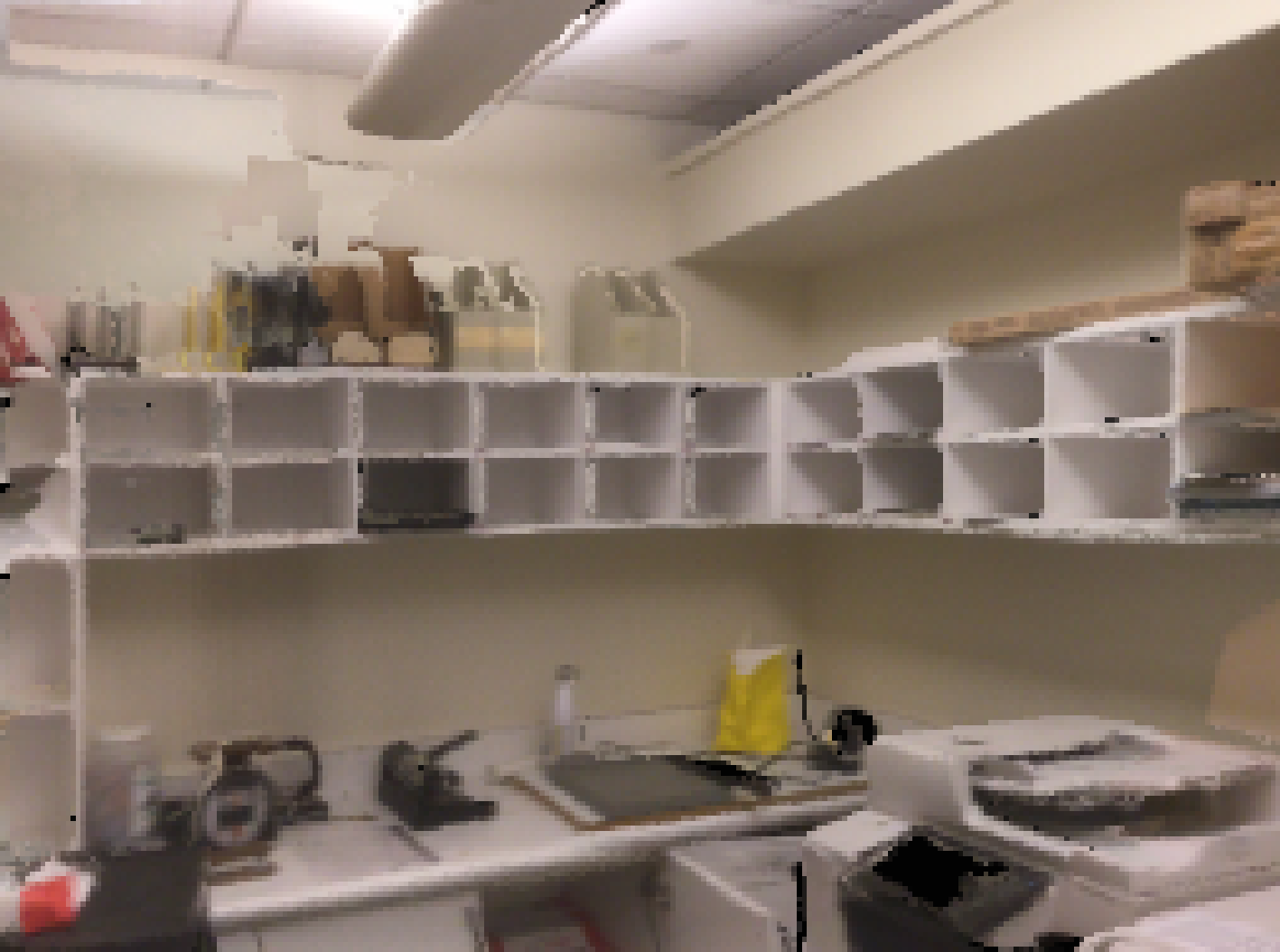}
        \caption{Image}
        \label{image0_30b}
    \end{subfigure}
    \centering
    \begin{subfigure}{0.3\linewidth}
    \includegraphics[width=1.0\linewidth]{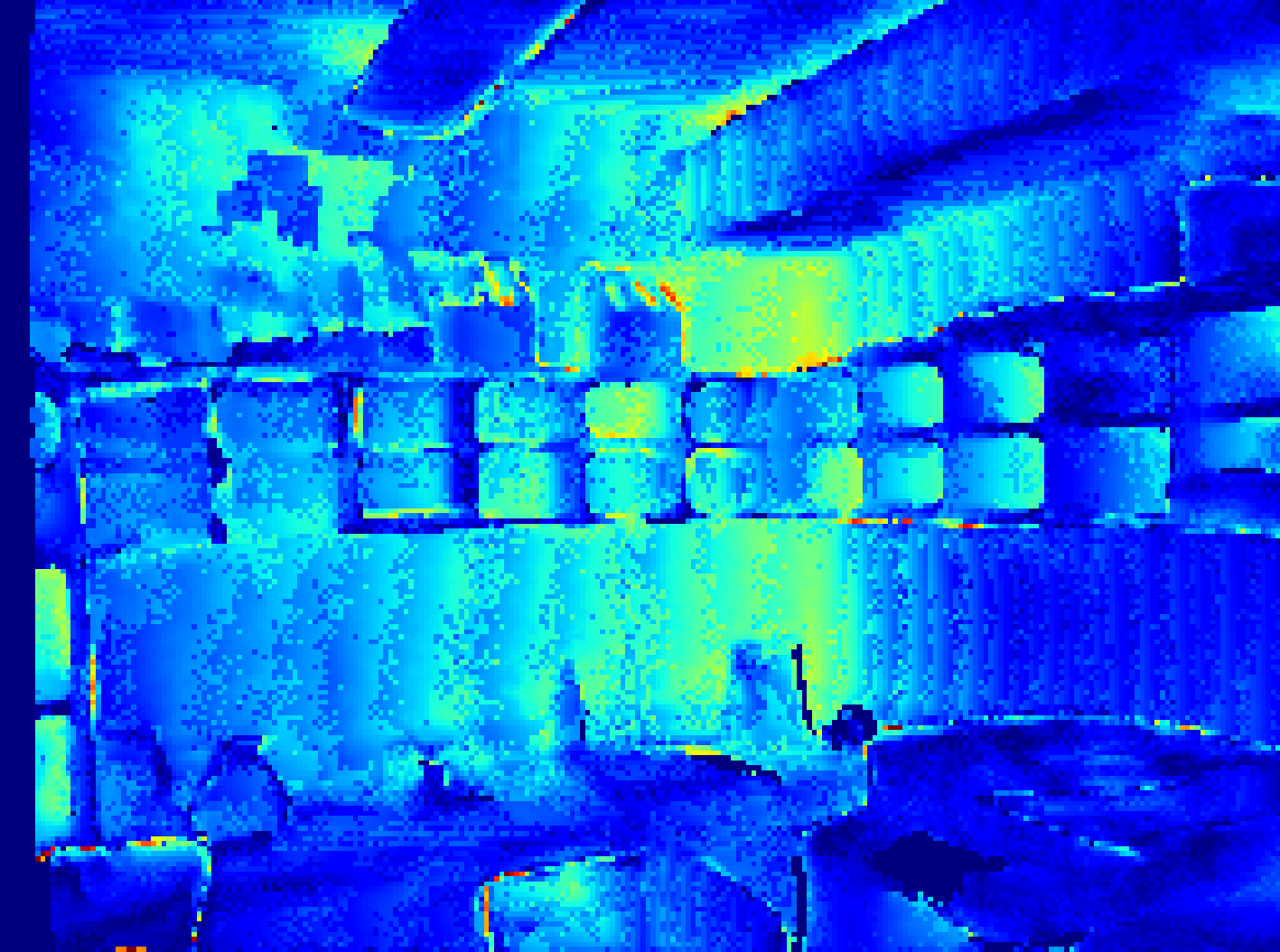}
    \caption{$|\text{GT - DPT}|$}
    \label{fig:gt_minus_dfs3}
    \end{subfigure}
\centering
    \begin{subfigure}{0.355\linewidth}
    \includegraphics[width=1.0\linewidth]{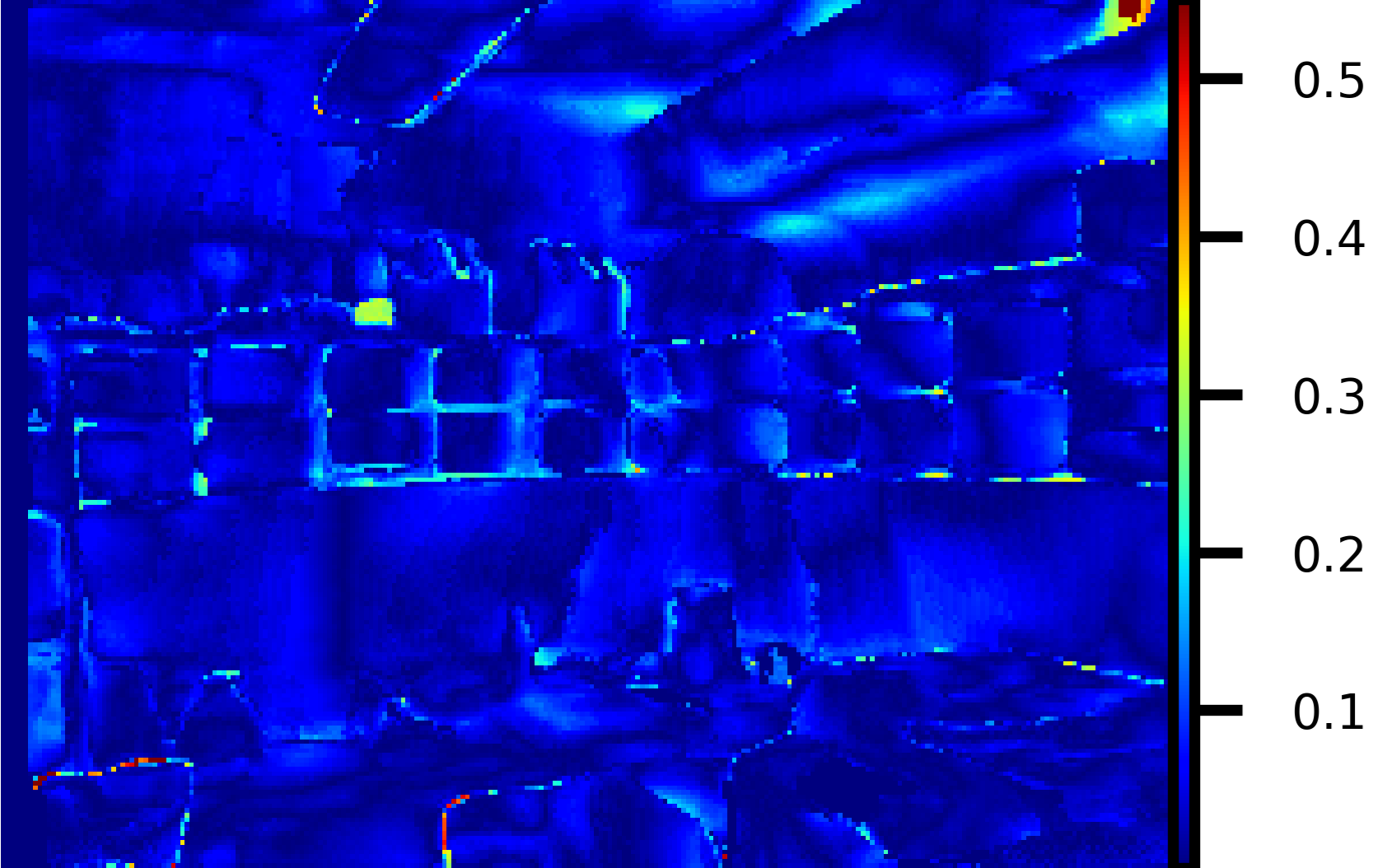}
    \caption{$|\text{GT - h3DPT}|\phantom{xxx}$}
    \label{fig:gt-h3dfs3}
    \end{subfigure}
    \caption{Depth errors of DPT and h3DPT models on DSP. ScanNet scene scene0804\_00.  All values are presented in meters.}
    \label{fig:depth_errorsdfs3}
\end{figure}

In Figs.~\ref{fig:depth_errors0}--\ref{fig:depth_errorsdfs3} we show additional examples of depth maps predicted by original and modified DispNet and DPT models. These examples represent both simpler (Figs.~\ref{fig:depth_errors0}--\ref{fig:depth_errorsdfs2}) and more complex scenes (Fig.~\ref{fig:depth_errorsdfs3}). In all examples, modified models have significantly smaller quantization error; remaining errors are concentrated on depth discontinuities. Errors in the vicinity of depth discontinuities are also present for FP32 models and not linked to the proposed approach.

\section{Human pose estimation experiment}
\label{sec:hpe_experiment}

Two approaches were realized for the Human pose estimation (HPE) experiment: direct keypoints regression~\cite{Toshev2013DeepPoseHP} and via representing keypoints as heatmaps~\cite{Xiao2018SimpleBF}. For training, the MS COCO~\cite{Lin2014MicrosoftCC} dataset with human pose keypoints labeling was used.
Both models have ResNet-RS~\cite{Bello2021RevisitingRI} as an encoder with alpha equals 0.5. The head for direct regression is composed of fully-connected layers as it was introduced by Toshev et~al. The model for heatmaps prediction includes DispNet decoder and convolutional head proposed by Xiao et~al.

For HPE experiment we trained two lightweight models: direct keypoints prediction with $AP_{50}$ = $43$, and heatmaps prediction with $AP_{50}$ = $83$. We found that the most stable way to train the modified HPE model is training the original model first, using it as a teacher model, and calculating GT Hilbert curve components $x_{\scriptscriptstyle\textrm{GT}}$ and $y_{\scriptscriptstyle\textrm{GT}}$ from the teacher model predictions in the loss term (\ref{eq:HILBERT_loss}). In this manner, the modified model can be trained even when GT data for calculation of $x_{\scriptscriptstyle\textrm{GT}}$ and $y_{\scriptscriptstyle\textrm{GT}}$ are not available, for example in case of heatmaps learning.

\begin{figure}
    \centering
    \begin{subfigure}{0.45\linewidth}
        \includegraphics[width=1.0\linewidth]{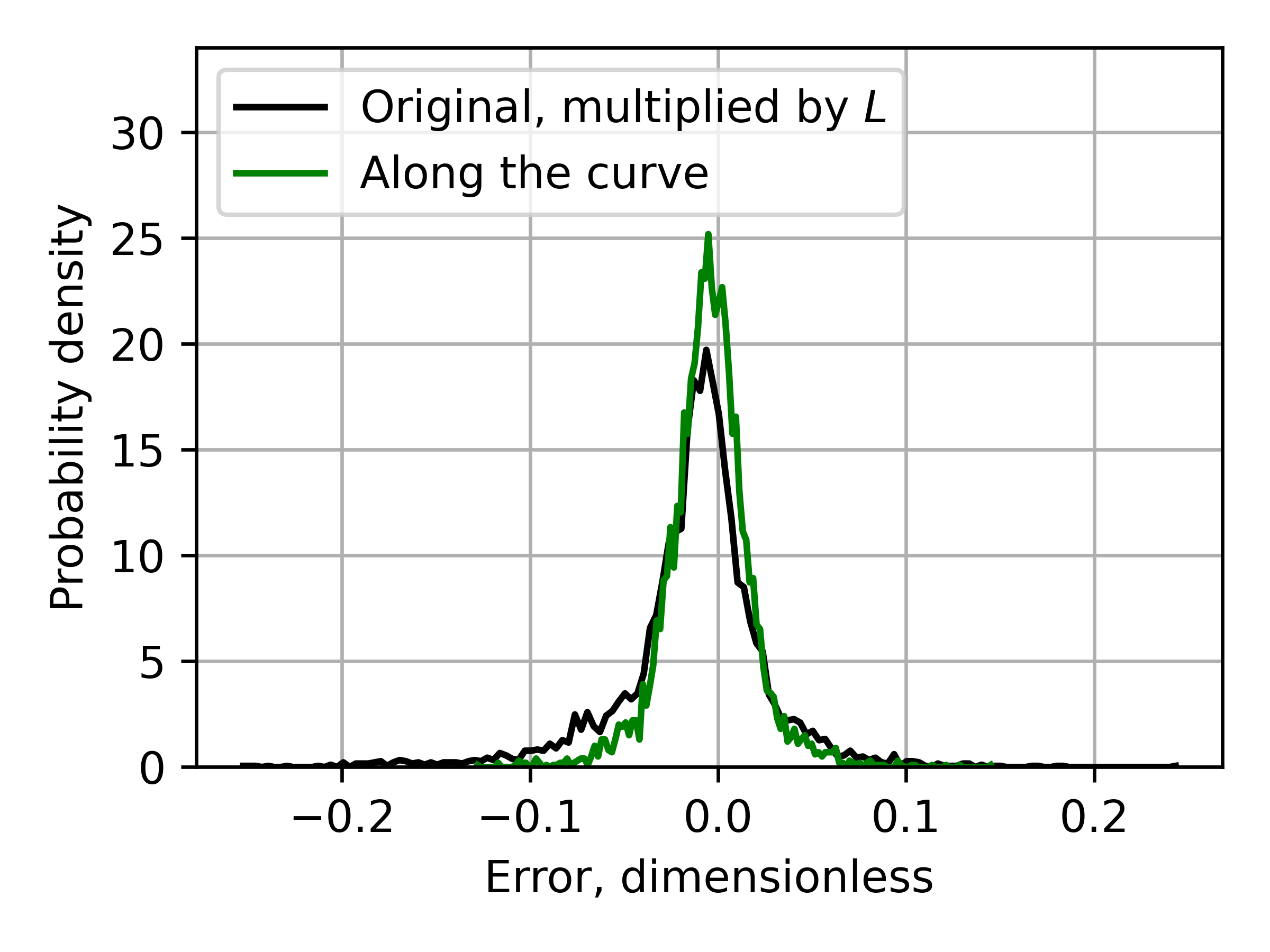}
        \caption{Keypoints error}
        \label{fig:key_error}
    \end{subfigure}\hfill
    \centering
    \begin{subfigure}{0.45\linewidth}
    \includegraphics[width=1.0\linewidth]{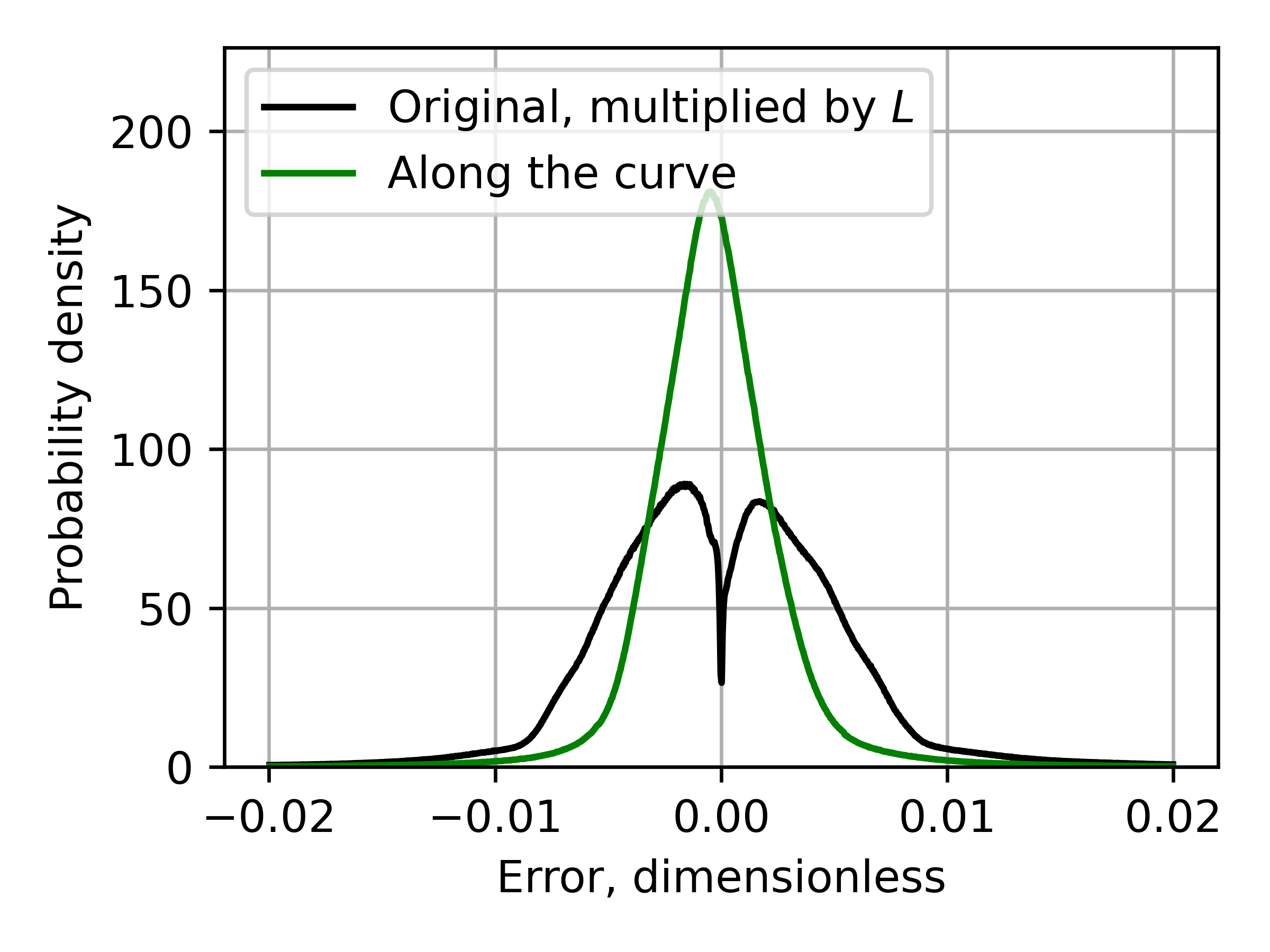}
       \caption{Heatmaps error}
       \label{fig:hm_error}
    \end{subfigure}
    \centering\hfill
    \begin{subfigure}{0.45\linewidth}
        \includegraphics[width=1.0\linewidth]{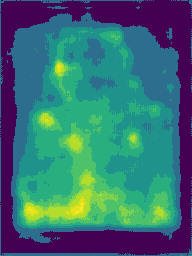}
        \caption{Heatmap by the original HPE model}
        \label{fig:hm_original}
    \end{subfigure}
    \centering\hfill
    \begin{subfigure}{0.45\linewidth}
    \includegraphics[width=1.0\linewidth]{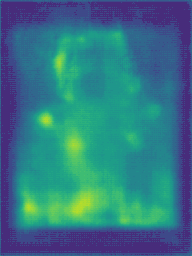}
        \caption{Heatmap by the modified HPE model}
        \label{fig:hm_modified}
    \end{subfigure}
    \caption{Quantization error for HPE using direct keypoints regression (a) and heatmaps (b). Comparison of heatmaps prediction on DSP using W8A8 models (c, d).}
    \label{fig:hpe_experiment}
\end{figure}

For model with direct keypoints prediction, along-the-curve quantization error for model with Hilbert curve-based representation for $p=1, 2$ increases approximately $L$ times as compared to the keypoints quantization error of the original model (see Fig. \ref{fig:key_error} for $p=1$ case); across-the-curve error is an order of magnitude smaller and is similar to the quantization error of the original model. As a result, quantization error of the modified model in keypoints space remains almost unchanged. This result suggests that for the direct regression task quantization error is strongly signal-dependent. The only effect from the proposed modification is in increase of output bit-precision.

The situation is different for the heatmaps prediction model. In this case, quantization error along the Hilbert curve of the second order increases less significantly as compared to the original model. From Fig.~\ref{fig:hm_error} it is seen that the along-the-curve error distribution is narrower than the original heatmaps quantization error multiplied by the curve length. As a result, quantization error of the modified model in heatmaps space is 2.69 times smaller than for the original model. In addition, the spatial structure of predicted heatmaps is improved as shown in Figs.~\ref{fig:hm_original} and \ref{fig:hm_modified}. The latter effect is caused by increase of the output bit-precision.

\end{document}